\documentclass[10pt]{article} 

\PassOptionsToPackage{numbers,square,sort&compress}{natbib}

\usepackage[preprint]{tmlr}

\usepackage{amsmath,amsfonts,bm}

\def\eqref#1{equation~\ref{#1}}

\def\1{\bm{1}}

\DeclareMathAlphabet{\mathsfit}{\encodingdefault}{\sfdefault}{m}{sl}
\SetMathAlphabet{\mathsfit}{bold}{\encodingdefault}{\sfdefault}{bx}{n}

\usepackage{hyperref}
\usepackage{url}

\usepackage{graphicx}%
\graphicspath{ {./figures_jpg/} }
\DeclareGraphicsExtensions{.jpg}

\usepackage{multirow}%
\usepackage{amsthm}%
\usepackage[title]{appendix}%
\usepackage{xcolor}%
\usepackage{manyfoot}%
\usepackage{booktabs}%
\usepackage{bm}

\usepackage[exponent-product = \cdot]{siunitx}

\usepackage{glossaries-extra}
\setabbreviationstyle{long-short-sc}
\newabbreviation{FEM}{FEM}{finite element method}
\newabbreviation{ML}{ML}{machine learning}
\newabbreviation{DL}{DL}{deep learning}
\newabbreviation{CDD}{CDD}{continuum dislocation dynamics}
\newabbreviation{PDE}{PDE}{partial differential equation}
\newabbreviation{pde}{PDE}{partial differential equation}
\newabbreviation{PSD}{PSD}{power spectral density}
\newabbreviation{RMSE}{RMSE}{root mean square error}
\newabbreviation{PCA}{PCA}{principal component analysis}
\newabbreviation{ood}{OOD}{out-of-distribution}
\newabbreviation{FNO}{FNO}{Fourier Neural Operator}
\newabbreviation{PINN}{PINN}{Physics-Informed Neural Network}
\newabbreviation{SciML}{SciML}{scientific machine learning}
\newabbreviation{DeepONet}{DeepONet}{Deep Operator Network}

\usepackage{subcaption}

\title{Out-of-distribution generalization of deep-learning surrogates for 2D PDE-generated dynamics in the small-data regime}

\author{\name Binh Duong Nguyen \thanks{Corresponding author: bi.nguyen@fz-juelich.de, s.sandfeld@fz-juelich.de} \\
	\addr Institute for Advanced Simulations -- Materials Data Science and Informatics (IAS-9), \\
	Forschungszentrum J\"ulich GmbH, \\
	J\"ulich 52425, Germany 
	\AND
	\name Stefan Sandfeld \footnotemark[1] \\
	\addr Institute for Advanced Simulations -- Materials Data Science and Informatics (IAS-9), \\
	Forschungszentrum J\"ulich GmbH, \\
	J\"ulich 52425, Germany 
	\AND
	\addr Chair of Materials Data Science and Materials Informatics, Faculty 5 -- Georesources and Materials Engineering, \\
	RWTH Aachen University, \\
	Aachen 52056, Germany \\
}

\def\month{MM}  
\def\year{YYYY} 
\def\openreview{\url{https://openreview.net/forum?id=XXXX}} 

\hypersetup{
	colorlinks=false,          
	pdfborder={0 0 1},         
	linkbordercolor={0 0 1},   
	citebordercolor={0 0 1},   
	urlbordercolor={0 0 1}     
}

\usepackage[noabbrev]{cleveref}
\crefname{equation}{eq.}{eqns.}
\crefname{figure}{Fig.}{Figs.}
\Crefname{figure}{Fig.}{Figs.}
\crefname{table}{Tab.}{Tabs.}
\Crefname{table}{Tab.}{Tabs.}

\begin{document}

	\maketitle
	
	\begin{abstract}
		Partial differential equations (PDEs) are a central tool for modeling the 
		dynamics of physical, engineering, and materials systems, but high-fidelity 
		simulations are often computationally expensive. At the same time, many 
		scientific applications can be viewed as the evolution of spatially distributed 
		fields, making data-driven forecasting of such fields a core task in scientific 
		machine learning. In this work we study autoregressive deep-learning surrogates 
		for two-dimensional PDE dynamics on periodic domains, focusing on generalization 
		to out-of-distribution initial conditions within a fixed PDE and parameter 
		regime and on strict small-data settings with at most $\mathcal{O}(10^2)$ 
		simulated trajectories per system. We introduce a multi-channel U-Net with 
		enforced periodic padding (me-UNet) that takes short sequences of past solution 
		fields of a single representative scalar variable as input and predicts the next 
		time increment. We evaluate me-UNet on five qualitatively different PDE families
		--- linear advection, diffusion, continuum dislocation dynamics, Kolmogorov 
		flow, and Gray--Scott reaction--diffusion---and compare it to ViT, AFNO, PDE-
		Transformer, and KAN-UNet under a common training setup. Across all datasets, me-UNet 
		matches or outperforms these more complex architectures in terms of field-
		space error, spectral similarity, and physics-based metrics for in-distribution 
		rollouts, while requiring substantially less training time. It also generalizes 
		qualitatively to unseen initial conditions 
		and, e.g.,  reaches comparable performance on continuum dislocation dynamics 
		with as few as $\approx 20$ training simulations. A data-efficiency study and 
		Grad-CAM analysis further suggest that, in small-data periodic 2D PDE settings, 
		convolutional architectures with inductive biases aligned to locality and 
		periodic boundary conditions remain strong contenders for accurate and 
		moderately out-of-distribution-robust surrogate modeling.
	\end{abstract}
	
	\section{Introduction}
	Scientific and engineering practice increasingly relies on numerical models and 
	data-driven methods to understand, predict, and control complex dynamical 
	systems. Many such systems can be viewed as the evolution of spatially 
	distributed fields---temperatures, concentrations of chemical species, velocity
	and pressure fields in fluids, elastic displacement and stress in solids, 
	electromagnetic fields, order parameters in phase-field models, probability 
	densities in population dynamics, or quantum-mechanical wave functions---over 
	time and space. A central goal of \gls{SciML}~~\cite{dietrich2025scientific}  
	is therefore to learn such field 
	dynamics from data, either in a purely data-driven fashion or by additionally 
	exploiting assumed or derived governing equations, for example by integrating 
	scientific knowledge with machine learning as discussed in recent surveys such 
	as~\cite{willard2023integrating}.
	
	The behavior of any real systems is extremely complex with a large number of
	unknown parameters and relations. Therefore, in practice, idealized 
	mathematical models expressed as systems of equations are used as a substitute. 
	For a large class of phenomena, these equations take the form of \glspl{PDE}, 
	which encode, e.g., local conservation laws, balance relations,	and 
	constitutive assumptions.  
	\Gls{PDE} models underpin applications from general relativity and quantum
	mechanics to fluid dynamics, elasticity, reaction--diffusion, and continuum 
	descriptions in materials science. Examples of famous \glspl{PDE} include 
	Einstein's equations of general relativity \cite{einstein1915feldgleichungen}, 
	which describe how the curvature of spacetime responds to the presence of 
	matter and energy, the Schrödinger equation \cite{schrodinger1926undulatory} 
	for the evolution of quantum-mechanical wave functions, and the Navier--Stokes 
	equations \cite{navier1821lois} governing the motion of viscous fluids. Their 
	flexibility and reach make \glspl{PDE} some of the most widely used tools for 
	representing spatio-temporal field evolution in science and engineering, even 
	though they always remain idealized approximations of reality. The particular 
	choice of \glspl{PDE} determines which classes of phenomena can be represented, 
	and the broad variety of \glspl{PDE} models reflects the breadth of field 
	dynamics encountered in practice.
	
	When high-fidelity \gls{PDE} models are available, they play a dual role in 
	\gls{SciML}. First, they are scientifically important in their own right: many 
	\glspl{PDE} of interest are challenging to solve numerically, and accurate 
	simulations can require substantial computational resources. Classical 
	examples include large-scale cosmological simulations such as 
	\emph{Illustris}~\cite{vogelsberger2014introducing}, which required more than 
	$8000$ CPU cores for several months of wall-clock time, or state-of-the-art 
	climate simulations that run for weeks to months on leadership-class 
	supercomputers with correspondingly large energy 
	consumption~\cite{randall2019100}. Second, \gls{PDE} solvers provide a 
	controlled and reproducible way to generate training data with known ground 
	truth. By sampling initial and boundary conditions, various parameters, and 
	forcing terms, one can construct benchmark datasets that probe a range of 
	dynamical regimes while avoiding experimental noise, measurement artifacts, 
	and uncertainty about the underlying equations. In this sense, 
	\gls{PDE}-based benchmarks are an attractive stepping stone towards the
	much harder problem of learning from real measurement data, where the 
	governing equations may be only partially known or not explicitly available. 
	Benchmark suites such as PDEBench~\cite{takamoto2022pdebench} already provide 
	standardized datasets and baselines for a broad range of time-dependent 
	\glspl{PDE}, primarily targeting large-data settings and broad parametric 
	coverage. Here we instead emphasize strict small-data regimes, periodic 2D 
	dynamics, and physics-based evaluation, aiming to mimic the constraints of 
	many real scientific applications where generating each high-fidelity 
	simulation is expensive.

	From this perspective, \glspl{PDE} in our work are not an end in themselves 
	but a convenient and stringent testbed for models that aim to learn complex 
	field dynamics. A \gls{SciML} method that fails to robustly learn dynamics 
	generated by well-posed \glspl{PDE}---in a setting where the discretization, 
	parameters, and numerical errors are under control---is unlikely to succeed 
	on heterogeneous, noisy, and partially observed real-world data. Conversely, 
	architectures that generalize well across \gls{PDE}-generated datasets,
	in particular under limited-data and \gls{ood} conditions, are promising 
	candidates for deployment on experimental or observational datasets. Thus, 
	our choice to study 2D \gls{PDE} dynamics is motivated both by the intrinsic 
	importance of \glspl{PDE} in science and engineering and by their role as a 
	controlled proxy for more complex real-world field evolution.
	
	\Gls{DL} surrogates fit naturally into this program. Instead of solving 
	the governing equations on the fly for each new query, one trains a parametric 
	model to approximate the time-advance operator that maps past field 
	configurations (and, where applicable, control parameters) to their future
	evolution. Once trained, such surrogates can produce approximate multi-step
	forecasts at drastically reduced inference cost compared to conventional 
	solvers, enabling tasks such as accelerated parameter studies, uncertainty 
	quantification, or real-time control. We refer to a sequence of such 
	predicted future fields, obtained by iteratively applying the learned 
	time-advance operator starting from a ground-truth context, as an 
	\emph{autoregressive rollout}.
	
	However, real scientific applications 
	typically operate in \emph{small-data regimes}, where only tens to hundreds 
	of high-fidelity simulations or experiments are available, and where test 
	conditions may differ markedly from those seen during training. In this 
	regime, the central challenge is not merely to fit a training distribution, 
	but to achieve qualitatively robust, moderately \gls{ood} generalization of 
	field dynamics while keeping training costs within the reach of typical 
	scientific users.

	A large body of work has explored neural networks as surrogates for 
	differential equations. Early approaches in the 1990s already exploited 
	universal approximation results for feed-forward 
	networks~\cite{hornik1989multilayer} to represent solutions of ordinary and 
	partial differential equations in collocation schemes, e.g. as in 
	\cite{lee1990neural, dissanayake1994neural, lagaris1998artificial}. More 
	recently, several influential lines of research have emerged. Neural 
	operators such as \gls{FNO}~\cite{li2020fourier} and 
	\glspl{DeepONet}~\cite{lu2021learning} aim to learn resolution- and 
	mesh-agnostic
	mappings between function spaces. They have demonstrated impressive 
	interpolation performance on benchmarks such as parametric Navier--Stokes 
	flows, often using on the order of $10^3$--$10^4$ simulated trajectories, 
	each of which consists of a 
	large number of time steps. \Glspl{PINN}
	\cite{raissi2019physics, zhu2019physics, shukla2020physics, zhang2020learning, ren2022phycrnet, ren2023physr, yuan2024f}
	take a complementary route by embedding the governing equations and boundary 
	conditions into the loss function via automatic differentiation, which allows 
	them to exploit sparse data but typically requires solving a challenging 
	\gls{PDE}-constrained optimization problem. Most recently, foundation-style 
	\gls{PDE} models such as PROSE-PDE~\cite{sun2025towards} and 
	PDE-Transformer~\cite{holzschuh2025pde} seek broad generalization across 
	systems and geometries through large-scale multi-task pretraining on billions
	of tokens of simulated data.
	
	Despite these advances, comparatively little work systematically investigates
	how relatively simple, carefully designed convolutional architectures perform 
	as time-stepping surrogates for 2D field dynamics under strict small-data 
	constraints and \gls{ood} initial conditions. Most existing benchmarks either 
	assume access to large numbers of simulated trajectories, focus on 
	in-distribution error metrics, or emphasize equation-informed and 
	foundation-scale models whose training requirements are far beyond the resources
	of many domain scientists. Moreover, \glspl{PDE} are typically treated as the 
	ultimate target of modeling, rather than as a controlled environment in which 
	to stress-test generalization in preparation for real measurement data.
	
	In this work, we adopt the latter viewpoint and use \gls{PDE}-generated field 
	dynamics as a controlled testbed for studying \gls{ood} generalization of 
	\gls{DL} surrogates in realistic small-data regimes. Concretely, we consider 
	two-dimensional, periodic field dynamics generated by five qualitatively 
	different \gls{PDE} families that cover transport, diffusion, pattern formation,
	and microstructure evolution. We then ask how different architectures behave 
	when trained on at most	a few dozen to one hundred simulations per \gls{PDE}, 
	and evaluated both on held-out simulations from the same distribution and on 
	initial conditions that differ substantially from those seen during	training. 
	Our focus is on autoregressive surrogates that operate as time-stepping models, 
	taking short sequences of past solution fields as input and predicting the next 
	temporal increment, and on how architectural inductive biases (such as 
	convolutional locality and periodic padding) influence data efficiency and 
	out-of-distribution behavior.

	\noindent%
	Our contributions are as follows:
	\begin{itemize}
		\item \textbf{A small-data benchmark for 2D periodic field dynamics.}
		We construct ten datasets derived from five representative \gls{PDE} families---linear advection, diffusion, continuum dislocation dynamics, Kolmogorov flow, and Gray--Scott reaction--diffusion---and use them as a controlled testbed for autoregressive forecasting on periodic $64\times64$ grids. For each system we define a strict small-data regime (at most $100$ simulations) and evaluate both in-distribution rollouts and generalization to qualitatively different initial conditions within the same \gls{PDE} and parameter regime.
		
		\item \textbf{A simple convolutional baseline with enforced periodicity.}
		We propose a multi-channel U-Net architecture with periodic padding (me-UNet) that operates as an incremental time-stepping surrogate: it takes a short sequence of past solution fields as input and predicts the next temporal increment. Under a common training protocol and fixed data budget, me-UNet consistently matches or outperforms several more complex neural-operator and transformer-based architectures (ViT, AFNO, \gls{PDE}-Transformer, and KAN-UNet) in terms of field-space error and spectral similarity, while requiring substantially less training time.
		
		\item \textbf{Physics-aware evaluation of long autoregressive rollouts.}
		We adopt an evaluation protocol that complements standard pixel-wise and spectral metrics with physics-aware diagnostics, tracking conserved or prescribed global quantities such as mass, energy, and total dislocation density over 100-step rollouts. Across \gls{PDE} families, me-UNet preserves these quantities more accurately than the competing architectures, including for moderately out-of-distribution initial conditions.
		
		\item \textbf{Data-efficiency and interpretability insights.}
		We perform a data-efficiency study that varies the number of simulations, time steps, and input sequence length, and we use Grad-CAM to analyze which spatial regions and U-Net blocks contribute most strongly to the predictions. These analyses indicate that convolutional inductive biases aligned with spatial locality and periodicity allow me-UNet to reach low-error regimes with as few as $\approx 20$ training simulations in continuum dislocation dynamics and to allocate representational capacity in a physically meaningful way.
	\end{itemize}

	Taken together, our results highlight that, in small-data, periodic 2D 
	field-dynamics settings, carefully designed convolutional baselines remain 
	strong contenders for accurate and moderately out-of-distribution-robust 
	surrogate modeling. More broadly, our study underscores the importance
	of evaluating \gls{SciML} architectures under realistic data constraints and 
	physics-aware metrics, and supports the use of \gls{PDE}-generated benchmarks 
	as a rigorous stepping stone toward models that can robustly learn from real 
	experimental and observational data.
	
	The remainder of this paper is organized as follows. 
	Sections~\ref{sec:math_model} and~\ref{sec:data_generation} introduce the 
	\gls{PDE} models and simulation datasets, and Section~\ref{sec:training} 
	describes the training setup. Section~\ref{sec:architecture} describes the 
	me-UNet and baseline architectures. Section~\ref{sec:results} presents our 
	empirical results, and Section~\ref{sec:discussion} discusses implications for 
	scientific machine learning and future work on real measurement data.

	\section{Mathematical model and neural network architecture}
	
	\subsection{Mathematical model}
	\label{sec:math_model}
	
	In this work we consider six mathematical models of increasing complexity, 
	listed in \cref{tab:abbrev}. They differ not only in the type of dynamics 
	(transport, diffusion, pattern formation, turbulence, microstructure evolution) 
	but also in the number and coupling of state variables.
	We organize them into three groups:
	\begin{itemize}
		\item \textbf{PDE-1--2: single-field linear transport and diffusion.}
		These are classical scalar advection and diffusion equations on periodic 
		domains, used as basic test cases for forecasting and spectral fidelity.
		
		\item \textbf{PDE-3--4: continuum dislocation dynamics (CDD)-based models.}
		These models describe the transport and evolution of possibly curved line-segments via various continuum densities. PDE-3a and PDE-3b are reduced \gls{CDD} systems 
		that capture expansion and motion of dislocation loops in a simplified setting 
		(with a smaller number of active fields and, in PDE-3a and 3b, curvature initialized 
		to zero), whereas PDE-4 is the full three-field \gls{CDD} model considered in 
		Appendix~\ref{sec:math_model} with an additional curvature density. 
		
		\item \textbf{PDE-5--6: multi-field fluid and reaction--diffusion systems.}
		PDE-5 is a Navier--Stokes system in Kolmogorov-flow configuration, evaluated 
		in terms of the scalar vorticity field, and PDE-6 is the Gray--Scott 
		reaction--diffusion system with two reacting species.
	\end{itemize}
	
\begin{figure*}[h!tb]
		\centering
		\begin{subfigure}{1.0\textwidth}
			\centering
			\includegraphics[width=0.75\textwidth]{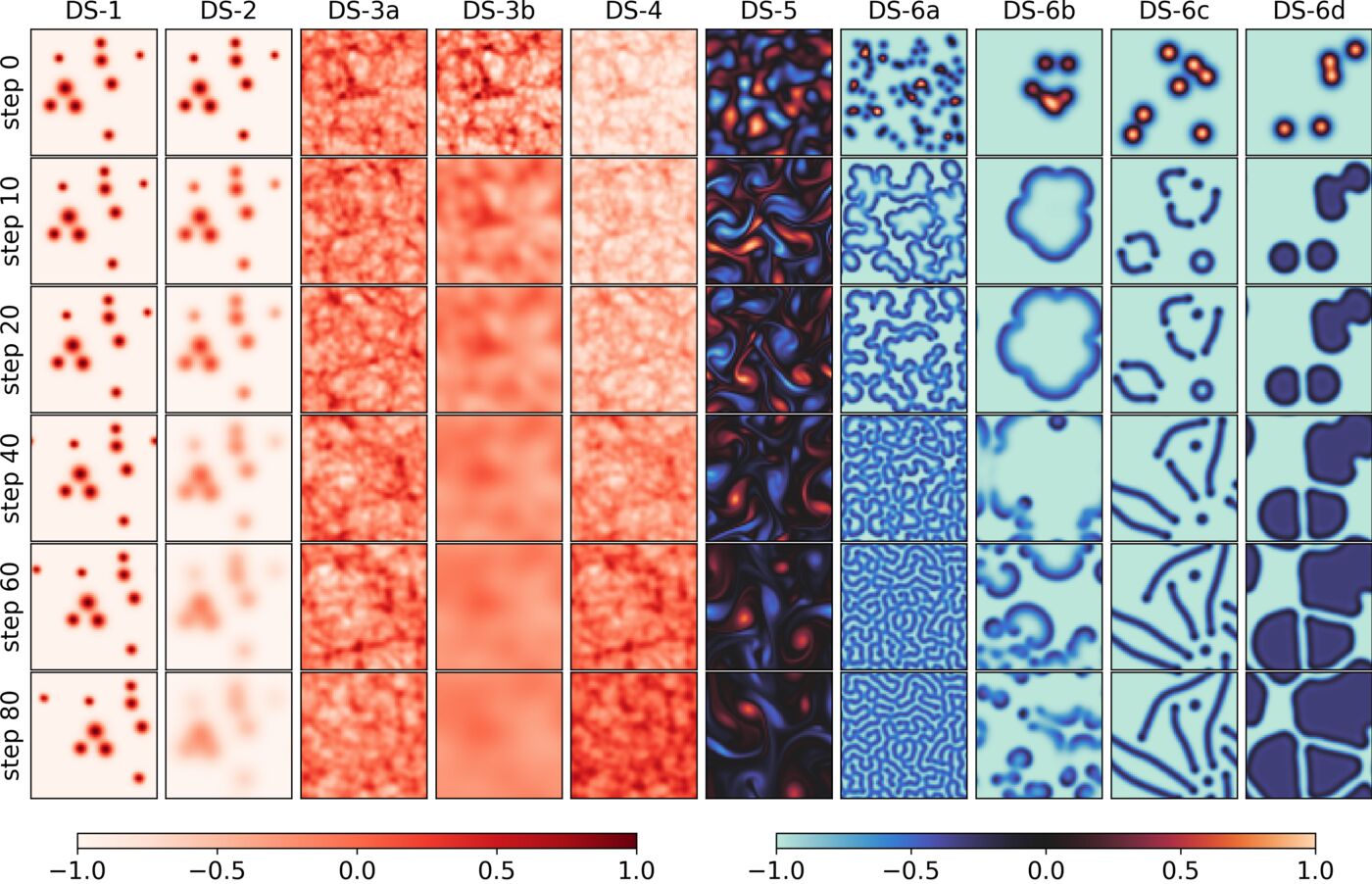}
			\bigskip
		\end{subfigure}
		\caption{Examples of simulation results from all mathematical models.
			Datasets DS-1 and 2 are obtained from simple convection and diffusion equations, while the complexity of the \glspl{PDE} and the number of involved state variables increases towards the right: DS-3a, 3b, and 4 are obtained from \gls{CDD} \glspl{PDE}, DS-5 from Navier-Stokes equations, while the variants of DS-6 are obtained from the Gray-Scott model.}
		\label{fig:example_mm}
	\end{figure*}

	Example simulations from all six models are shown in \cref{fig:example_mm}. 
	Together they cover a range from simple scalar transport and diffusion to more 
	complex systems such as Kolmogorov flow, Gray--Scott pattern formation, and the 
	statistical evolution of systems of curved lines in \gls{CDD}. Detailed 
	descriptions of each model and the chosen parameters are provided in the Appendix.

	\begin{table}[htb!]
		\caption{Overview of the six \gls{PDE} models and resulting datasets used in this work.
			The column ``Fields'' refers to the number and type of continuous fields used for the numerical solution (``simulation''). The training dataset uses only \emph{a single field} of that simulation as input to the ML model, cf.\ \cref{sec:architecture}.}
		\label{tab:abbrev}
		\centering
		\resizebox{\textwidth}{!}{
			\begin{tabular}{clcc|cc}
				\toprule
				\multirow{2}{*}{\bf No.} & \multicolumn{1}{c}{\multirow{2}{*}{\bf Mathematical model}} & \multicolumn{2}{c}{\bf Simulation} & \multicolumn{2}{c}{Training} \\ 
				\cmidrule(r){3-6}
				& \multicolumn{1}{c|}{}                                    & \multicolumn{1}{c}{\bf Equation}    & \multicolumn{1}{c|}{\bf Fields}    & \multicolumn{1}{c}{\bf Datasets}      & \multicolumn{1}{c}{\bf Features}     \\
				\cmidrule(r){1-6}
				1   & Advection of a distribution of blobs                      & PDE-1	& 1 (scalar)               		& DS-1							& 1 (concentration)                       \\
				\cmidrule(r){1-6}
				2   & Diffusion of a distribution of blobs                      & PDE-2	& 1 (scalar)               		& DS-2                        	& 1 (concentration)	\\
				\cmidrule(r){1-6}
				\multirow{2}{*}{3}   & Expansion and diffusion of         & \multirow{2}{*}{PDE-3}	& \multirow{2}{*}{3 (1 scalar, 1 vectorial)}   		& \multirow{2}{*}{DS-3a, DS-3b}                        	& \multirow{2}{*}{1 (total dislocation density)}	\\
				& a distribution of loops (reduced CDD)        					& 		&    							&                        		& 	\\
				\cmidrule(r){1-6}
				5   & Statistical evolution of systems of curved lines (CDD)    & PDE-4	& 4 (2 scalars, 1 vectorial)  		& DS-4                        	& 1 (total dislocation density)	\\
				\cmidrule(r){1-6}
				6   & Kolmogorov flow (Navier--Stokes, vorticity formulation)   & PDE-5	& 2 (1 vector)     				& DS-5                        	& 1 (vorticity)	\\
				\cmidrule(r){1-6}
				\multirow{2}{*}{7}   & \multirow{2}{*}{Reaction--diffusion (Gray--Scott model)}                   & \multirow{2}{*}{PDE-6}	& \multirow{2}{*}{2 (scalar} 	& DS-6a, DS-6b,   	& \multirow{2}{*}{1 (concentration)}	\\
				&			& 			& 			& DS-6c, DS-6d  	& 	\\
				\bottomrule
			\end{tabular}
		}
	\end{table}
	
	To keep comparisons uniform and to mimic partial observability in real measurements, we train all models on rollouts of a single representative scalar field per system (Table~\ref{tab:abbrev}), even when the underlying \gls{PDE} couples multiple state variables.

	\subsection{Simulation datasets and training samples}
	\label{sec:data_generation}
	
	The dataset abbreviations are listed in \cref{tab:abbrev}: for PDE-1, 2, 4, 5 we obtain datasets DS-1, 2, 4 and 5; for PDE-3 we consider 2 parameter settings, denoted DS-3a and DS-3b; 
	for PDE-6 we consider four different parameter settings, denoted DS-6a, DS-6b, DS-6c, and DS-6d.
	Each of the resulting $10$ datasets consists of $110$ simulations.
	Examples from each dataset are shown in \cref{fig:example_mm}, and detailed descriptions are given in the Appendix.
	For each dataset, $100$ simulations are used for training, and the remaining $10$ are kept untouched
	for validation and testing.
	Each simulation contains $100$ time steps.
	
	For autoregressive training, the input data are constructed by sampling a random time index $n$ with
	$7 \leq n \leq 99$ (using one-based indexing for time steps).
	For each such $n$, we stack the $7$ consecutive fields
	\begin{align}
		u^{n-6}, u^{n-5}, \dots, u^{n}
	\end{align}
	along the channel dimension to form a multi-channel input image.
	The corresponding target is an \emph{increment field} that represents the change between time steps $n$ and $n+1$,
	\begin{align}
		\Delta u^{n+1} = u^{n+1} - u^{n}.
	\end{align}
	Thus, our models are trained to predict $\Delta u^{n+1}$ given the past sequence
	$\{u^{n-6}, \dots, u^{n}\}$, rather than the absolute field $u^{n+1}$.
	During rollout, the next state is obtained via the residual update
	\begin{align}
		\hat{u}^{n+1} = \hat{u}^{n} + \Delta \hat{u}^{n+1},
	\end{align}
	where hats denote model predictions.
	Focusing on local temporal increments stabilizes long autoregressive rollouts:
	the network only needs to learn short-term corrections instead of reconstructing the full state,
	and it can concentrate on the dynamics of the system rather than on absolute field values.
	This residual parameterization empirically leads to substantially more stable long-horizon predictions,
	which is consistent with the fact that many PDE solutions change gradually between adjacent time steps.
	
	All data values are scaled to the interval $[-1, 1]$ separately for each simulation and state variable.
	For a given state variable, the minimum and maximum over all time steps within a simulation are used
	to linearly map the values in each frame to the required range; scaling is not recomputed per frame.
	This per-simulation, per-variable normalization follows common practice in \gls{PDE} surrogate modeling,
	but it can have drawbacks in multi-field systems, which we discuss later in Appendix~\ref{app:physics_diagnostic_results}.
	All images are resized to $64 \times 64$ pixels.
	
	\paragraph{In-distribution vs out-of-distribution initial conditions}
	For each PDE family we fix the governing equations, physical parameters, numerical scheme, and spatial resolution. Training and test simulations always share this fixed configuration. The only source of distribution shift we study in this work is the distribution of initial conditions: ``in-distribution'' (ID) test rollouts use the same random initial-condition generator as training, whereas ``out-of-distribution'' (OOD) test rollouts use qualitatively different initial-condition families within the same PDE and parameter regime (e.g., sparse loops or line-like structures for CDD, or structured line/blob perturbations for Gray--Scott). Thus, throughout this paper, OOD refers specifically to shifts in the initial-condition distribution under a fixed PDE and parameter setting.

	\subsection{Neural network architecture}
	\label{sec:architecture}
	
	\begin{figure*}[h!tb]
		\centering
		\begin{subfigure}{1.0\textwidth}
			\centering
			\includegraphics[width=1.0\textwidth]{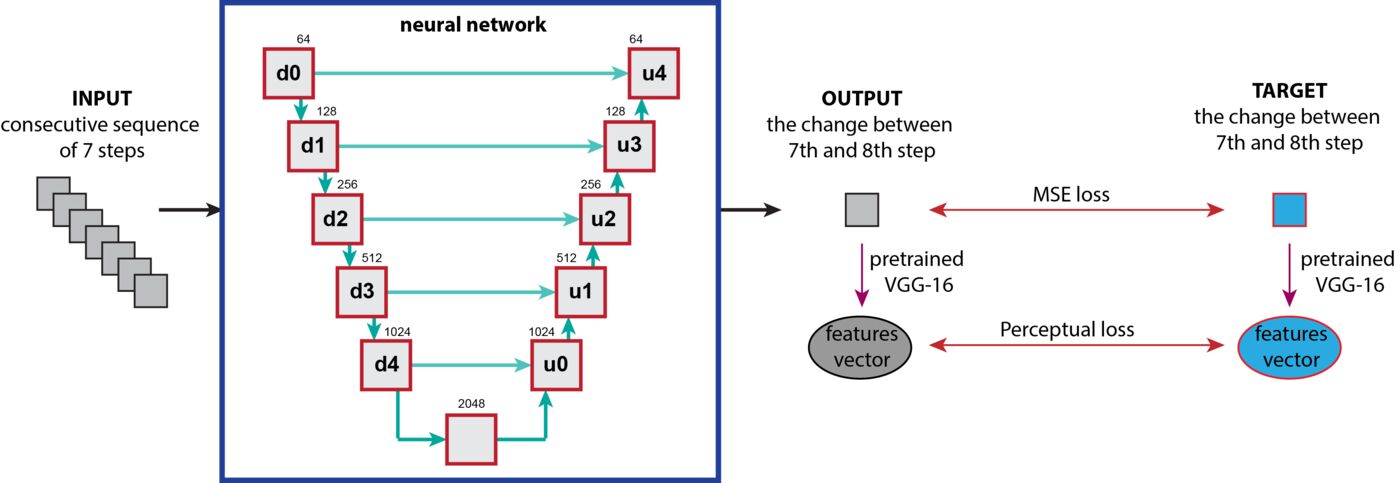}
			\bigskip
		\end{subfigure}
		\caption{Illustration of the me-UNet architecture, showing the encoder--decoder structure, periodic padding, and multi-channel input/output setup for autoregressive time-stepping.}
		\label{fig:me_UNet}
	\end{figure*}
	
	Our U-Net architecture is an extension of the original architecture by Ronneberger et al.~\cite{ronneberger2015u} and consists of two parts: an encoder (contracting path) that captures the spatial context of the input fields, and a decoder (expanding path) that enables precise localization of features. The input consists of a sequence of images, i.e., even though some of the original \glspl{PDE} consist of multiple or vectorial field quantities, we only use a single, scalar field. The encoder comprises five stages, each with a double convolutional module followed by an average-pooling operator for downsampling. The decoder also contains five stages, each with a transposed convolution operator for upsampling followed by a double convolutional module. Between encoder and decoder we place a double convolutional module acting as a bottleneck. 
	
	Each double convolutional module consists of two convolutional layers, each followed by batch normalization and a nonlinear activation function. A final convolutional layer is added at the end of the network to reduce the number of channels to match the number of output state variables of the considered problem. In total, the network has 23 convolutional layers, five average-pooling operators for downsampling, and five transposed convolution operators for upsampling. Skip connections between encoder and decoder stages allow low-level features from earlier layers to be combined with higher-level features in the decoder. This feature fusion helps preserve spatial detail and improves accuracy in tasks that require precise localization.
	
	The first substantial modification in our approach is the use of \emph{periodic padding}. In order to enforce periodic boundary conditions during training, we apply periodic padding in all convolutional layers whenever the data have periodic boundary conditions~\cite{qu2022learning, rao2023encoding}. In this way, prior knowledge about the boundary behavior is encoded directly into the network architecture so that the trained model cannot ignore periodicity. Concretely, instead of using the default zero padding in the \texttt{Conv2d} layers, we pad by copying values from the opposite side of the image: the leftmost columns are copied to the right, the rightmost columns to the left, and analogously for the top and bottom rows, yielding a fully periodic domain. This can also be implemented using the PyTorch built-in \texttt{CircularPad2d} operator. 
	
	We also adjust the downsampling operator by using average pooling with stride $2$ instead of max pooling. According to~\cite{zhao2024improved}, average pooling is a spatially invariant operation whose output does not depend on the precise position of features within the pooling window, whereas max pooling can neglect weaker but still important features. In our experiments, average pooling leads to more stable rollouts and better preservation of fine-scale structures.
	
	A second design aspect is the ability to adjust the number of input and output channels. In general, the architecture can handle multiple state variables and multiple time steps as separate channels. In our setup we use an input sequence of $L=7$ past time steps of a single representative state variable per \gls{PDE}, so the input has seven channels. For systems with multiple state variables (e.g., the \gls{CDD} model involves four fields), the architecture could in principle output several channels; however, for the cross-dataset comparisons in this work we restrict training and evaluation to one state variable per \gls{PDE} to keep the setup consistent across datasets. The choice of sequence length $L$ is a hyperparameter that controls how much temporal context the model sees relative to the intrinsic time scale of the dynamics. For the datasets considered here, the simulation time steps were chosen such that the evolution over six consecutive intervals exhibits nontrivial but still smooth changes; for systems with slower or faster dynamics, one would typically adapt $L$ or sub-sample time steps accordingly.
	
	We denote by $f_\theta$ the parametric mapping implemented by a given architecture, with trainable parameters $\theta$. All architectures considered in this paper, including me-UNet and the baselines, are trained in an \emph{incremental} or residual form. Given an input sequence of $L$ past fields
	\begin{align}
		\{u^{n-L+1}, \dots, u^{n}\} \subset \mathbb{R}^{H \times W},
	\end{align}
	the network does not predict the next field $u^{n+1}$ directly. Instead, it predicts a temporal update
	\begin{align}
		\Delta u^{n+1} = f_\theta\big(u^{n-L+1}, \dots, u^{n}\big),
	\end{align}
	and the next state is obtained as
	\begin{align}
		u^{n+1} &= u^{n} + \Delta u^{n+1}.
		\label{eq:increment-update}
	\end{align}
	Thus, for $L=7$ the output image represents the change in the field between the $7^\text{th}$ and $8^\text{th}$ time step. This residual-style parameterization empirically stabilizes long autoregressive rollouts and encourages the network to focus on learning short-time dynamics rather than absolute field values.
	
	To enable a fair comparison with other architectures, we keep the input sequence length (seven consecutive time steps), the incremental output representation in~\eqref{eq:increment-update}, and the loss function fixed across models. Only the internal architecture of the mapping $f_\theta$ is changed. Specifically, in addition to me-UNet we adapt four alternative architectures: a vision transformer (ViT)~\cite{dosovitskiy2020image}, an adaptive Fourier neural operator (AFNO), which is the core architecture of FourCastNet~\cite{pathak2022fourcastnet}, the PDE-Transformer (PDE-T)~\cite{holzschuh2025pde}, and a KAN-UNet in which the double convolutional layers of U-Net are replaced by KAN convolutional layers~\cite{xiangbo2024}.
	
	During training we minimize a combination of a pixel-wise loss $\mathcal L_{\mathrm{MSE}}$ and a perceptual loss $\mathcal L_{\mathrm{pc}}$. The total loss $\mathcal L$ for a predicted field $\hat{y}$ and reference field $y$ is
	\begin{align}
		\mathcal L(\hat{y}, y) &= \mathcal L_{\mathrm{MSE}}(\hat{y}, y) + \lambda_{\mathrm{pc}}\, \mathcal L_{\mathrm{pc}}(\hat{y}, y),
	\end{align}
	where $\mathcal L_{\mathrm{MSE}}$ is the mean-squared error defined in \cref{eq:mse_loss}, and $\mathcal L_{\mathrm{pc}}$ is the perceptual loss based on feature maps $\Phi_i$ of a pre-trained VGG-16 network as in \cref{eq:perceptual_loss}. In all experiments we keep the VGG-16 weights fixed and use feature maps from layer \texttt{relu2\_2}, and we set $\lambda_{\mathrm{pc}} = 1$.

	\subsection{Training setup}
	\label{sec:training}
	
	For each \gls{PDE} dataset and each architecture, we train a separate model using the same training protocol. Unless otherwise stated, we use the Adam optimizer with an initial learning rate of $1\times 10^{-4}$, weight decay $1\times 10^{-5}$, and a batch size of $40$. Each model is trained for $1000$ epochs without early stopping, and we select the final model based on the validation loss. 
	
	For me-UNet, the channel widths at the five encoder levels are 
	$[64, 128, 256, 512, 1024]$ and mirrored in the decoder. For the 
	baselines, we follow standard configurations adapted to our 
	$64\times 64$ input size: for ViT we use a patch size of $8$, 
	embedding dimension $768$, depth $12$, and $8$ attention heads; for 
	AFNO we use a patch size of $4$, embedding dimension $768$, depth 
	$12$, MLP ratio $4$, and $16$ AFNO blocks; for PDE-Transformer we 
	use the PDE-B configuration of~\cite{holzschuh2025pde}; and for KAN-
	UNet we mirror the me-UNet channel widths in encoder and decoder, 
	replacing the \texttt{Conv2d} layers in the double-convolution 
	blocks by \texttt{FastKANConvLayer} modules with Chebyshev basis 
	functions. All other architectural details are kept as close as 
	possible to the original implementations referenced above.

	Periodic boundary conditions are implemented in most of the mentioned architectures: in AFNO, periodic padding is used through an additional 2D convolutional layer \cite{pathak2022fourcastnet}; PDE-Transformer mimics periodic boundary conditions by rolling the tokens along the x- and y-axis when shifting the attention windows \cite{holzschuh2025pde}; KAN-UNet implements periodic padding similar to me-UNet. Only our implementation of the vision transformer use the vanilla architecture.
	
	\subsection{Evaluation metrics and physics-aware measures}
	\label{sec:metrics}
	
	We evaluate all models using three complementary criteria: (i) field-space error, (ii) spectral similarity, and (iii) physics-aware metrics that monitor conserved or prescribed global quantities.
	
	\paragraph{Field-space error}
	For each rollout we compute the \gls{RMSE} between the predicted fields $\hat{y}$ and reference fields $y$, averaged over spatial grid points, output channels, and time steps:
	\begin{align}
		\mathrm{RMSE}
		= \sqrt{\frac{1}{N T C} \sum_{t=1}^{T} \sum_{c=1}^{C} \sum_{i=1}^{N} \bigl(\hat{y}_{t,c,i} - y_{t,c,i}\bigr)^2},
	\end{align}
	where $N$ is the number of spatial grid points, $C$ the number of state variables, and $T$ the number of time steps in the rollout.
	
	\paragraph{Spectral similarity}
	To compare spatial frequency content we compute the azimuthally averaged \gls{PSD} for each snapshot and channel similar to what is done in \cite{koehler2024apebench, nguyen_efficient_2024}. 
	Let $P_{t,c}(k)$ and $\hat{P}_{t,c}(k)$ denote the \gls{PSD} of the reference and predicted fields, respectively, as a function of the radial wavenumber $k$ 
	(obtained by averaging $|\mathcal{F}[y_{t,c}]|^2$ and $|\mathcal{F}[\hat{y}_{t,c}]|^2$ over angular directions in Fourier space). 
	For each time step $t$ and channel $c$ we then compute the cosine similarity
	\begin{align}
		\mathrm{cos\_sim}_{t,c}
		= \frac{\sum_{k} \hat{P}_{t,c}(k)\, P_{t,c}(k)}
		{\sqrt{\sum_{k} \hat{P}_{t,c}(k)^2}\,\sqrt{\sum_{k} P_{t,c}(k)^2}}.
	\end{align}
	The reported spectral similarity is the mean of $\mathrm{cos\_sim}_{t,c}$ over channels and time steps.

	\paragraph{Physics-based metrics}
	In addition to these generic metrics we track physically meaningful global quantities for each \gls{PDE}. All such quantities are defined as spatial integrals (or sums over grid points) of appropriate functions of the state variables. For Kolmogorov flow we monitor the kinetic energy $E(t)$ of the velocity field (obtained from the vorticity field) and compare the temporal evolution of $E(t)$ between reference and prediction. For the \gls{CDD} model we monitor the spatial integral of the total dislocation density $\rho_t(t)$, which is known to follow a prescribed linear evolution according to the governing equations, and we assess how closely the learned surrogate reproduces this relationship. For Gray--Scott systems we track the total mass of the reactants (spatial integrals of $u$ and $v$), which should remain approximately conserved up to reaction terms, and compare their evolution under the learned models. For the advection equations we track conservation of total mass, which should remain constant during transport up to numerical errors. 
	
	For each architecture and dataset we report summary statistics of these diagnostics over the rollout horizon in \cref{sec:results}. The normalized absolute errors of the monitored quantities are visualized in \cref{fig:physics_diagnostic}.

	\subsection{Software and implementation details}
	\label{sec:software}
	
	All neural-network models are implemented in PyTorch~\cite{paszke2019pytorch} using Python~3.10.12. Experiments are run on a workstation equipped with an NVIDIA RTX~A6000 GPU with 48~GB of memory and a multi-core CPU, using CUDA~12.8 and cuDNN~91002. The me-UNet architecture and training loops are implemented from scratch, while the baseline models (ViT, AFNO, PDE-Transformer, KAN-UNet) build on publicly available reference implementations that we adapt to our input--output format and loss functions.
	
	The synthetic datasets are generated with finite-element and finite-volume solvers implemented in C++/Python using FEniCS~\cite{logg2012automated} for PDE-1--PDE-4, JAX-CFD~\cite{kochkov2021machine} for PDE-5, and exponax~\cite{koehler2024apebench} for PDE-6. For each \gls{PDE} dataset we fix the numerical scheme, time step, and resolution as described in \cref{sec:math_model,sec:data_generation}. Random initial conditions are sampled using NumPy~1.26.4 with a fixed random seed, and we use consistent seeds across models to ensure that training and evaluation splits are identical.
	
	All configuration files specifying all hyperparameters (learning rates, batch sizes, channel widths, etc.) are provided in the accompanying code repository. Where possible we also release pre-trained model checkpoints and scripts to regenerate all figures and tables from this paper, in order to facilitate reproducibility and further comparison.

	\section{Results}\label{sec:results}
	We now evaluate the performance of the different architectures on the 
	\gls{PDE}-generated datasets described in 
	\cref{sec:math_model,sec:data_generation}. Our goal is not to obtain the lowest 
	possible error on a single benchmark by combining a very large model with massive 
	training data. Instead, we focus on a realistic small-data regime with 
	autoregressive rollouts of $T=100$ time steps and ask the following questions:
	\begin{enumerate}
		\item How do convolutional, operator-based, and transformer-based 
		architectures compare for \emph{in-distribution} rollouts across five 
		qualitatively different \gls{PDE} families?
		
		\item How robust are these models to \emph{out-of-distribution} 
		(\gls{ood}) initial conditions within a fixed \gls{PDE} and parameter 
		regime?
		
		\item How does performance depend on the amount of available data and 
		temporal context (number of simulations, time steps, and input sequence 
		length)?
		
		\item What do the models actually learn, and how do architectural inductive 
		biases manifest in their internal activations?
	\end{enumerate}
	All models are trained with the common setup described in \cref{sec:training}. 
	For each dataset and architecture we train a separate model and generate 
	autoregressive predictions starting from seven ground-truth snapshots; 
	these seven frames form the input sequence, and the model predicts increments 
	for the subsequent $T=100$ time steps.
	
	\subsection{In-distribution prediction}
	We first consider in-distribution prediction, where training and test 
	simulations are drawn from the same distribution of initial conditions for each 
	\gls{PDE}. For each dataset and architecture, we generate rollouts of length 
	$T=100$ and compute the \gls{RMSE} and spectral similarity defined in 
	\cref{sec:metrics}.
	
	\begin{figure*}[htb!]
		\centering
		\begin{subfigure}[t]{0.49\textwidth}
			\centering
			\includegraphics[width=0.995\textwidth]{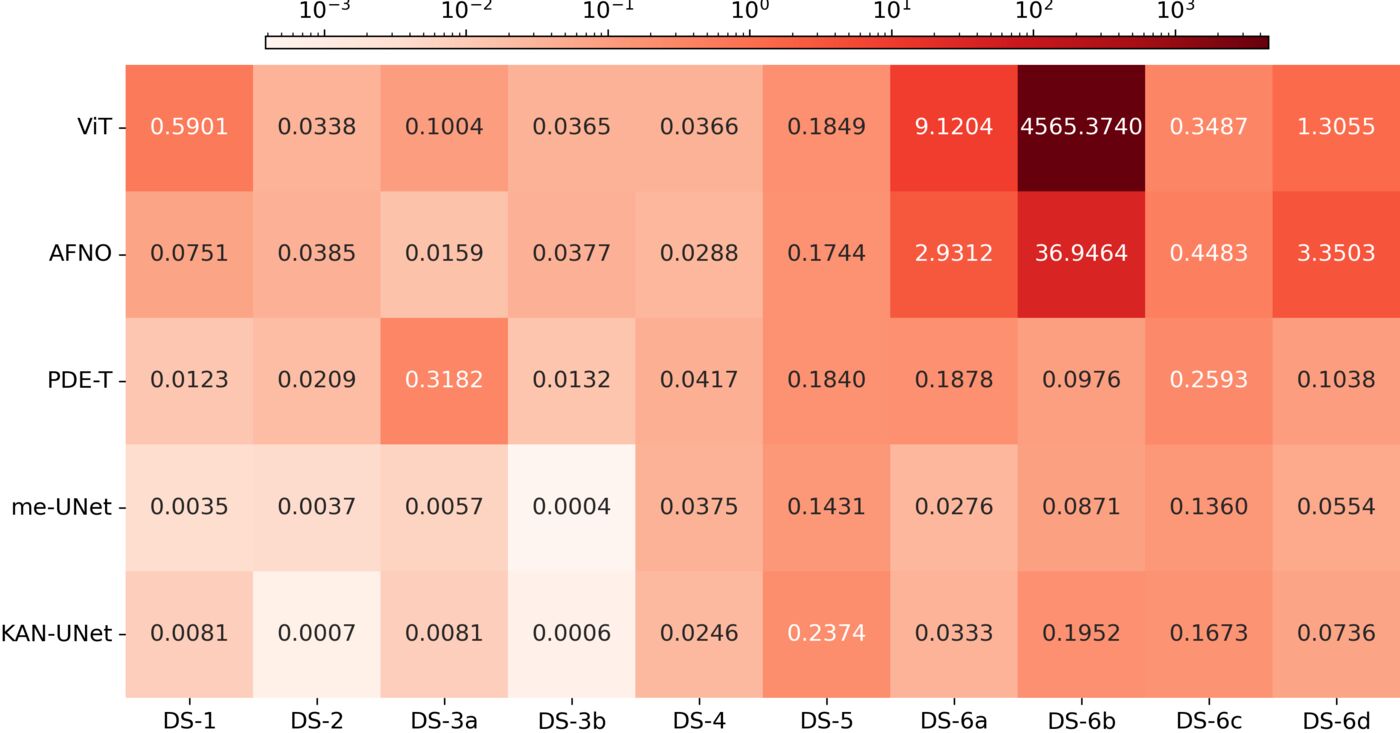}\\
			\caption{}
		\end{subfigure}
		\hfill 
		\begin{subfigure}[t]{0.49\textwidth}
			\centering
			\includegraphics[width=1.\textwidth]{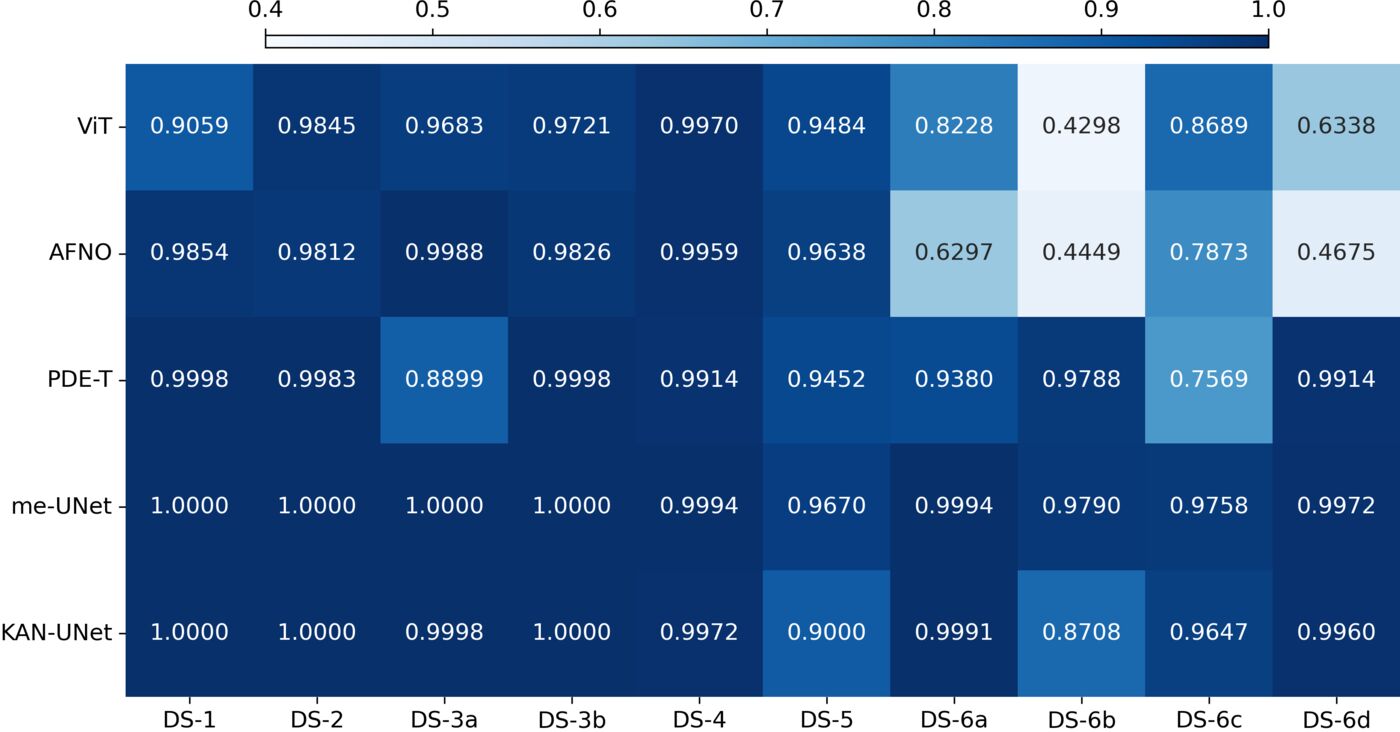}\\
			\caption{}
		\end{subfigure}
		\caption{Comparison of average values over $T=100$ time steps: (a) \gls{RMSE} and (b) cosine-similarity scores of the power spectral density (PSD). Rows correspond to architectures and columns to datasets; colors are shown on a logarithmic scale for \gls{RMSE}.}
		\label{fig:results}
	\end{figure*}
	
	\begin{figure*}[htb!]
		\centering
		\tabskip=0pt
		\valign{#\cr
			\hbox{%
				\begin{subfigure}[b]{.69\textwidth}
					\centering
					\includegraphics[width=1\textwidth]{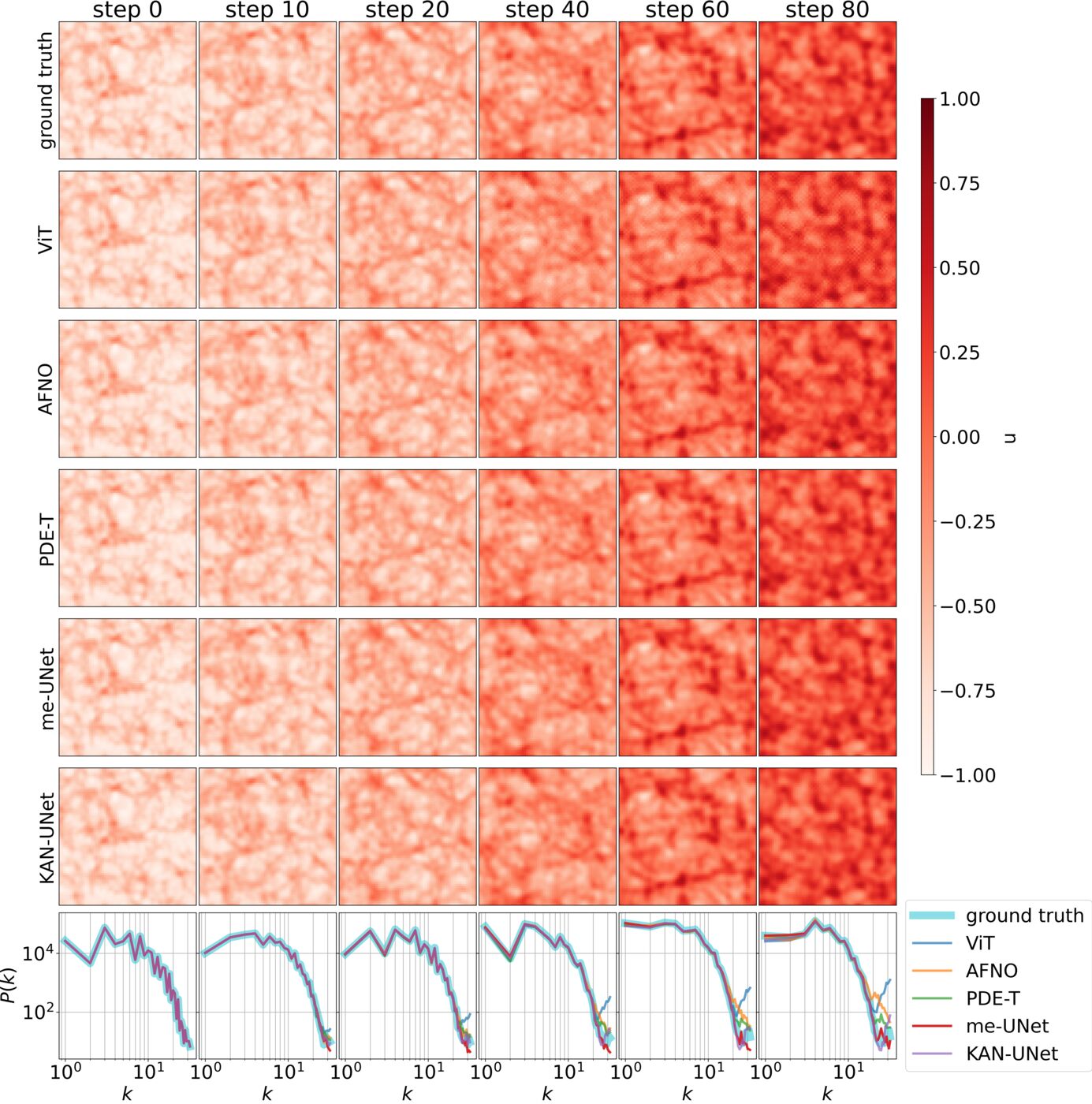}
					\caption{}
				\end{subfigure}%
			}\cr
			\hbox{%
				\begin{subfigure}{.3\textwidth}
					\centering
					\includegraphics[width=0.9\textwidth]{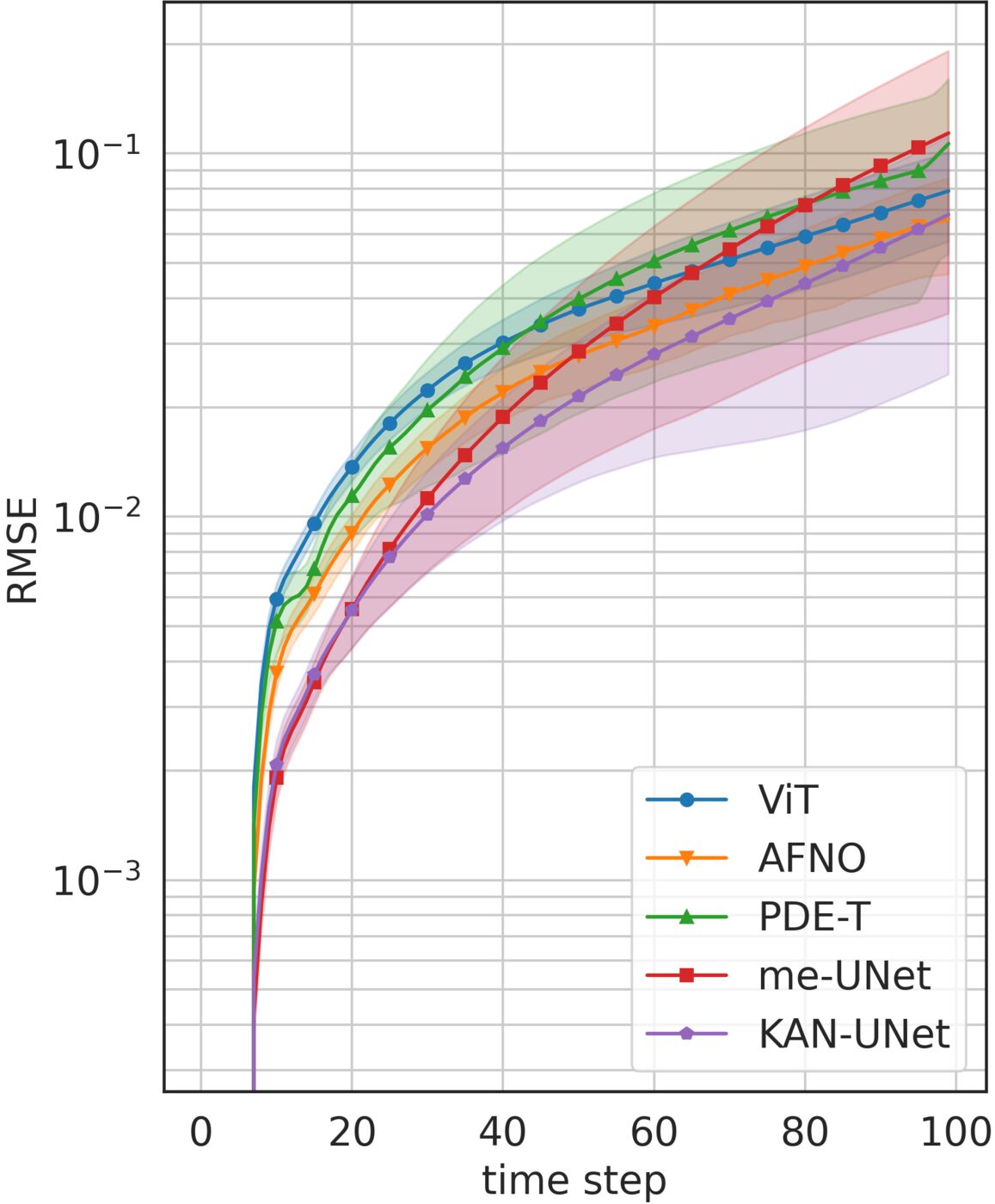}
					\caption{}
				\end{subfigure}%
			}\vfill
			\hbox{%
				\begin{subfigure}{.3\textwidth}
					\centering
					\includegraphics[width=0.9\textwidth]{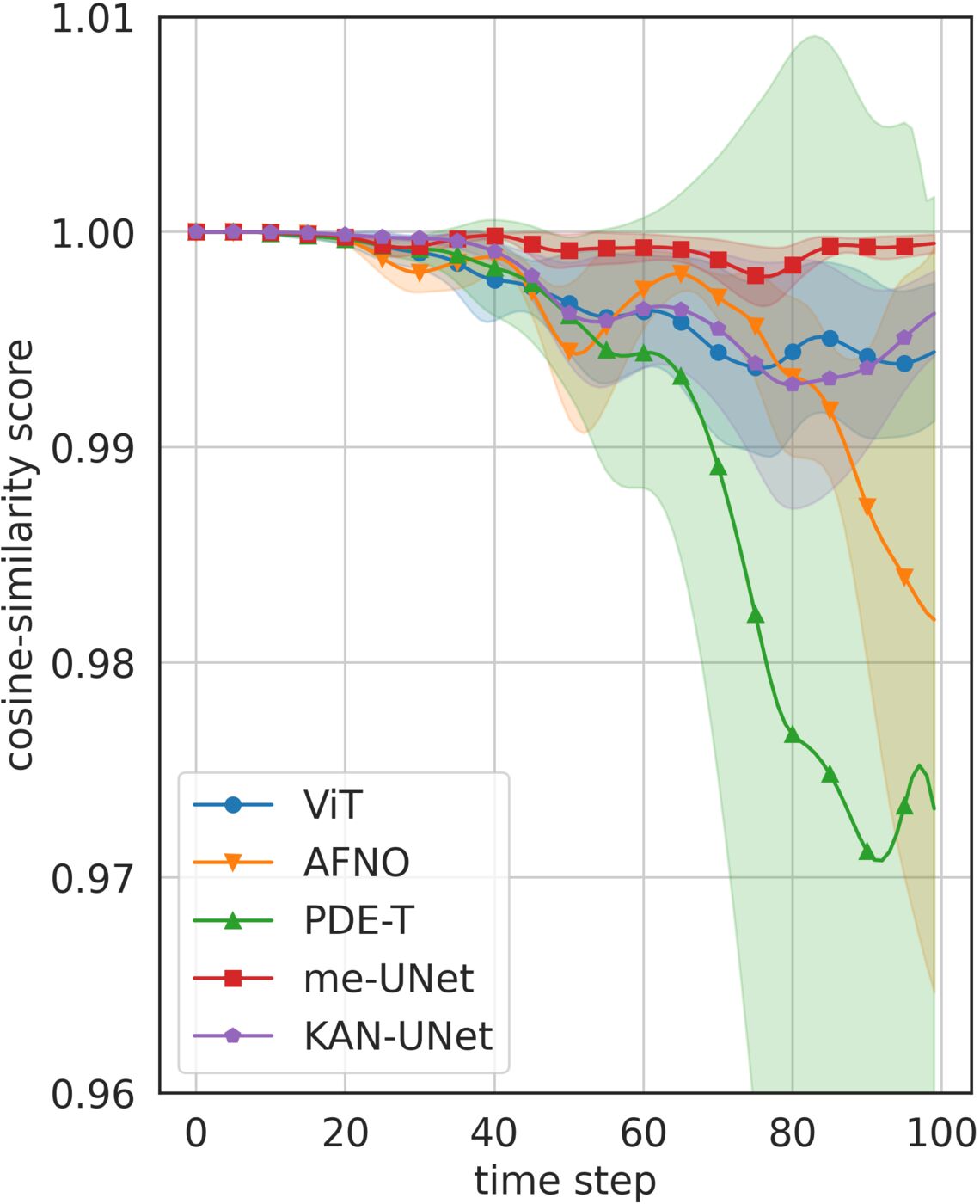}
					\caption{}
				\end{subfigure}%
			}\cr
		}
		\caption{Performance of the different neural network architectures on the \gls{CDD} dataset (DS-4): (a) example autoregressive rollouts; (b) per-time-step \gls{RMSE}; and (c) cosine similarity of the PSD curves between prediction and reference.}
		\label{fig:id_cdd}
	\end{figure*}
	
	\begin{figure*}[htb!]
		\centering
		\tabskip=0pt
		\valign{#\cr
			\hbox{%
				\begin{subfigure}[b]{.69\textwidth}
					\centering
					\includegraphics[width=1\textwidth]{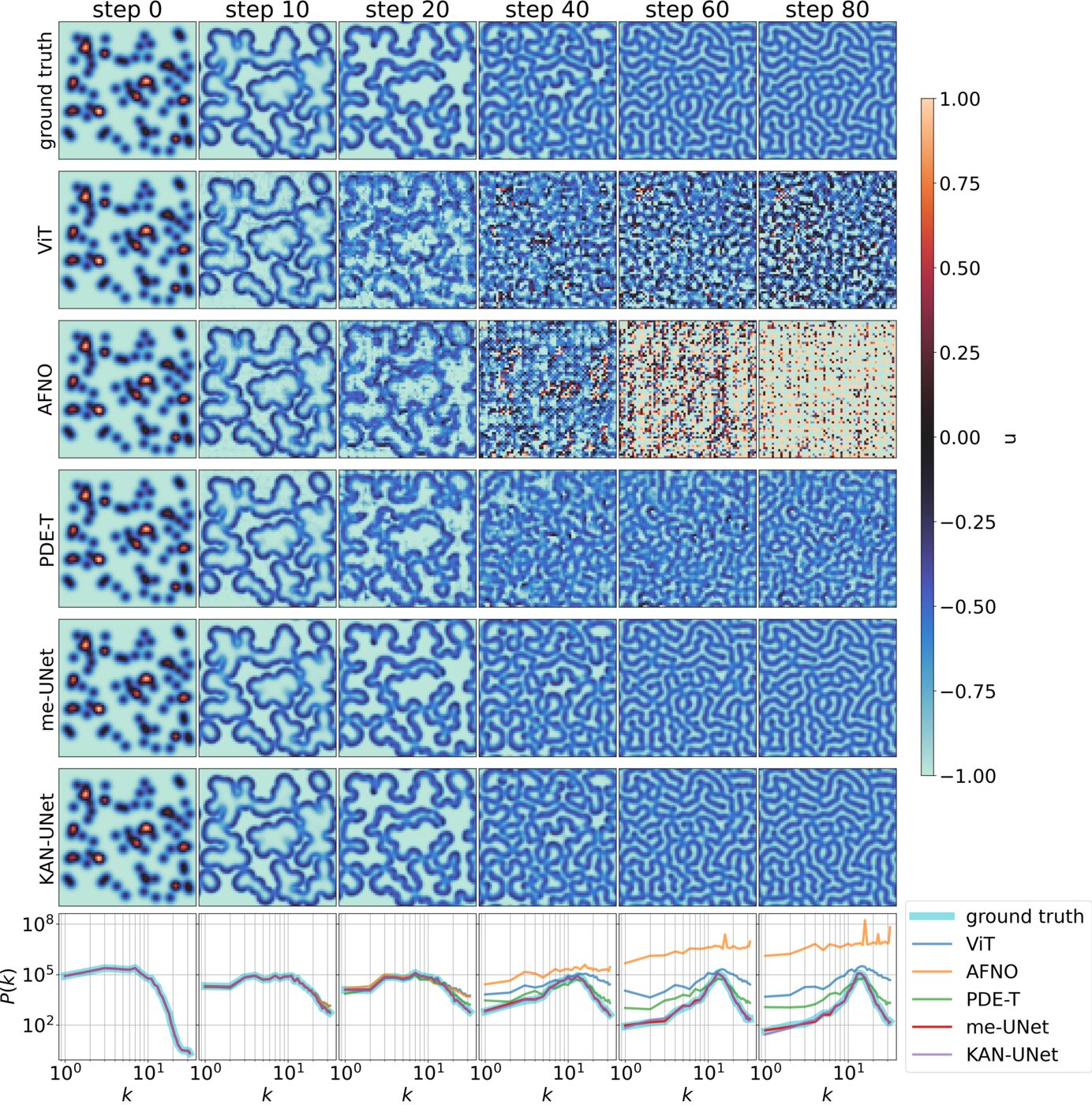}
					\caption{}
				\end{subfigure}%
			}\cr
			\hbox{%
				\begin{subfigure}{.3\textwidth}
					\centering
					\includegraphics[width=0.9\textwidth]{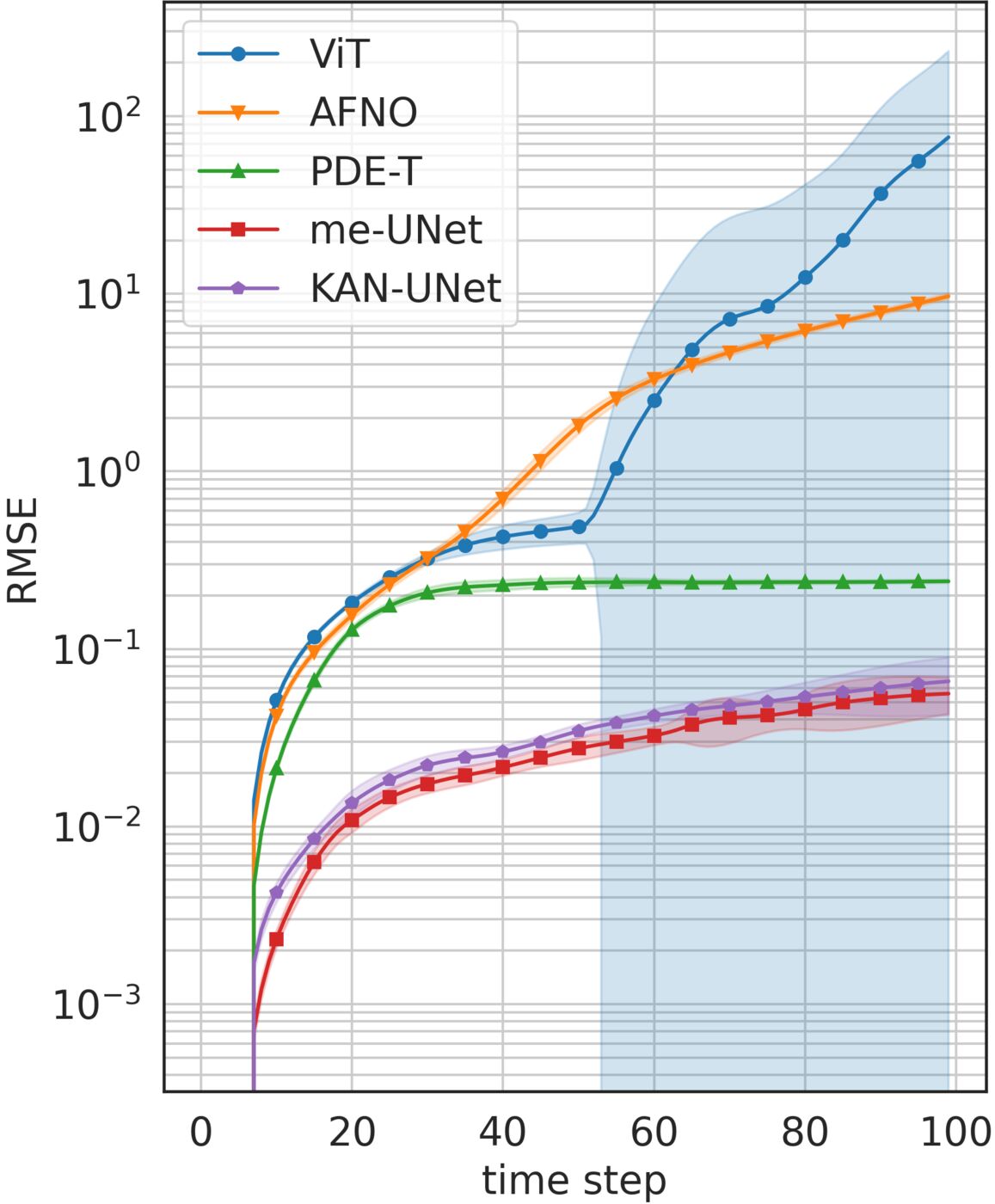}
					\caption{}
				\end{subfigure}%
			}\vfill
			\hbox{%
				\begin{subfigure}{.3\textwidth}
					\centering
					\includegraphics[width=0.9\textwidth]{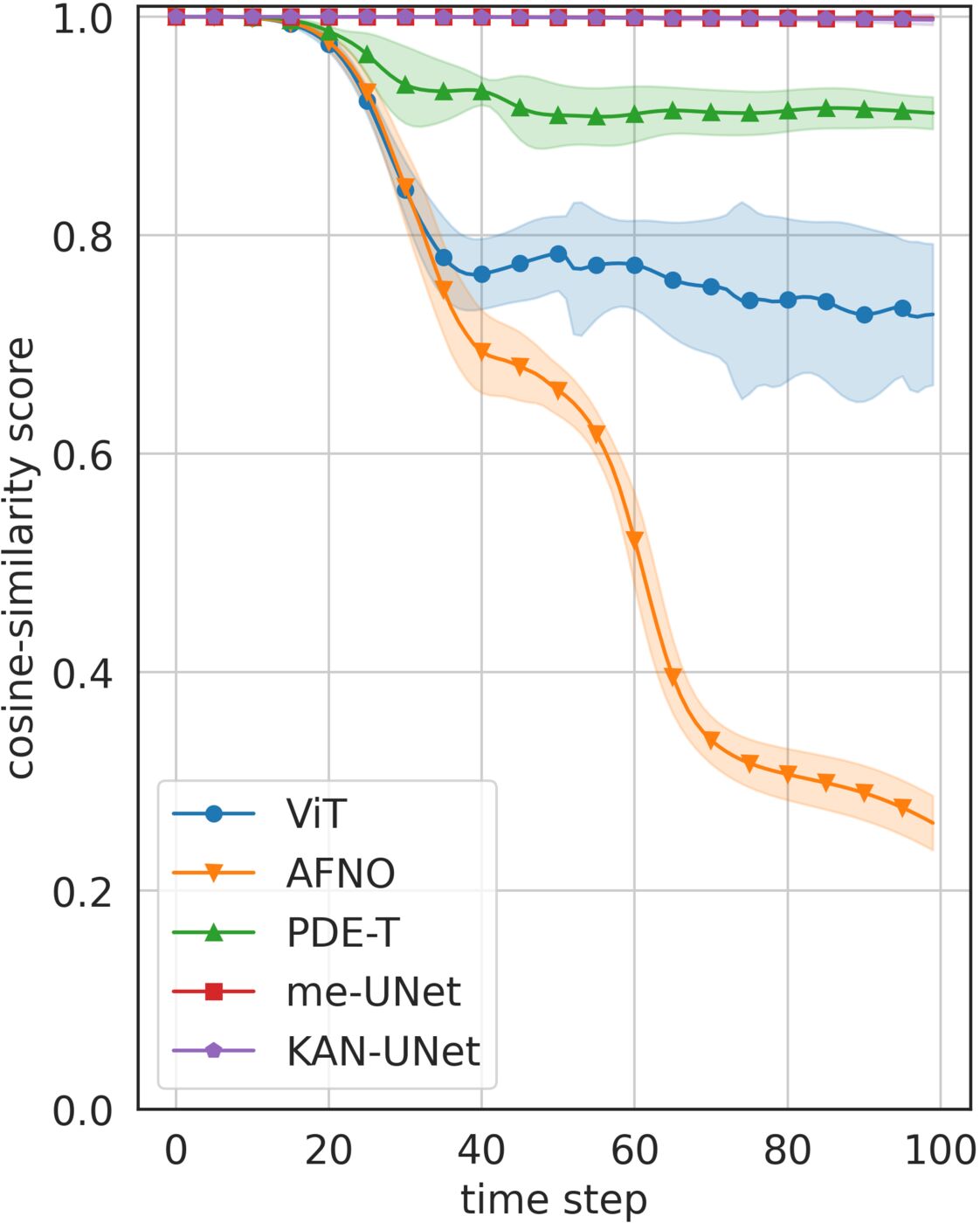}
					\caption{}
				\end{subfigure}%
			}\cr
		}
		\caption{Performance of the different neural network architectures on the Gray--Scott dataset DS-6a: (a) example autoregressive rollouts; (b) per-time-step \gls{RMSE}; and (c) cosine similarity of the PSD curves between prediction and reference.}
		\label{fig:id_gs_maze_d2}
	\end{figure*}
	
	Figure~\ref{fig:results} presents an overview of in-distribution performance 
	across all datasets: panel~(a) shows the average \gls{RMSE} over $T=100$ 
	prediction steps, and panel~(b) shows the corresponding average PSD cosine 
	similarity. Rows correspond to the investigated neural network architectures, 
	and columns to the datasets derived from the mathematical models in 
	\cref{sec:math_model}. Lower \gls{RMSE} indicates better field-space accuracy, 
	while cosine similarity values closer to $1$ indicate a closer match of the 
	spatial frequency content.
	
	Overall, me-UNet achieves the lowest \gls{RMSE} on almost all datasets. In 
	particular, its average \gls{RMSE} on DS-1, DS-3a, DS-3b, DS-5, DS-6a, DS-6b, 
	DS-6c, and DS-6d is $0.0035 \pm 0.0002$, $0.0057 \pm 0.0007$, 
	$0.0004 \pm 0.00005$, $0.1431 \pm 0.02$, $0.0276 \pm 0.005$, 
	$0.0871 \pm 0.03$, $0.1360 \pm 0.08$, and $0.0554 \pm 0.006$, respectively. 
	The corresponding predicted snapshots show good agreement with the ground truth 
	(see the me-UNet rows in \cref{fig:id_diff}a, \cref{fig:id_cdd_adv}a, 
	\cref{fig:id_cdd_diff}a, \cref{fig:id_cfd}a, \cref{fig:id_gs_maze_d2}a, 
	\cref{fig:id_gs_alpha_v2}a, \cref{fig:id_gs_worms_mu}a, and 
	\cref{fig:id_gs_bubbles}a). KAN-UNet achieves the lowest \gls{RMSE} on DS-2 and 
	DS-4 with $0.0007 \pm 0.00002$ and $0.0246 \pm 0.0125$, respectively (see the 
	KAN-UNet rows in \cref{fig:id_diff}a and \cref{fig:id_cdd}a).
	
	In contrast, ViT attains the highest \gls{RMSE} on several datasets 
	(e.g., $0.5901 \pm 0.2$ on DS-1, $9.1204 \pm 18.96$ on DS-6a, and 
	$4565.3740 \pm 1860.8349$ on DS-6b), which is reflected in visibly degraded 
	predictions (see the first rows of \cref{fig:id_adv}a, \cref{fig:id_gs_maze_d2}a, 
	and \cref{fig:id_gs_alpha_v2}a). PDE-Transformer performs worst on DS-3a 
	($0.3182 \pm 0.06$, see \cref{fig:id_cdd_adv}a), while AFNO shows the largest 
	errors on DS-6c ($0.4483 \pm 0.1384$) and DS-6d ($3.3502 \pm 2.2589$). 
	
	Spectrally, me-UNet also consistently achieves the highest PSD cosine similarity, 
	with values close to $1.0$ across all datasets. KAN-UNet is typically the 
	second-best model, with cosine similarity near $1.0$ except on DS-5, where it 
	attains $0.9 \pm 0.07$. ViT and AFNO exhibit the lowest spectral similarity on 
	the most challenging Gray--Scott datasets (e.g., $0.4298 \pm 0.02$ and 
	$0.4449 \pm 0.02$ on DS-6b, and $0.4675 \pm 0.08$ on DS-6d), indicating 
	substantial distortion of the spatial frequency content over long rollouts.
	
	Figures~\ref{fig:id_cdd} and \ref{fig:id_gs_maze_d2} show representative 
	in-distribution rollouts for the \gls{CDD} dataset (DS-4) and the Gray--Scott 
	dataset DS-6a. Each figure contains snapshot sequences, per-time-step 
	\gls{RMSE}, and PSD cosine similarity for all architectures. For \gls{CDD}, 
	me-UNet preserves both the large-scale microstructure and the fine dislocation 
	patterns over $T=100$ time steps, with \gls{RMSE} remaining low and PSD 
	similarity close to~1. Some baselines show accumulation of high-frequency 
	artifacts and microstructure degradation over time. For Gray--Scott, me-UNet 
	again produces stable pattern evolution with accurate spectral content, whereas 
	several heavier architectures exhibit pattern drift and loss of small-scale 
	structures.
	
	Additional in-distribution rollouts for all datasets are provided in Appendix~C, 
	together with the corresponding PSD curves and cosine similarity plots.

	Training and validation losses converge smoothly for me‑UNet under the common protocol; ViT and AFNO show less stable training dynamics (see Fig.~\ref{fig:train_val_loss}).

	\subsection{\gls{ood} prediction}
	We now investigate \gls{ood} prediction performance on the \gls{CDD} dataset 
	(DS-4) and the Gray--Scott dataset DS-6a, for which the corresponding 
	in-distribution results are shown in \cref{fig:id_cdd} and 
	\cref{fig:id_gs_maze_d2}. In both cases, the underlying \gls{PDE} and 
	parameters are kept fixed, but the initial conditions are qualitatively very 
	different from those seen during training.
	
\begin{figure*}[htb!]
		\centering
		\begin{subfigure}[t]{0.49\textwidth}
			\centering
			\includegraphics[width=0.995\textwidth]{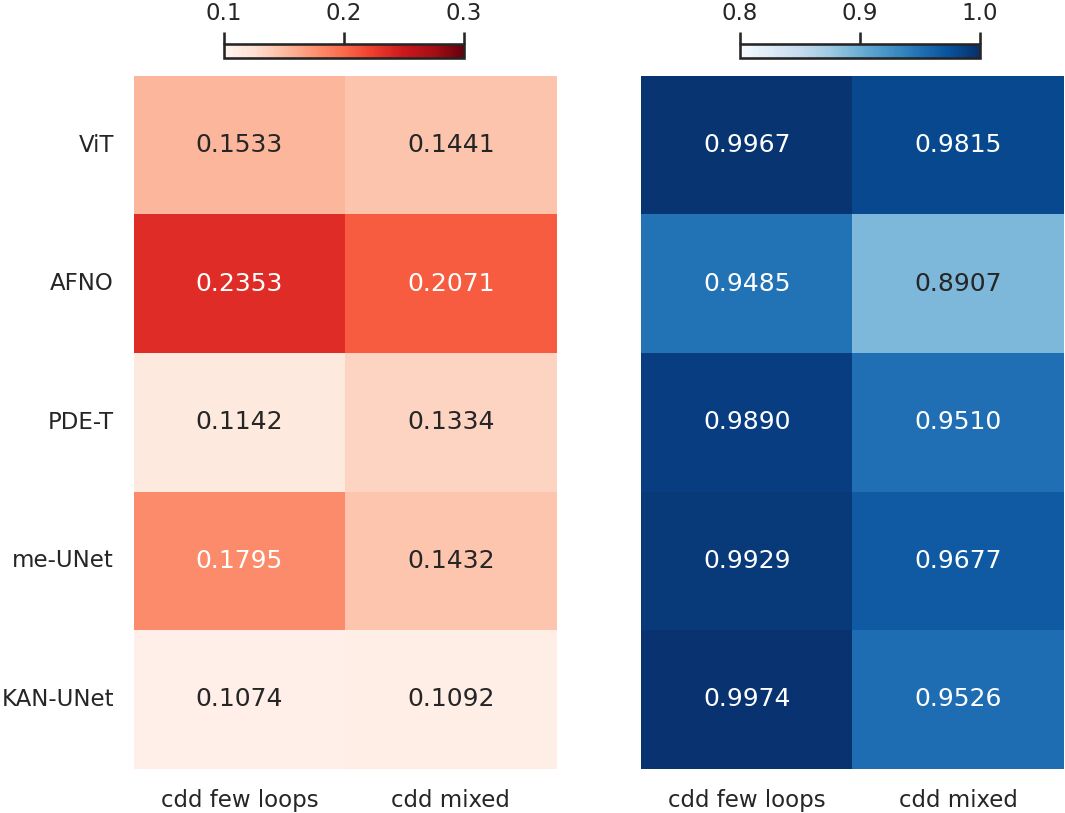}\\
			\caption{}
		\end{subfigure}
		\hfill 
		\begin{subfigure}[t]{0.49\textwidth}
			\centering
			\includegraphics[width=1.\textwidth]{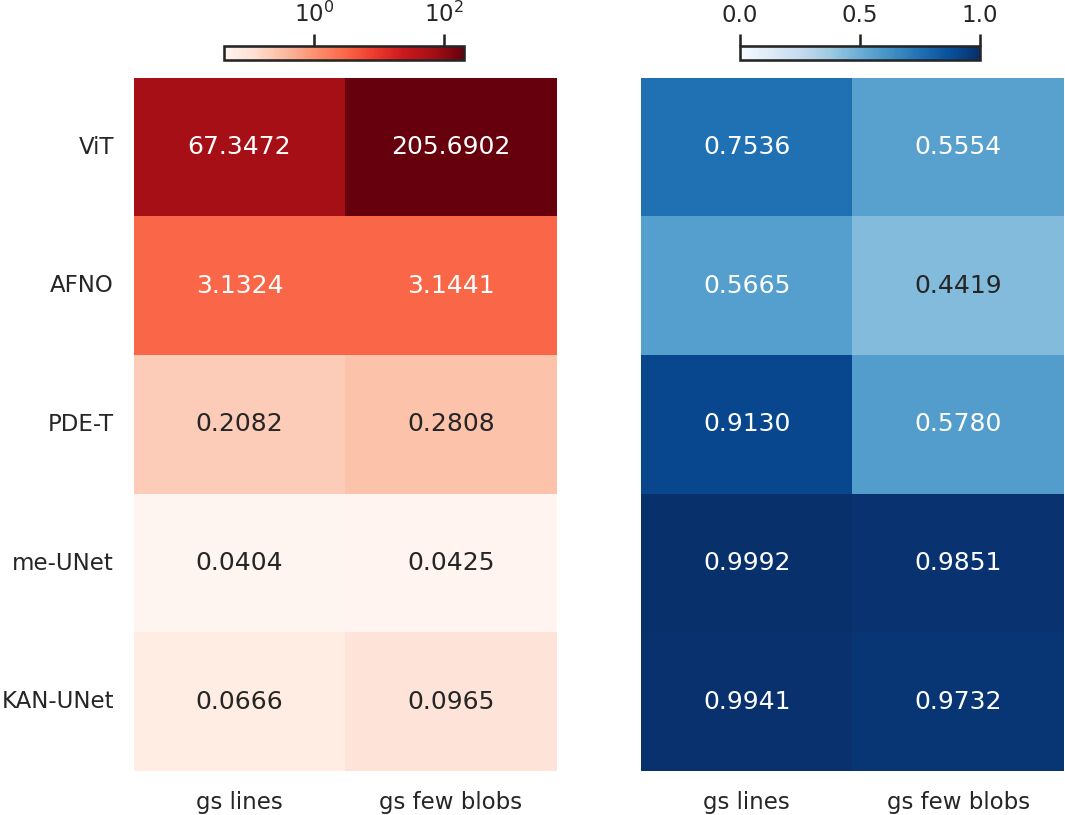}\\
			\caption{}
		\end{subfigure}
		\caption{Comparison of average \gls{RMSE} (red color scheme) and cosine-similarity score (blue color scheme) for \gls{ood} prediction on (a) the \gls{CDD} dataset (DS-4) and (b) the Gray--Scott dataset DS-6a.}
		\label{fig:ood_results}
	\end{figure*}

\begin{figure*}[h!tb]
		\centering
		\begin{subfigure}[t]{0.49\textwidth}
			\centering
			\includegraphics[width=1\textwidth]{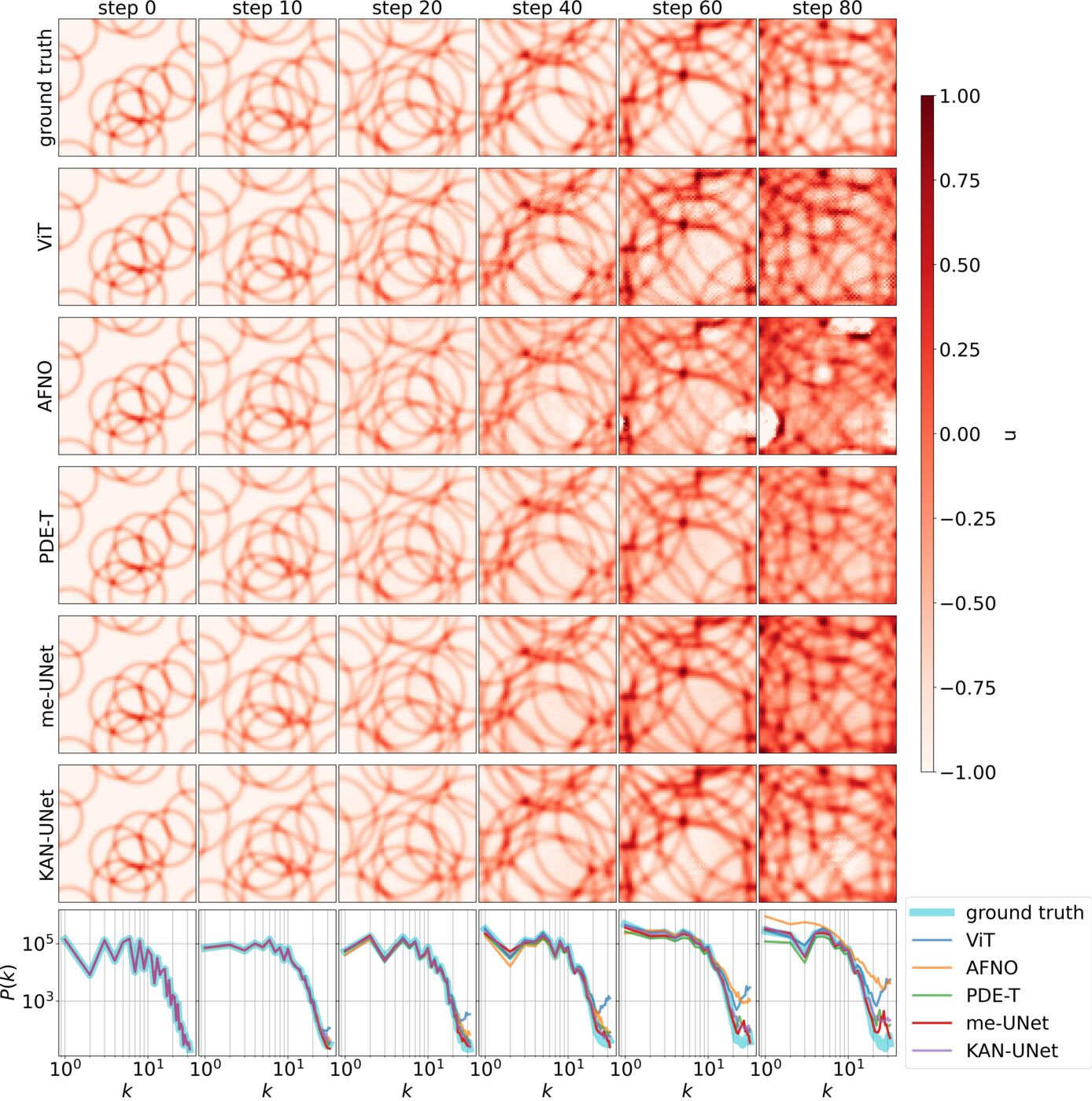}\\
			\caption{}
		\end{subfigure}
		\hfill 
		\begin{subfigure}[t]{0.49\textwidth}
			\centering
			\includegraphics[width=1.\textwidth]{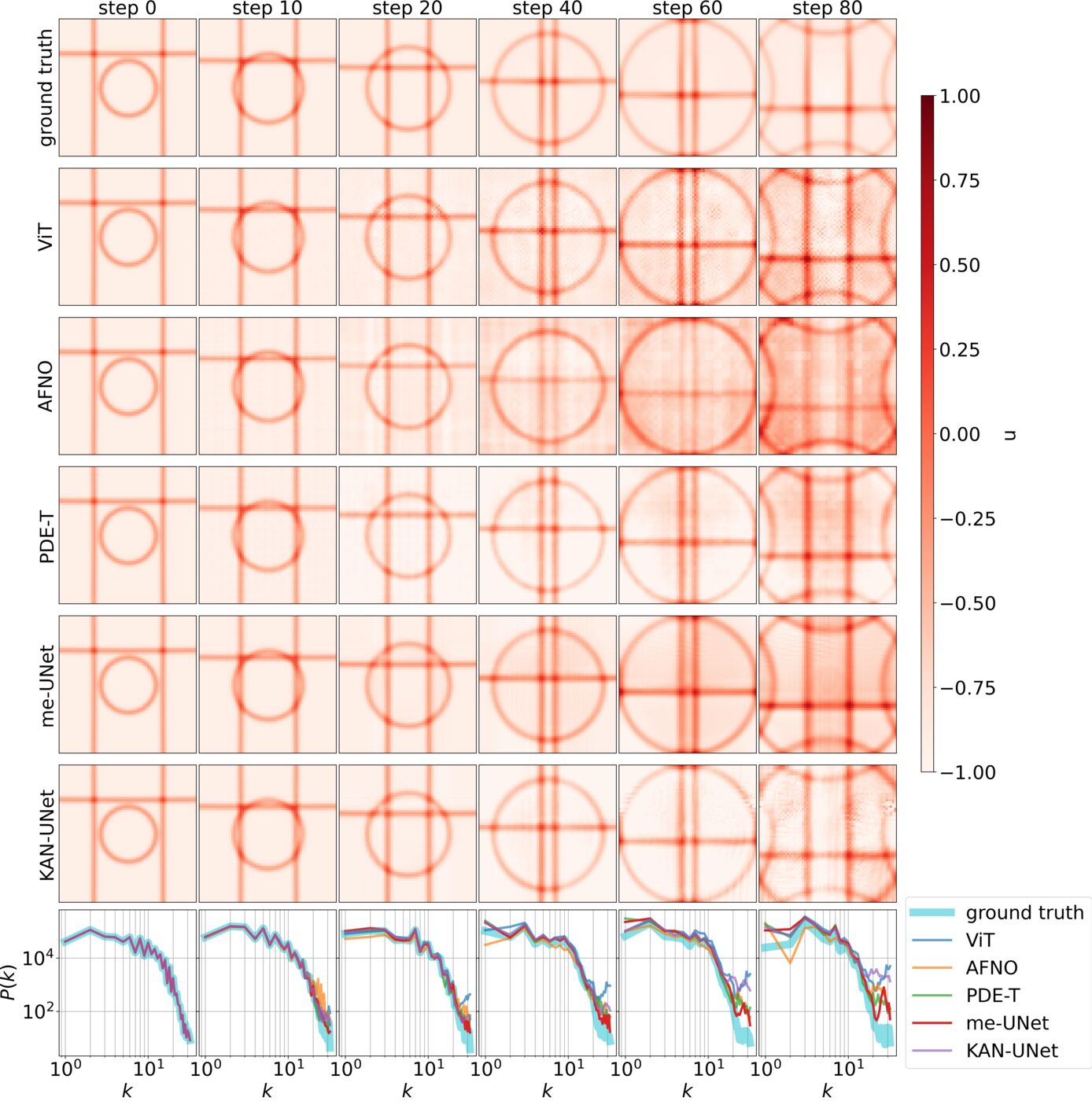}\\
			\caption{}
		\end{subfigure}
		\caption{Example \gls{ood} prediction results of the different neural networks on the \gls{CDD} model: (a) initial condition with several loops; (b) initial condition with three lines and one loop. The training data was visually very different but still contained the same underlying dynamical behavior that is also observed here.}
		\label{fig:ood_cdd}
	\end{figure*}

	For \gls{CDD}, we train on simulations with a very large number of
	superimposed blob- and loop-like initial patterns, resulting in images where 
	the individual blobs or loops can no longer be identified. Testing is then 
	performed on configurations with a very small number of loops or with mixed 
	line and loop arrangements, which are visually very different from the 
	training data. Figures~\ref{fig:ood_cdd}a and~\ref{fig:ood_cdd}b illustrate two 
	such \gls{ood} cases. The main physical phenomenon---expansion of dislocation 
	loops throughout the domain---is captured by most trained networks, and the 
	predicted patterns remain close to the reference for me-UNet and KAN-UNet. The 
	PSD curves in the last rows of \cref{fig:ood_cdd} show a good match across a 
	wide range of frequencies, with only small deviations at high frequencies for 
	the best-performing models. AFNO, however, exhibits unstable behavior as time
	progresses (see the third rows of \cref{fig:ood_cdd}), leading to the highest
	\gls{RMSE} values of $0.2353$ and $0.2071$ in the two \gls{ood} tests, as well
	as the lowest cosine similarities ($0.9485$ and $0.8907$, see
	\cref{fig:ood_results}a and \cref{fig:ood_rmse_cos_sim}a). The fact that 
	me-UNet can extrapolate from densely populated microstructures to sparse, 
	geometrically simple initial conditions strongly suggests that it has learned 
	the underlying dislocation dynamics rather than merely memorizing typical 
	training configurations.
	
\begin{figure*}[h!tb]
		\centering
		\begin{subfigure}[t]{0.49\textwidth}
			\centering
			\includegraphics[width=1.\textwidth]{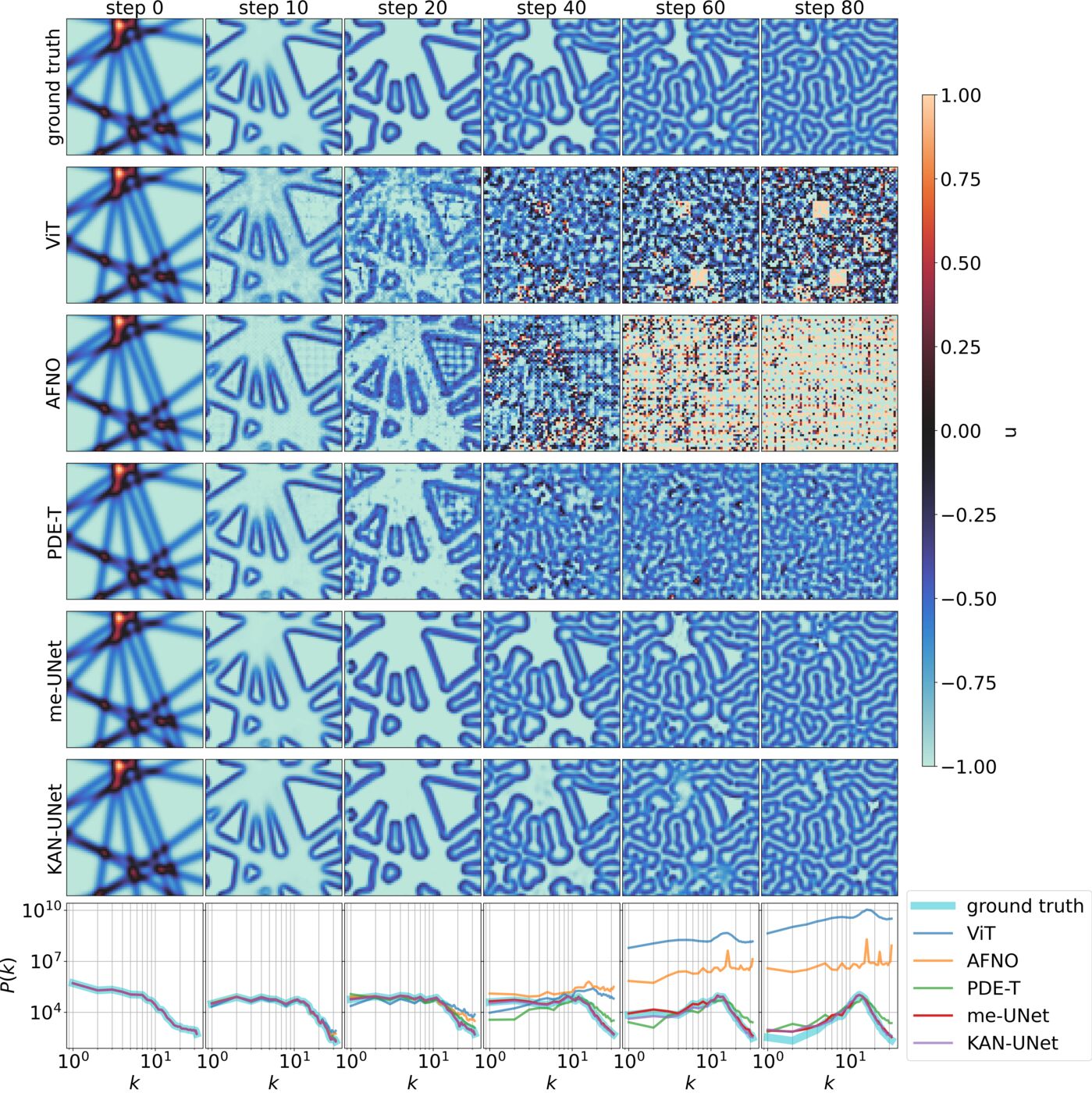}\\
			\caption{}
		\end{subfigure}
		\hfill 
		\begin{subfigure}[t]{0.49\textwidth}
			\centering
			\includegraphics[width=1.\textwidth]{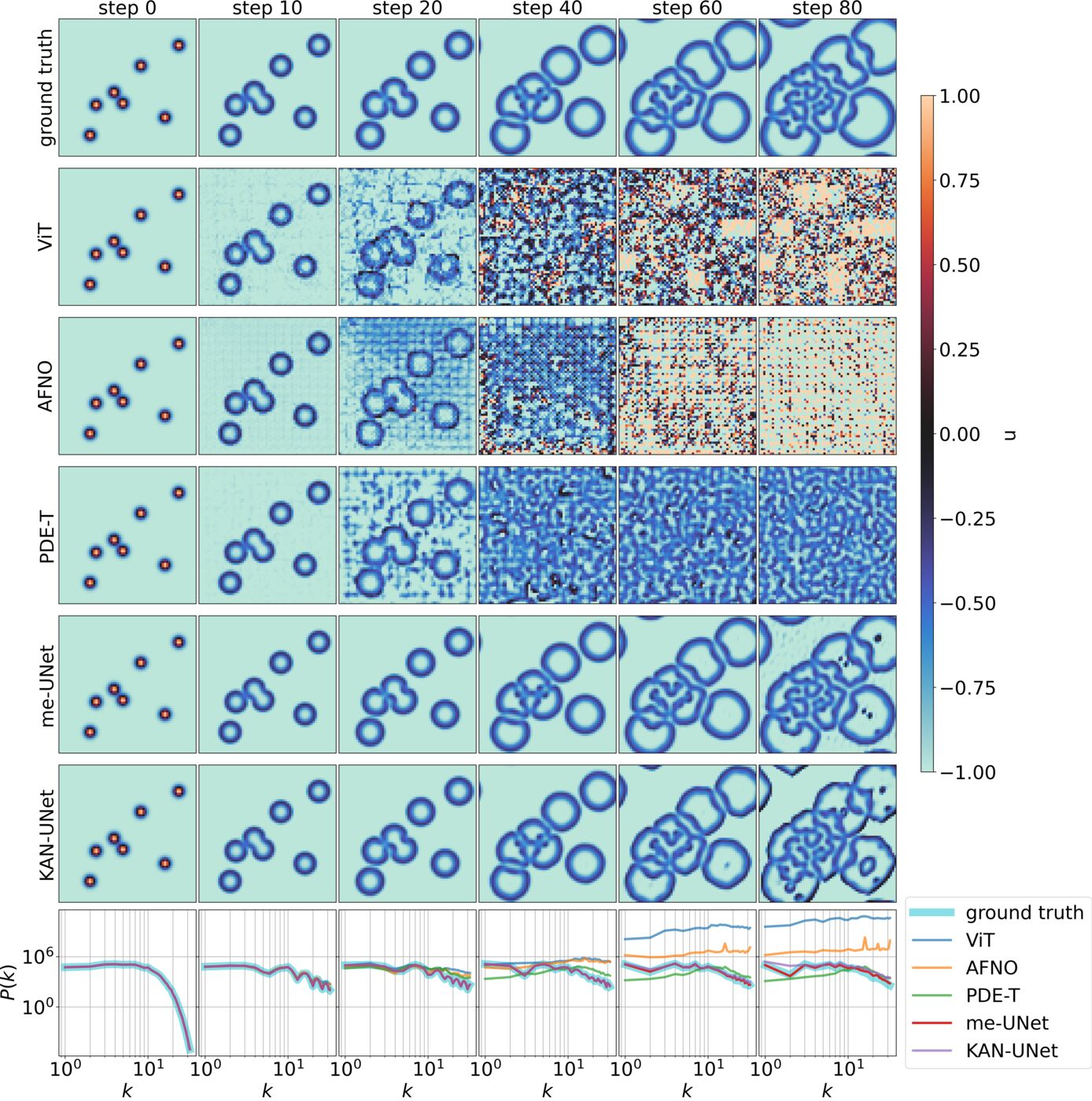}\\
			\caption{}
		\end{subfigure}
		\caption{Example \gls{ood} prediction results of the different neural networks on the Gray--Scott model (DS-6a): (a) line-like initial condition; (b) blob-like initial condition.}
		\label{fig:ood_gs_maze_d2}
	\end{figure*}

	For the Gray--Scott system, we train on randomly perturbed initial states and 
	test on structured line and blob initial conditions that differ strongly from 
	the training distribution. Figures~\ref{fig:ood_gs_maze_d2}a and 
	\ref{fig:ood_gs_maze_d2}b show \gls{ood} rollouts for line and blob initial 
	patterns, respectively (compare the first columns of \cref{fig:id_gs_maze_d2} 
	and \cref{fig:ood_gs_maze_d2}). Transformer-based models such as ViT, AFNO, 
	and PDE-Transformer already struggle to train in-distribution on this dataset 
	and therefore perform poorly on \gls{ood} initial conditions; their predictions 
	are very noisy and quickly diverge from the ground truth (see the second, 
	third, and fourth columns of \cref{fig:ood_gs_maze_d2}).
	
	In contrast, me-UNet and KAN-UNet produce qualitatively correct \gls{ood} 
	dynamics, capturing the reaction--diffusion behavior and generating patterns 
	that remain similar to the reference (fifth and sixth columns of 
	\cref{fig:ood_gs_maze_d2}a). For me-UNet, the \gls{RMSE} is lowest among all 
	models ($0.0404$ and $0.0425$ for the two \gls{ood} cases), and the cosine 
	similarity is highest ($0.9992$ and $0.9851$, see \cref{fig:ood_results}b and 
	\cref{fig:ood_rmse_cos_sim}b). Some noise appears at later time steps due to 
	accumulated prediction errors, but the overall patterns remain stable. ViT 
	attains the highest \gls{RMSE} ($67.3472$ and $205.6902$), and AFNO reaches the 
	lowest cosine similarities ($0.5665$ and $0.4419$), indicating severe breakdown 
	of \gls{ood} generalization for these heavy architectures.
	
	An overview of \gls{ood} performance across architectures is given in 
	\cref{fig:ood_results}, which summarizes average \gls{RMSE} and PSD cosine 
	similarity for the \gls{CDD} and Gray--Scott \gls{ood} tests. me-UNet exhibits 
	the most consistent behavior, with relatively small degradation from 
	in-distribution to \gls{ood} initial conditions, while several baselines show 
	substantial increases in error or spectral mismatch on at least one of the two 
	systems. Physics-aware metrics, such as energy, mass, and total dislocation 
	density, further corroborate these trends: \cref{fig:physics_diagnostic} shows that me-UNet preserves these quantities more accurately than 
	the baselines, especially under \gls{ood} conditions.
	
	\subsection{Data efficiency}
	Here we study data efficiency on the \gls{CDD} dataset, which serves as a 
	representative benchmark for evaluating how the architectures behave under data 
	limitations. We do not repeat these ablations on all other datasets, but we 
	expect qualitatively similar trends to hold.
	
	We vary three factors: (i) the number of time steps per simulation used during 
	training, (ii) the number of training simulations, and (iii) the input 
	sequence length $L$ (number of past time steps provided as context). For each 
	setting we retrain all architectures with the same protocol as in 
	\cref{sec:training} and evaluate in-distribution rollouts.
	
	\begin{figure*}[h!tb]
		\centering
		\begin{subfigure}{0.33\textwidth}
			\centering
			\includegraphics[width=0.9\textwidth]{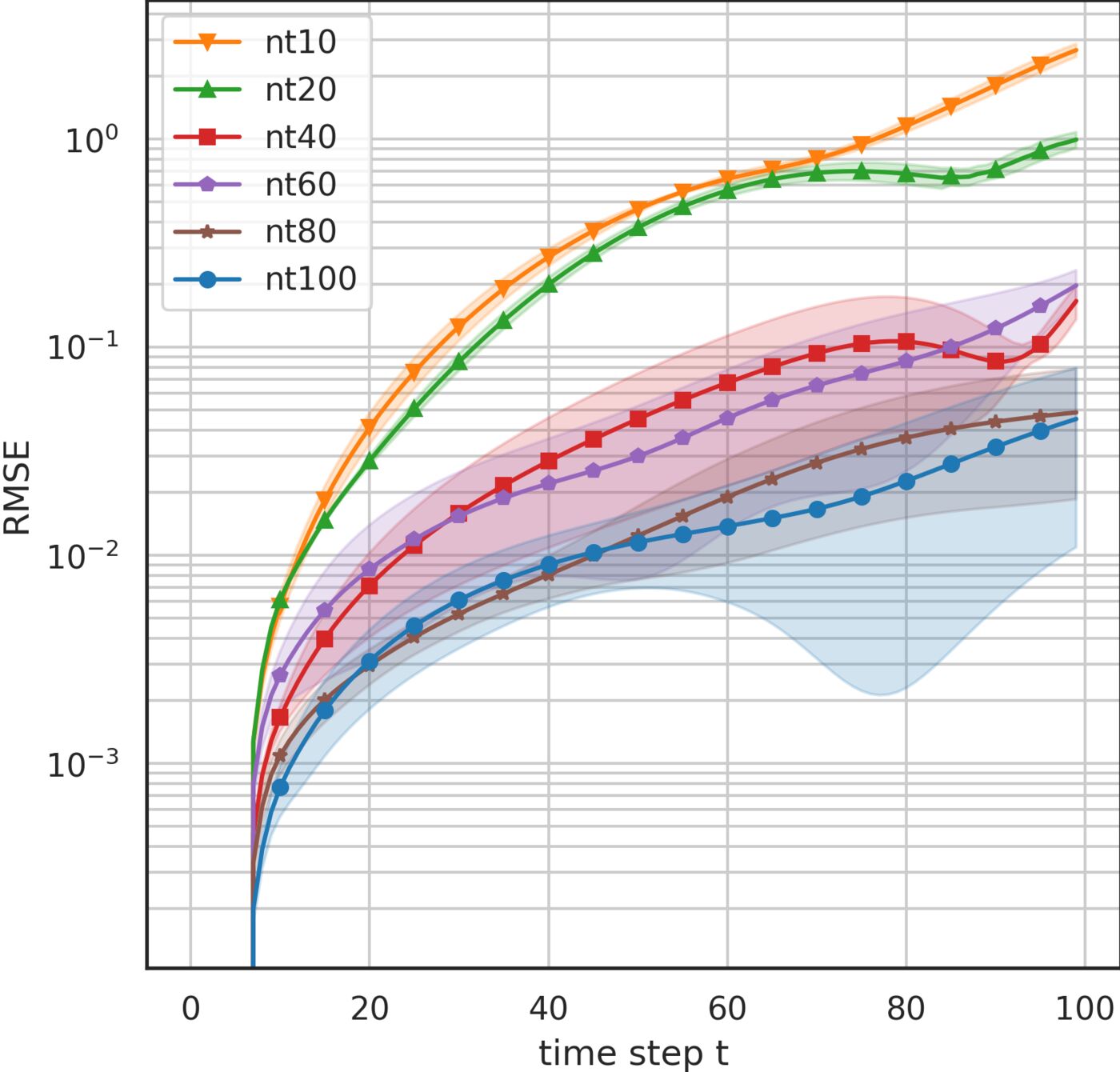}\\
			\bigskip
			\caption{}
		\end{subfigure}
		\centering
		\begin{subfigure}{0.33\textwidth}
			\centering
			\includegraphics[width=0.9\textwidth]{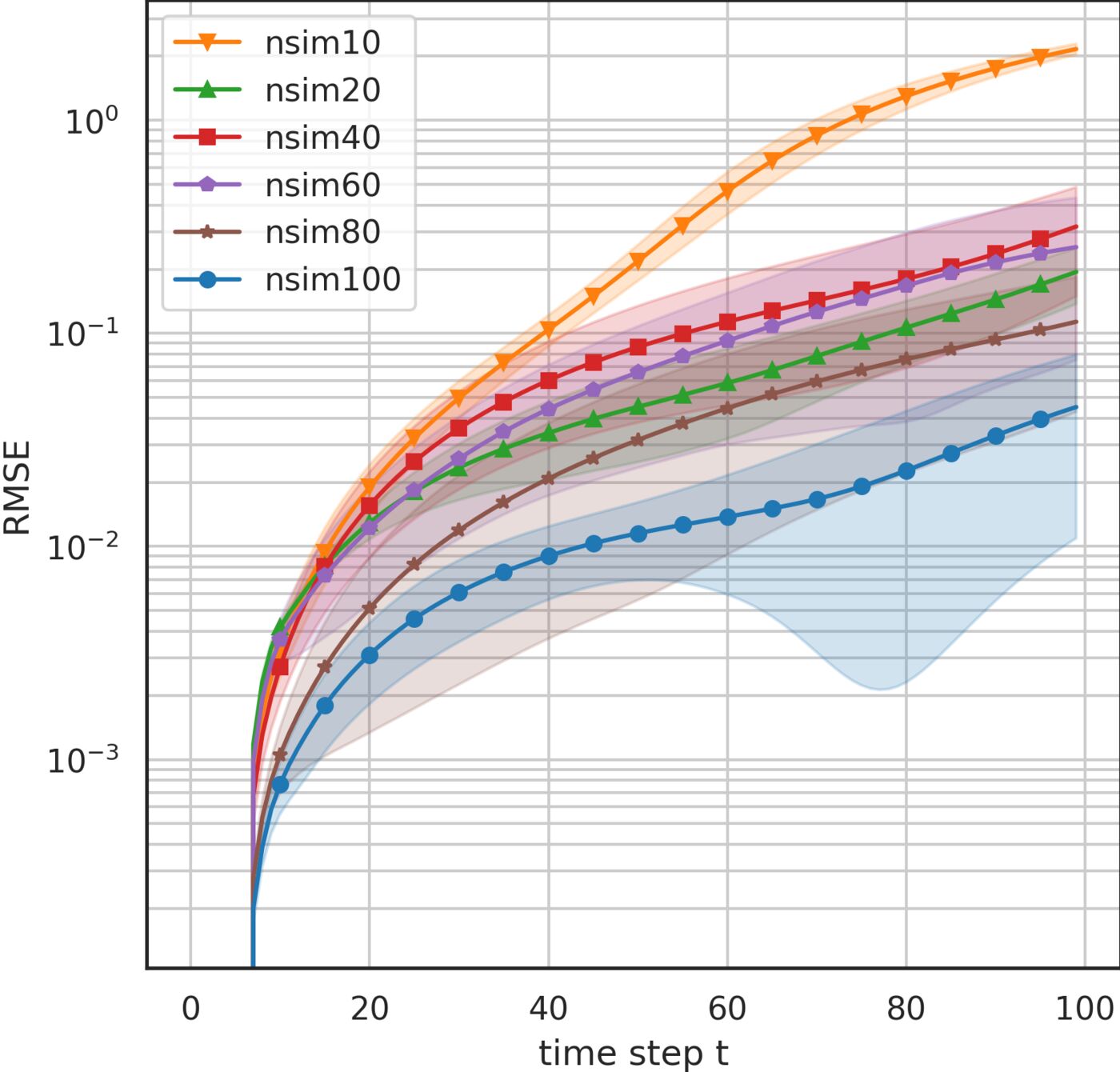}\\
			\bigskip
			\caption{}
		\end{subfigure}
		\centering
		\begin{subfigure}{0.33\textwidth}
			\centering
			\includegraphics[width=0.9\textwidth]{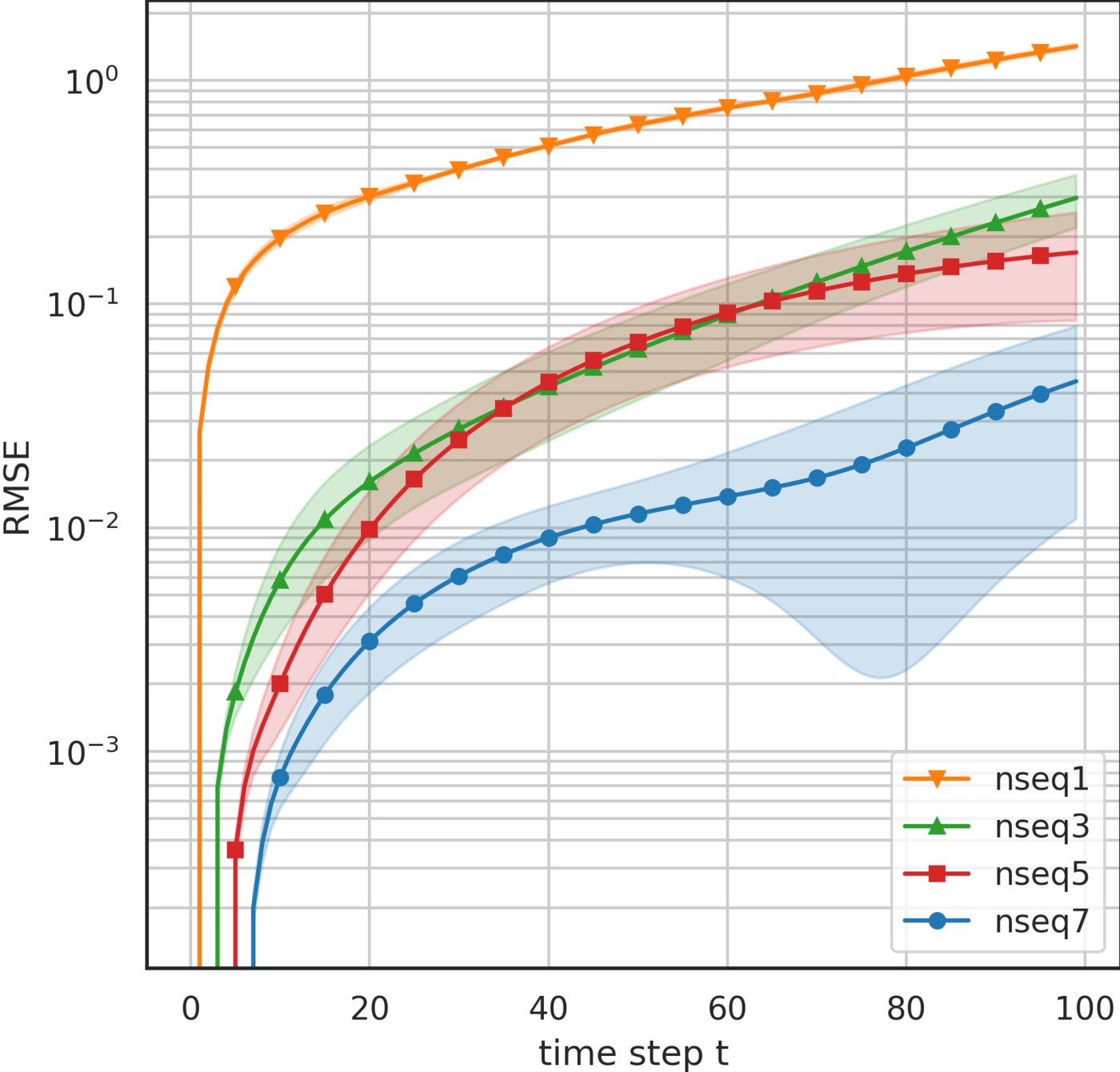}\\
			\bigskip
			\caption{}
		\end{subfigure}
		\caption{Data-efficiency study on the \gls{CDD} dataset (DS-4): \gls{RMSE} over 100-step rollouts as a function of (a) the number of time steps per simulation used during training, (b) the number of training simulations, and (c) the input sequence length $L$ (number of past frames used as context).}
		\label{fig:cdd_rmse_nt_nsim_nseq}
	\end{figure*}
	
	Figure~\ref{fig:cdd_rmse_nt_nsim_nseq} summarizes the \gls{RMSE} obtained under 
	these variations. In panel~(a), labels $\mathrm{nt10}$, $\mathrm{nt20}$, 
	$\mathrm{nt40}$, $\mathrm{nt60}$, $\mathrm{nt80}$, and $\mathrm{nt100}$ 
	indicate the number of time steps per simulation used to construct training 
	samples (with the number of simulations fixed at 100 and the input sequence 
	length at $L=7$). We observe that training with $40$ time steps per simulation 
	is already sufficient for me-UNet and the better-performing baselines; using 
	more time steps yields only marginal improvements in \gls{RMSE}.
	
	In panel~(b), labels $\mathrm{nsim10}$, $\mathrm{nsim20}$, $\mathrm{nsim40}$, 
	$\mathrm{nsim60}$, $\mathrm{nsim80}$, and $\mathrm{nsim100}$ indicate the 
	number of simulations used for training (again with $L=7$ and $T=100$ time 
	steps available). Here, me-UNet reaches a low-error regime with as few as $20$ 
	training simulations, with \gls{RMSE} already close to the values obtained with 
	$40$, $60$, or $80$ simulations. In contrast, the transformer- and 
	operator-based models benefit more noticeably from additional simulations, 
	indicating that their data requirements are higher in this setting.
	
	Finally, panel~(c) varies the input sequence length $L \in \{1, 3, 5, 7\}$ 
	while keeping the number of simulations and time steps fixed at 100. We find 
	that using $L=5$ or $L=7$ input frames yields substantially lower \gls{RMSE} 
	than using a single-frame context ($L=1$), confirming that limited temporal 
	history is important for accurate next-step prediction. Performance saturates 
	between $L=5$ and $L=7$, consistent with the choice of time step in the 
	simulations: the dynamics change appreciably over a window of about six 
	intervals, and longer contexts bring limited additional benefit.
	
	Overall, these data-efficiency experiments show that, in the small-data regime, 
	the convolutional inductive biases of me-UNet allow it to make more effective 
	use of limited training data than the heavier baselines. They also provide 
	practical guidance for selecting the number of simulations, time steps, and 
	input sequence length in future studies on similar systems.
	
	\subsection{Studying the network behavior}
	In this section we analyze the internal behavior of me-UNet during training on 
	the different mathematical models. We use Grad-CAM to visualize which spatial 
	regions each block of the network attends to over the course of training. The 
	resulting visualizations are shown in 
	\cref{fig:grad_cam_adv,fig:grad_cam_diff,fig:grad_cam_cdd_adv,fig:grad_cam_cdd_diff,fig:grad_cam_cdd,fig:grad_cam_cfd,fig:grad_cam_gs_maze_d2,fig:grad_cam_gs_alpha_v2,fig:grad_cam_gs_worms_mu,fig:grad_cam_gs_bubbles}. 
	The horizontal axis in these figures corresponds to training epochs, while the 
	vertical axis indexes the U-Net blocks.
	
	We denote the five encoder (downsampling) blocks by $\mathrm{d0}$, $\mathrm{d1}$, 
	$\mathrm{d2}$, $\mathrm{d3}$, and $\mathrm{d4}$, and the five decoder 
	(upsampling) blocks by $\mathrm{u0}$, $\mathrm{u1}$, $\mathrm{u2}$, 
	$\mathrm{u3}$, and $\mathrm{u4}$. These blocks are annotated in 
	\cref{fig:me_UNet}. The spatial resolutions of the feature maps at 
	$\mathrm{d0}$--$\mathrm{d4}$ are $64\times64$, $32\times32$, $16\times16$, 
	$8\times8$, and $4\times4$, respectively, and the decoder resolutions are 
	mirrored. For visualization purposes we show Grad-CAM maps every 40 epochs over 
	the total of 1000 training epochs, since the early stages of training exhibit 
	the most pronounced changes.
	
	For each block, we compute Grad-CAM by multiplying the backward gradients with 
	the forward activations, averaging over channels, and scaling the resulting 
	values to the range $[0,1]$. This highlights the spatial regions that 
	contribute most strongly to the loss for each block at a given epoch, and 
	therefore reveals which structures the network focuses on during learning.
	
	We observe that image sequences from different datasets affect the attention 
	patterns of me-UNet in characteristic ways. In the blocks closest to the input 
	($\mathrm{d0}$, $\mathrm{d1}$) and output ($\mathrm{u3}$, $\mathrm{u4}$), the 
	model initially responds to a broad set of features, but as training progresses 
	the activations concentrate on the occupied regions of the domain (e.g., 
	dislocation lines, vortices, or reaction fronts) in most datasets (see 
	\cref{fig:grad_cam_adv}a, \cref{fig:grad_cam_diff}a, 
	\cref{fig:grad_cam_cdd_adv}a, \cref{fig:grad_cam_cdd_diff}a, 
	\cref{fig:grad_cam_cdd}, \cref{fig:grad_cam_cfd}, and 
	\cref{fig:grad_cam_gs_maze_d2}a). In some datasets (e.g., Gray--Scott DS-6b, 
	DS-6c, and DS-6d), the model already focuses on low-level features from the 
	first epochs (see \cref{fig:grad_cam_gs_alpha_v2}a, 
	\cref{fig:grad_cam_gs_worms_mu}a, and \cref{fig:grad_cam_gs_bubbles}a).
	
	In deeper encoder blocks ($\mathrm{d2}$, $\mathrm{d3}$, $\mathrm{d4}$) and the 
	corresponding decoder blocks ($\mathrm{u0}$, $\mathrm{u1}$, $\mathrm{u2}$), the 
	model concentrates on more global, semantic features. This is particularly 
	evident for the advection dataset DS-1, where vertical streaks in the Grad-CAM 
	maps reflect the horizontal motion of the advected structures (see 
	\cref{fig:grad_cam_adv}). For the Gray--Scott datasets (DS-6a, DS-6b) and 
	Kolmogorov flow (DS-5), deeper decoder blocks remain active throughout training 
	(see \cref{fig:grad_cam_gs_maze_d2}, \cref{fig:grad_cam_gs_alpha_v2}, and 
	\cref{fig:grad_cam_cfd}), reflecting the multi-scale nature of the pattern 
	dynamics.
	
	Taken together, these observations suggest that me-UNet leverages its 
	convolutional and periodic inductive biases to allocate representational 
	capacity in a physically meaningful way: shallow blocks capture local, 
	fine-scale details, while deeper blocks encode global structures and 
	long-range interactions. This aligns with the strong empirical performance of 
	me-UNet in both in-distribution and \gls{ood} prediction tasks.
	
	\section{Discussion}\label{sec:discussion}
	
	Our experiments provide a comparative picture of how different deep-learning 
	architectures behave as surrogates for 2D periodic \glspl{PDE} in a small-data 
	regime. Across all five \gls{PDE} families, the proposed me-UNet achieves 
	consistently low field-space error and high spectral similarity for in-
	distribution rollouts and remains stable over hundreds of time steps.
	
	In the same setting, neural-operator and transformer-based models (AFNO, PDE-
	Transformer, ViT, KAN-UNet) display more heterogeneous behavior: some of them 
	perform competitively on selected \glspl{PDE}, but they are typically less 
	robust across all datasets and more prone to the accumulation of high-frequency 
	artifacts in long rollouts. When evaluated with physics-based metrics, me-UNet 
	also tends to preserve conserved quantities such as energy (see 
	\cref{fig:physics_diagnostic}c) and mass (see \cref{fig:physics_diagnostic}a 
	and \cref{fig:physics_diagnostic}d) more accurately than the competing 
	architectures, which suggests that local convolutions with periodic padding 
	constitute a favorable inductive bias for these periodic \glspl{PDE}.
	
	More concretely, me-UNet performs well across all datasets DS-1--DS-6d, which 
	span models ranging from simple diffusion to more complex systems such as 
	Navier--Stokes (Kolmogorov flow) and continuum dislocation dynamics, using a 
	single set of hyperparameters. In contrast, the performance of the neural-
	operator and transformer-based baselines varies strongly between datasets when 
	trained with the same hyperparameters. For example, ViT and AFNO are 
	competitive on the \gls{CDD} (DS-4, \cref{fig:id_cdd}) and Kolmogorov flow 
	(DS-5, \cref{fig:id_cfd}) datasets, but their performance degrades 
	substantially on the Gray--Scott variants DS-6a (\cref{fig:id_gs_maze_d2}), 
	DS-6b (\cref{fig:id_gs_alpha_v2}), DS-6c (\cref{fig:id_gs_worms_mu}) and DS-6d 
	(\cref{fig:id_gs_bubbles}), where they tend to produce noisy or unstable 
	patterns.
	
	Beyond in-distribution performance, our results indicate that purely data-
	driven models can capture qualitative aspects of the dynamics even for initial 
	conditions that differ substantially from those seen during training. For 
	example, in the \gls{CDD} \gls{ood} experiments shown in \cref{fig:ood_cdd}, 
	the learned models reproduce the expansion and motion of individual loops or 
	lines, although the training data (\cref{fig:id_cdd}) contain only highly 
	crowded superpositions in which the motion of individual objects is not 
	visually apparent. The fact that me-UNet generalizes from these visually 
	complex initial states to much simpler line- and loop-like configurations 
	strongly suggests that it has internalized the underlying dislocation dynamics 
	rather than merely memorizing typical microstructure textures.
	
	Viewed through the lens of model complexity versus data complexity, our results 
	suggest that, in the considered setting, additional architectural 
	expressiveness does not automatically translate into better surrogates. The 
	training datasets are relatively small, and the underlying dynamics, while 
	nonlinear and sometimes chaotic, are governed by smooth fields on periodic 
	domains. In this regime, a convolutional architecture with strong inductive 
	biases---local filters, translation equivariance, multi-scale feature 
	aggregation, and periodic padding---appears to be well matched to the data 
	complexity, whereas higher-capacity models such as AFNO, PDE-Transformer, and 
	ViT may be harder to optimize and more prone to overfitting or the build-up of 
	high-frequency artifacts. We do not claim a universal relationship between 
	model capacity and performance for \glspl{PDE}, but our empirical study 
	highlights that carefully chosen inductive biases can be more important than 
	sheer architectural complexity when data are scarce.
	
	Our notion of out-of-distribution generalization is deliberately restricted to initial-condition shifts within a fixed PDE and parameter regime on periodic domains, as defined in Section~\ref{sec:data_generation}. We do not claim generalization across PDE families, parameter ranges, domain geometries, boundary conditions, or resolutions; extending the benchmark and architectures to such forms of distribution shift (parameter OOD, PDE-family OOD, and domain OOD) is an important direction for future work.

	Additionally, me-UNet works well with a relatively small amount of data. Based 
	on the experimental results shown in \cref{fig:cdd_rmse_nt_nsim_nseq}b, 
	sufficient performance on \gls{CDD} is obtained with only $20$ training simulations.
	While many works require a substantially larger number of simulations for 
	training---for example, \cite{oommen2022learning} use $2000$ simulations to train 
	\gls{DeepONet}, and the smallest number of simulations in \cite{kontolati2024learning} 
	is more than $260$---our results suggest that, at least for \gls{CDD}, a modest number 
	of simulations can suffice when using an architecture with appropriate 
	inductive biases. Furthermore, the time required for training me-UNet is much 
	lower than that of the state-of-the-art architectures considered here. Training 
	from scratch for $1000$ epochs with a batch size of $40$ on an NVIDIA RTX A6000 
	GPU takes roughly $5$~hours for me-UNet, while ViT, AFNO, PDE-Transformer, and 
	KAN-UNet take approximately $6$, $42$, $12$, and $9$~hours, respectively.
	
	Notably, several of the benchmarks are only partially observed during training
	(one scalar field per system; cf.~Table~\ref{tab:abbrev}). The strong performance 
	of me-UNet suggests that short temporal context can provide an effective 
	data-driven closure for the chosen observable in these periodic small-data 
	regimes.
	
	Moreover, Grad-CAM provides a qualitative window into the representations learned 
	by me-UNet during training. It highlights which spatial regions and U-Net blocks 
	most influence the prediction loss and how this focus evolves over epochs (see 
	Appendix~\ref{app:grad_cam_results}). Because the Grad-CAM maps are aggregated 
	over batches that span all simulation time steps, they do not resolve individual 
	physical times; nevertheless, they reveal dataset-dependent signatures such as 
	streak- or cross-shaped activations in transport-dominated systems and more 
	localized responses around reaction fronts or vortex cores for the Gray--Scott 
	and Kolmogorov-flow datasets. These patterns indicate that me-UNet learns internal 
	representations that reflect the dominant transport and pattern-formation 
	mechanisms in each benchmark, consistent with its strong in-distribution and 
	\gls{ood} performance.

	This benchmark is deliberately scoped to highlight small-data autoregressive forecasting on 2D periodic grids, which leaves several important extensions open.
	First, all experiments are carried 
	out on two-dimensional, structured, periodic grids with a fixed spatial 
	resolution. While this setting is representative of many canonical benchmarks, 
	it does not cover unstructured meshes, complex geometries, or adaptive 
	resolution. Second, the architectures are trained separately for each \gls{PDE} 
	family rather than as a single multi-task model, and we have not attempted 
	extensive per-architecture hyperparameter tuning beyond a common training 
	protocol. Third, although we incorporate physics-based metrics that quantify 
	the conservation of selected invariants and show that me-UNet can preserve them 
	reasonably well, we only \emph{encode periodic boundary conditions} explicitly, 
	via periodic padding; we do not impose any hard conservation constraints in the 
	architecture or loss. Future work could combine the periodic U-Net backbone 
	with physics-informed loss terms or hard constraints to more strictly enforce 
	mass, energy, or other invariants, and could explore similar inductive biases 
	within neural-operator architectures designed for more general domains.

	In addition, me-UNet, like most traditional CNN-based approaches, is naturally 
	suited to image-like datasets where values are distributed on a structured 
	grid, because it relies on fixed filter kernels (e.g., a $3 \times 3$ kernel 
	for most 2D convolutional operators). Such architectures struggle to handle 
	unstructured grids with irregular shapes and non-uniform node distributions. To 
	address this, it may be necessary either to combine me-UNet with techniques 
	that transform an unstructured grid into a structured grid, such as space-
	filling curves \cite{heaney2024applying}, or to use graph convolutional 
	networks \cite{jiang2019semi} to perform convolution-like operations on 
	irregular data.
	
	Furthermore, the current version of me-UNet is still limited to 2D simulation 
	results; however, there is room to extend the architecture to 3D problems, 
	which will be the next step following this work. The possible grid size used as 
	input for me-UNet is not limited to $64 \times 64$, but can be increased to 
	higher-resolution images (e.g., $128 \times 128$, $256 \times 256$, 
	$512 \times 512$, or higher), depending on the datasets produced, the 
	availability of GPU hardware, and the purpose of the research. Also, the 
	datasets produced for this study all have periodic boundary conditions in both 
	the horizontal and vertical directions for comparison purposes. Our neural 
	network architecture is in principle able to train with different types of 
	boundary conditions (e.g., Dirichlet boundaries) by changing the padding 
	technique, as discussed in \cref{sec:architecture}. The generalization ability 
	across boundary conditions is another important aspect to be addressed in 
	future work.
	
	Finally, we emphasize again that in this work \glspl{PDE} serve both as 
	scientifically important models and as controlled, interpretable generators of 
	high-dimensional spatio-temporal data. A surrogate model that cannot robustly 
	learn dynamics generated by well-posed \glspl{PDE} under controlled numerical 
	conditions is unlikely to succeed on heterogeneous, noisy measurement data 
	where the governing equations are only approximately known. Conversely, 
	architectures that perform well under the small-data and \gls{ood}-initial-
	condition regime studied here are promising candidates for subsequent 
	evaluation on real experimental and observational datasets.
	
	\section{Conclusion}
	We have presented a systematic empirical study of autoregressive deep-learning surrogates for two-dimensional periodic \glspl{PDE} in a small-data regime. Across five representative \gls{PDE} families, we compared the proposed me-UNet to several recent neural-operator and transformer-based architectures under a common training protocol. Using field-space error, spectral similarity, and physics-based metrics to quantify the preservation of invariants such as energy and mass, we found that me-UNet consistently provides accurate and stable rollouts for in-distribution test cases while requiring substantially less training time than the competing architectures.
	
	Within the considered setting, me-UNet also exhibits robust generalization to out-of-distribution initial conditions whose spatial structure differs markedly from the training data, including cases where the training set consists of highly crowded microstructures and the tests involve only a few well-separated objects. Together with the data-efficiency experiments, this supports the view that strong inductive biases in convolutional U-Nets---local filters, translation equivariance, multi-scale feature aggregation, and periodic padding that matches the boundary conditions---can be more beneficial than additional model complexity when data are scarce and the underlying dynamics are smooth on periodic domains.
	
	At the same time, our study is limited to 2D, structured, periodic grids at a fixed resolution, and to initial-condition \gls{ood} within a fixed \gls{PDE} and parameter regime. Extending the benchmark and models to parameter and geometry shifts, non-periodic boundary conditions, unstructured meshes, and three-dimensional problems, and combining the periodic U-Net backbone with physics-informed loss terms or hard constraints, are important directions for future work. More broadly, \glspl{PDE} in this work serve as controlled, interpretable generators of high-dimensional spatio-temporal data, and we expect that the same architectural and evaluation principles will also be relevant for surrogate modeling and forecasting in other domains where the dynamics are described by, or well approximated by, continuum models and related dynamical systems.
	
	
	\section*{Authors' contributions}\noindent
	BDN proposed the initial me-UNet architecture, implemented all models, generated the datasets, and ran all experiments. 
	BDN and SS jointly analyzed the data and wrote the manuscript. 
	SS guided the design of the study.

	
	\section*{Competing interests}\noindent
	The authors declare no competing interests.
	
	\section*{Data availability}\noindent
	All training datasets used in this study, together with scripts to generate them from the underlying PDE solvers, are provided in anonymized form as part of the supplementary material for review and will be released in a public repository upon acceptance.
	
	\section*{Code availability}\noindent
	All code required to reproduce the experiments, including model implementations, training scripts, and configuration files, is provided in anonymized form as part of the supplementary material for review and will be released in a public repository upon acceptance.
	

	\bibliography{literature}

@article{willard2023integrating,
  author  = {Willard, Jared and Jia, Xiaowei and Xu, Shaoming and Steinbach, Michael and Kumar, Vipin},
  title   = {Integrating Scientific Knowledge with Machine Learning for Engineering and Environmental Systems},
  journal = {ACM Computing Surveys},
  year    = {2023},
  volume  = {55},
  number  = {4},
  pages   = {1--37},
  doi     = {10.1145/3514228}
}

@article{dietrich2025scientific,
  author  = {Dietrich, Felix and Schilders, Wil},
  title   = {Scientific Machine Learning},
  journal = {Mathematische Semesterberichte},
  year    = {2025},
  volume  = {72},
  number  = {2},
  pages   = {89--115},
  doi     = {10.1007/s00591-025-00399-4}
}

@inproceedings{takamoto2022pdebench,
  title     = {PDEBench: An Extensive Benchmark for Scientific Machine Learning},
  author    = {Takamoto, Makoto and Praditia, Timothy and Leiteritz, Raphael
               and MacKinlay, Daniel and Alesiani, Francesco and Pfl{\"u}ger, Dirk
               and Niepert, Mathias},
  booktitle = {Advances in Neural Information Processing Systems (NeurIPS), Datasets and Benchmarks Track},
  year      = {2022},
  eprint    = {2210.07182},
  archivePrefix = {arXiv}
}

@article{nguyen_efficient_2024,
	title = {Efficient surrogate models for materials science simulations: {Machine} learning-based prediction of microstructure properties},
	issn = {26668270},
	shorttitle = {Efficient surrogate models for materials science simulations},
	url = {https://linkinghub.elsevier.com/retrieve/pii/S2666827024000203},
	doi = {10.1016/j.mlwa.2024.100544},
	language = {en},
	urldate = {2024-03-13},
	journal = {Machine Learning with Applications},
	author = {Nguyen, Binh Duong and Potapenko, Pavlo and Demirci, Aytekin and Govind, Kishan and Bompas, Sébastien and Sandfeld, Stefan},
	month = mar,
	year = {2024},
	pages = {100544},
}

@article{hornik1989multilayer,
	author  = {Hornik, Kurt and Stinchcombe, Maxwell and White, Halbert},
	title   = {Multilayer feedforward networks are universal approximators},
	journal = {Neural Networks},
	volume  = {2},
	number  = {5},
	pages   = {359--366},
	year    = {1989},
	issn    = {0893-6080},
	doi     = {10.1016/0893-6080(89)90020-8}
}

@article{lagaris1998artificial,
	author  = {Lagaris, Isaac E. and Likas, Aristidis and Fotiadis, Dimitrios I.},
	title   = {Artificial neural networks for solving ordinary and partial differential equations},
	journal = {IEEE Transactions on Neural Networks},
	volume  = {9},
	number  = {5},
	pages   = {987--1000},
	year    = {1998},
	issn    = {1045-9227},
	doi     = {10.1109/72.712178}
}

@article{einstein1915feldgleichungen,
  title={{Die Feldgleichungen der Gravitation}},
  author={Einstein, Albert},
  journal={Sitzungsberichte der K{\"o}niglich Preu{\ss}ischen Akademie der Wissenschaften},
  pages={844--847},
  year={1915},
  doi={https://adsabs.harvard.edu/pdf/1915SPAW.......844E}
}

@article{schrodinger1926undulatory,
  title={An undulatory theory of the mechanics of atoms and molecules},
  author={Schr{\"o}dinger, Erwin},
  journal={Physical review},
  volume={28},
  number={6},
  pages={1049},
  year={1926},
  publisher={APS},
  doi={https://doi.org/10.1103/PhysRev.28.1049}
}

@inproceedings{navier1821lois,
  title={Sur les Lois des Mouvement des Fluides, en Ayant Egard a L’adhesion des Molecules},
  author={Navier, CLMH},
  booktitle={Annales de Chimie et de Physique},
  volume={19},
  number={244},
  pages={1821},
  year={1821},
  organization={Lavoisier Paris, France}
}

@article{vogelsberger2014introducing,
  title={Introducing the Illustris Project: simulating the coevolution of dark and visible matter in the Universe},
  author={Vogelsberger, Mark and Genel, Shy and Springel, Volker and Torrey, Paul and Sijacki, Debora and Xu, Dandan and Snyder, Greg and Nelson, Dylan and Hernquist, Lars},
  journal={Monthly Notices of the Royal Astronomical Society},
  volume={444},
  number={2},
  pages={1518--1547},
  year={2014},
  publisher={Oxford University Press},
  doi={https://doi.org/10.1093/mnras/stu1536}
}

@article{randall2019100,
  title={100 years of earth system model development},
  author={Randall, David A and Bitz, Cecilia M and Danabasoglu, Gokhan and Denning, A Scott and Gent, Peter R and Gettelman, Andrew and Griffies, Stephen M and Lynch, Peter and Morrison, Hugh and Pincus, Robert and others},
  journal={Meteorological Monographs},
  volume={59},
  pages={12--1},
  year={2019}
}

@article{lee1990neural,
  title={Neural algorithm for solving differential equations},
  author={Lee, Hyuk and Kang, In Seok},
  journal={Journal of Computational Physics},
  volume={91},
  number={1},
  pages={110--131},
  year={1990},
  publisher={Elsevier},
  doi={https://doi.org/10.1016/0021-9991(90)90007-N}
}

@article{dissanayake1994neural,
  title={Neural-network-based approximations for solving partial differential equations},
  author={Dissanayake, MWM Gamini and Phan-Thien, Nhan},
  journal={communications in Numerical Methods in Engineering},
  volume={10},
  number={3},
  pages={195--201},
  year={1994},
  publisher={Wiley Online Library},
  doi={https://doi.org/10.1002/cnm.1640100303}
}

@article{rao2023encoding,
  title={Encoding physics to learn reaction--diffusion processes},
  author={Rao, Chengping and Ren, Pu and Wang, Qi and Buyukozturk, Oral and Sun, Hao and Liu, Yang},
  journal={Nature Machine Intelligence},
  volume={5},
  number={7},
  pages={765--779},
  year={2023},
  publisher={Nature Publishing Group UK London},
  doi={https://doi.org/10.1038/s42256-023-00685-7}
}

@inproceedings{ronneberger2015u,
  title={U-net: Convolutional networks for biomedical image segmentation},
  author={Ronneberger, Olaf and Fischer, Philipp and Brox, Thomas},
  booktitle={Medical image computing and computer-assisted intervention--MICCAI 2015: 18th international conference, Munich, Germany, October 5-9, 2015, proceedings, part III 18},
  pages={234--241},
  year={2015},
  organization={Springer},
  doi={https://doi.org/10.1007/978-3-319-24574-4_28}
}

@article{pearson1993complex,
  title={Complex patterns in a simple system},
  author={Pearson, John E},
  journal={Science},
  volume={261},
  number={5118},
  pages={189--192},
  year={1993},
  publisher={American Association for the Advancement of Science},
  doi={https://doi.org/10.1126/science.261.5118.189}
}

@article{hochrainer2014continuum,
  title={Continuum dislocation dynamics: towards a physical theory of crystal plasticity},
  author={Hochrainer, Thomas and Sandfeld, Stefan and Zaiser, Michael and Gumbsch, Peter},
  journal={Journal of the Mechanics and Physics of Solids},
  volume={63},
  pages={167--178},
  year={2014},
  publisher={Elsevier}, 
  doi={https://doi.org/10.1016/j.jmps.2013.09.012}
}

@article{sandfeld2015microstructural,
  title={Microstructural comparison of the kinematics of discrete and continuum dislocations models},
  author={Sandfeld, Stefan and Po, Giacomo},
  journal={Modelling and Simulation in Materials Science and Engineering},
  volume={23},
  number={8},
  pages={085003},
  year={2015},
  publisher={IOP Publishing},
  doi={https://doi.org/10.1088/0965-0393/23/8/085003F}
}

@article{monavari2014comparison,
  title={Comparison of closure approximations for continuous dislocation dynamics},
  author={Monavari, Mehran and Zaiser, Michael and Sandfeld, Stefan},
  journal={MRS Online Proceedings Library (OPL)},
  volume={1651},
  pages={mrsf13--1651},
  year={2014},
  publisher={Cambridge University Press},
  doi={https://doi.org/10.1557/opl.2014.62}
}

@inproceedings{johnson2016perceptual,
  title={Perceptual losses for real-time style transfer and super-resolution},
  author={Johnson, Justin and Alahi, Alexandre and Fei-Fei, Li},
  booktitle={Computer Vision--ECCV 2016: 14th European Conference, Amsterdam, The Netherlands, October 11-14, 2016, Proceedings, Part II 14},
  pages={694--711},
  year={2016},
  organization={Springer},
  doi={https://doi.org/10.1007/978-3-319-46475-6_43}
}

@article{qu2022learning,
  title={Learning time-dependent PDEs with a linear and nonlinear separate convolutional neural network},
  author={Qu, Jiagang and Cai, Weihua and Zhao, Yijun},
  journal={Journal of Computational Physics},
  volume={453},
  pages={110928},
  year={2022},
  publisher={Elsevier},
  doi={https://doi.org/10.1016/j.jcp.2021.110928}
}

@article{kochkov2021machine,
  title={Machine learning--accelerated computational fluid dynamics},
  author={Kochkov, Dmitrii and Smith, Jamie A and Alieva, Ayya and Wang, Qing and Brenner, Michael P and Hoyer, Stephan},
  journal={Proceedings of the National Academy of Sciences},
  volume={118},
  number={21},
  pages={e2101784118},
  year={2021},
  publisher={National Acad Sciences},
  doi={https://doi.org/10.1073/pnas.2101784118}
}

@article{zhao2024improved,
  title={A improved pooling method for convolutional neural networks},
  author={Zhao, Lei and Zhang, Zhonglin},
  journal={Scientific Reports},
  volume={14},
  number={1},
  pages={1589},
  year={2024},
  publisher={Nature Publishing Group UK London},
  doi={https://doi.org/10.1038/s41598-024-51258-6}
}

@book{kolmogorov1995turbulence,
  title={Turbulence: the legacy of A. N. Kolmogorov},
  author={Kolmogorov, Andre{\u\i} Nikolaevich},
  year={1995},
  publisher={Cambridge university press}
}

@article{raissi2019physics,
  title={Physics-informed neural networks: A deep learning framework for solving forward and inverse problems involving nonlinear partial differential equations},
  author={Raissi, Maziar and Perdikaris, Paris and Karniadakis, George E},
  journal={Journal of Computational physics},
  volume={378},
  pages={686--707},
  year={2019},
  publisher={Elsevier},
  doi={https://doi.org/10.1016/j.jcp.2018.10.045}
}

@article{li2020fourier,
  title={Fourier neural operator for parametric partial differential equations},
  author={Li, Zongyi and Kovachki, Nikola and Azizzadenesheli, Kamyar and Liu, Burigede and Bhattacharya, Kaushik and Stuart, Andrew and Anandkumar, Anima},
  journal={arXiv preprint arXiv:2010.08895},
  year={2020},
  doi={https://arxiv.org/pdf/2010.08895}
}

@article{lu2021learning,
  title={Learning nonlinear operators via DeepONet based on the universal approximation theorem of operators},
  author={Lu, Lu and Jin, Pengzhan and Pang, Guofei and Zhang, Zhongqiang and Karniadakis, George Em},
  journal={Nature machine intelligence},
  volume={3},
  number={3},
  pages={218--229},
  year={2021},
  publisher={Nature Publishing Group UK London},
  doi={https://doi.org/10.1038/s42256-021-00302-5}
}

@article{oommen2022learning,
  title={Learning two-phase microstructure evolution using neural operators and autoencoder architectures},
  author={Oommen, Vivek and Shukla, Khemraj and Goswami, Somdatta and Dingreville, R{\'e}mi and Karniadakis, George Em},
  journal={npj Computational Materials},
  volume={8},
  number={1},
  pages={190},
  year={2022},
  publisher={Nature Publishing Group UK London},
  doi={https://doi.org/10.1038/s41524-022-00876-7}
}

@article{kontolati2024learning,
  title={Learning nonlinear operators in latent spaces for real-time predictions of complex dynamics in physical systems},
  author={Kontolati, Katiana and Goswami, Somdatta and Em Karniadakis, George and Shields, Michael D},
  journal={Nature Communications},
  volume={15},
  number={1},
  pages={5101},
  year={2024},
  publisher={Nature Publishing Group UK London},
  doi={https://doi.org/10.1038/s41467-024-49411-w}
}

@article{pathak2022fourcastnet,
  title={Fourcastnet: A global data-driven high-resolution weather model using adaptive fourier neural operators},
  author={Pathak, Jaideep and Subramanian, Shashank and Harrington, Peter and Raja, Sanjeev and Chattopadhyay, Ashesh and Mardani, Morteza and Kurth, Thorsten and Hall, David and Li, Zongyi and Azizzadenesheli, Kamyar and others},
  journal={arXiv preprint arXiv:2202.11214},
  year={2022},
  doi={https://doi.org/10.48550/arXiv.2202.11214}
}

@article{dosovitskiy2020image,
  title={An image is worth 16x16 words: Transformers for image recognition at scale},
  author={Dosovitskiy, Alexey and Beyer, Lucas and Kolesnikov, Alexander and Weissenborn, Dirk and Zhai, Xiaohua and Unterthiner, Thomas and Dehghani, Mostafa and Minderer, Matthias and Heigold, Georg and Gelly, Sylvain and others},
  journal={arXiv preprint arXiv:2010.11929},
  year={2020},
  doi={https://doi.org/10.48550/arXiv.2010.11929}
}

@misc{xiangbo2024,
  author = {Xiangbo Gao},
  title = {KA-Conv: Kolmogorov-Arnold Convolutional Networks with Various Basis Functions},
  year = {2024},
  publisher = {GitHub},
  journal = {GitHub repository},
  howpublished = {\url{https://github.com/XiangboGaoBarry/KA-Conv}}
}

@article{holzschuh2025pde,
  title={PDE-Transformer: Efficient and Versatile Transformers for Physics Simulations},
  author={Holzschuh, Benjamin and Liu, Qiang and Kohl, Georg and Thuerey, Nils},
  booktitle={Forty-second International Conference on Machine Learning, {ICML} 2025, Vancouver, Canada, July 13-19, 2025},
  year={2025}
}

@article{koehler2024apebench,
  title={{APEBench}: A Benchmark for Autoregressive Neural Emulators of {PDE}s},
  author={Felix Koehler and Simon Niedermayr and Ruediger Westermann and Nils Thuerey},
  journal={Advances in Neural Information Processing Systems (NeurIPS)},
  volume={38},
  year={2024}
}

@article{sun2025towards,
  title={Towards a foundation model for partial differential equations: Multioperator learning and extrapolation},
  author={Sun, Jingmin and Liu, Yuxuan and Zhang, Zecheng and Schaeffer, Hayden},
  journal={Physical Review E},
  volume={111},
  number={3},
  pages={035304},
  year={2025},
  publisher={APS},
  doi={https://doi.org/10.1103/PhysRevE.111.035304}
}

@article{heaney2024applying,
  title={Applying convolutional neural networks to data on unstructured meshes with space-filling curves},
  author={Heaney, Claire E and Li, Yuling and Matar, Omar K and Pain, Christopher C},
  journal={Neural Networks},
  volume={175},
  pages={106198},
  year={2024},
  publisher={Elsevier},
  doi={https://doi.org/10.1016/j.neunet.2024.106198}
}

@inproceedings{jiang2019semi,
  title={Semi-supervised learning with graph learning-convolutional networks},
  author={Jiang, Bo and Zhang, Ziyan and Lin, Doudou and Tang, Jin and Luo, Bin},
  booktitle={Proceedings of the IEEE/CVF conference on computer vision and pattern recognition},
  pages={11313--11320},
  year={2019}
}

@article{paszke2019pytorch,
  title={Pytorch: An imperative style, high-performance deep learning library},
  author={Paszke, Adam and Gross, Sam and Massa, Francisco and Lerer, Adam and Bradbury, James and Chanan, Gregory and Killeen, Trevor and Lin, Zeming and Gimelshein, Natalia and Antiga, Luca and others},
  journal={Advances in neural information processing systems},
  volume={32},
  year={2019}
}

@book{logg2012automated,
  title={Automated solution of differential equations by the finite element method: The FEniCS book},
  author={Logg, Anders and Mardal, Kent-Andre and Wells, Garth},
  volume={84},
  year={2012},
  publisher={Springer Science \& Business Media}
}

@article{brooks1982streamline,
  title={Streamline upwind/Petrov-Galerkin formulations for convection dominated flows with particular emphasis on the incompressible Navier-Stokes equations},
  author={Brooks, Alexander N and Hughes, Thomas JR},
  journal={Computer methods in applied mechanics and engineering},
  volume={32},
  number={1-3},
  pages={199--259},
  year={1982},
  publisher={Elsevier}
}

@article{zhu2019physics,
  title={Physics-constrained deep learning for high-dimensional surrogate modeling and uncertainty quantification without labeled data},
  author={Zhu, Yinhao and Zabaras, Nicholas and Koutsourelakis, Phaedon-Stelios and Perdikaris, Paris},
  journal={Journal of Computational Physics},
  volume={394},
  pages={56--81},
  year={2019},
  publisher={Elsevier},
  doi={https://doi.org/10.1016/j.jcp.2019.05.024}
}

@article{shukla2020physics,
  title={Physics-informed neural network for ultrasound nondestructive quantification of surface breaking cracks},
  author={Shukla, Khemraj and Di Leoni, Patricio Clark and Blackshire, James and Sparkman, Daniel and Karniadakis, George Em},
  journal={Journal of Nondestructive Evaluation},
  volume={39},
  number={3},
  pages={61},
  year={2020},
  publisher={Springer},
  doi={https://doi.org/10.1007/s10921-020-00705-1}
}

@article{zhang2020learning,
  title={Learning in modal space: Solving time-dependent stochastic PDEs using physics-informed neural networks},
  author={Zhang, Dongkun and Guo, Ling and Karniadakis, George Em},
  journal={SIAM Journal on Scientific Computing},
  volume={42},
  number={2},
  pages={A639--A665},
  year={2020},
  publisher={SIAM}
}

@article{ren2022phycrnet,
  title={PhyCRNet: Physics-informed convolutional-recurrent network for solving spatiotemporal PDEs},
  author={Ren, Pu and Rao, Chengping and Liu, Yang and Wang, Jian-Xun and Sun, Hao},
  journal={Computer Methods in Applied Mechanics and Engineering},
  volume={389},
  pages={114399},
  year={2022},
  publisher={Elsevier},
  doi={https://doi.org/10.1016/j.cma.2021.114399}
}

@article{ren2023physr,
  title={PhySR: Physics-informed deep super-resolution for spatiotemporal data},
  author={Ren, Pu and Rao, Chengping and Liu, Yang and Ma, Zihan and Wang, Qi and Wang, Jian-Xun and Sun, Hao},
  journal={Journal of Computational Physics},
  volume={492},
  pages={112438},
  year={2023},
  publisher={Elsevier},
  doi={https://doi.org/10.1016/j.jcp.2023.112438}
}

@article{yuan2024f,
  title={f-PICNN: a physics-informed convolutional neural network for partial differential equations with space-time domain},
  author={Yuan, Biao and Wang, He and Heitor, Ana and Chen, Xiaohui},
  journal={Journal of Computational Physics},
  volume={515},
  pages={113284},
  year={2024},
  publisher={Elsevier},
  doi={https://doi.org/10.1016/j.jcp.2024.113284}
}
	\bibliographystyle{unsrtnat}

	\appendix
	\section{Loss function}
	
	\subsection{MSE loss}
	The mean-squared error (MSE) loss is widely used in the machine-learning community. It computes the average squared difference between the predicted and true values:
	\begin{align}
		\label{eq:mse_loss}
		\mathcal{L}_\mathrm{MSE}
		= \frac{1}{N} \sum_{i=1}^{N} \bigl(\hat{y}_{i} - y_{i}\bigr)^2,
	\end{align}
	where $N$ is the number of data points, $y_i$ is the true value, and $\hat{y}_i$ is the predicted value for the $i$-th sample.
	
	\subsection{Perceptual loss}
	Perceptual loss was introduced by Johnson et al.~\cite{johnson2016perceptual}; it is computed based on features extracted from a pre-trained network rather than directly from pixel-wise differences. Perceptual loss captures more abstract and global image qualities that are often more relevant for human perception, whereas pixel-based losses such as MSE penalize local differences and can lead to overly smooth or blurry images:
	\begin{align}
		\label{eq:perceptual_loss}
		\mathcal{L}_\mathrm{pc}
		= \sum_{i} \bigl\lVert \Phi_{i}(\hat{y}) - \Phi_{i}(y) \bigr\rVert^2,
	\end{align}
	where $\Phi_{i}$ denotes the feature map from the $i$-th layer of the pre-trained network, and $y$ and $\hat{y}$ are the true and predicted images, respectively. In this work, we use VGG-16 as the pre-trained network to obtain feature maps.
	
	\section{Detailed description of the mathematical models}
	
	\paragraph{Advection equation}
	The advection (or transport) equation is a fundamental mathematical model used to describe a variety of physical phenomena in fields such as biology, chemistry, and materials science. It predicts how a quantity is transported over time due to the motion of particles or a background flow.
	
	The general form of the advection equation in our setting is
	\begin{align}
		\label{eq:advection}
		\frac{\partial u}{\partial t}
		= v_x \frac{\partial u}{\partial x} + v_y \frac{\partial u}{\partial y},
	\end{align}
	where $u$ is the scalar quantity being transported (e.g., temperature or concentration), and $v_x = 1$, $v_y = 0$ are the velocity components in the $x$- and $y$-directions.
	
	The initial values consist of ``blobs'' introduced into the system as a Gaussian exponential function,
	\begin{align}
		\label{eq:gaussian}
		c = \exp\bigl(-a\bigl((x-x_0)^2 + (y-y_0)^2\bigr)\bigr).
	\end{align}
	The number of these blobs is randomly chosen within the range $(1, 10)$, and the blobs are randomly positioned inside the quadratic domain with edge length $l_x = l_y = 10$. For the Gaussian distribution the length scale parameter was chosen as $a = 10$. The spatial mesh is $100 \times 100$.
	
	The simulation is implemented in FEniCS, an open-source computing platform for solving \glspl{PDE} with the finite element method. Each simulation is run for $100$ time steps, and the field is saved at every step. The time step size is $0.01$, and a Streamline-Upwind Petrov--Galerkin (SUPG) stabilization term~\cite{brooks1982streamline} is used to prevent numerical oscillations and instabilities in this convection-dominated problem. We use the default ``Sparse LU'' solver provided by FEniCS.
	
	\paragraph{Diffusion equation}
	The diffusion equation is another fundamental mathematical model used to describe a wide range of physical processes, including heat conduction and the spreading of chemical species. It predicts how a quantity spreads out over time due to random motion of particles.
	
	The general form of the diffusion equation is
	\begin{align}
		\label{eq:diffusion}
		\frac{\partial u}{\partial t}
		= c_x \frac{\partial^2 u}{\partial x^2}
		+ c_y \frac{\partial^2 u}{\partial y^2},
	\end{align}
	where $u$ is the diffusing quantity (e.g., temperature), and $c_x = c_y = 1$ are the diffusion coefficients in the $x$- and $y$-directions.
	
	As before, the number of initial blobs is randomly chosen in the range $(1, 10)$ and the blobs are randomly distributed inside the domain with $l_x = l_y = 10$. Each blob again is again obtained from a  Gaussian function \cref{eq:gaussian} with $a = 10$. The spatial mesh is $100 \times 100$.
	
	The simulation is implemented in FEniCS and run for $100$ time steps with the field saved at every step. The time step size is $0.01$, and we use the default ``Sparse LU'' solver for this diffusion equation.
	
	\paragraph{Reaction--diffusion equation (Gray--Scott model)}
	A two-dimensional Gray--Scott system~\cite{pearson1993complex} describes the behavior of two reacting and diffusing substances by
	\begin{align}
		\label{eq:gray_scott}
		\frac{\partial u}{\partial t} &= -u v^2 + c_u \Delta u + f(1-u), \\
		\frac{\partial v}{\partial t} &= \phantom{-}u v^2 + c_v \Delta v - (f+k)v,
	\end{align}
	where $u$ and $v$ are the concentrations of the two reacting species $U$ and $V$, $c_u$ and $c_v$ are their diffusion coefficients, and $f$ and $k$ are feed and kill rates.
	
	We use the exponax package~\cite{koehler2024apebench} to simulate a set of Gray--Scott systems with different parameter choices corresponding to DS-6a, DS-6b, DS-6c, and DS-6d; the parameters are given in \cref{tab:gs_params}. The spatial mesh sizes are either $128 \times 128$ or $256 \times 256$, as listed in the table. Each simulation is run for $200$ time steps, and the field values are saved every $2$ steps, resulting in $100$ stored snapshots per simulation.
	
	\begin{table}[htb!]
		\caption{Gray--Scott model simulation parameters.}
		\label{tab:gs_params}
		\centering
		\begin{tabular}{l|c|c|c|c}
			\toprule
			& Feed rate $f$  & Kill rate $k$ & \# Gaussian blobs & Spatial mesh \\ 
			\cmidrule(r){1-5}
			DS-6a & $0.03$ &  $0.0565$ &  $50$--$200$ &  $256 \times 256$ \\ 
			DS-6b & $0.018$ & $0.051$ &  $5$--$10$   &  $128 \times 128$ \\ 
			DS-6c & $0.058$ & $0.065$ &  $5$--$10$   &  $128 \times 128$ \\ 
			DS-6d & $0.082$ & $0.059$ &  $5$--$10$   &  $128 \times 128$ \\ 
			\bottomrule
		\end{tabular}
	\end{table}
	
	\paragraph{Navier--Stokes equation (Kolmogorov flow)}
	The Navier--Stokes equations~\cite{kolmogorov1995turbulence} describe the motion of incompressible fluid flows:
	\begin{align}
		\label{eq:navier_stokes}
		\frac{\partial \boldsymbol{u}}{\partial t} &=
		-\nabla \cdot (\boldsymbol{u} \otimes \boldsymbol{u})
		+ \frac{1}{\mathrm{Re}} \nabla^2 \boldsymbol{u}
		- \frac{1}{\rho} \nabla p
		+ \boldsymbol{f}, \\
		\nabla \cdot \boldsymbol{u} &= 0,
	\end{align}
	where $\boldsymbol{u}$ is the velocity field, $\boldsymbol{f}$ is an external forcing, $\rho$ is the density, $\mathrm{Re}$ is the Reynolds number, and $p$ is the pressure. In two dimensions, the scalar vorticity field
	\[
	\omega = \partial_x u_y - \partial_y u_x
	\]
	is often used to characterize incompressible flows.
	
	We simulate Kolmogorov flow for decaying turbulence using the open-source JAX-CFD code~\cite{kochkov2021machine}, which employs finite-volume/finite-difference discretizations on a staggered grid (pressure at cell centers and velocity components on corresponding faces). The spatial grid size is $256 \times 256$, and each simulation runs for $100$ time steps with the vorticity field saved at every step. The density is set to $\rho = 1$, the Reynolds number to $\mathrm{Re} = 1000$, and the Kolmogorov forcing terms are kept at the default values provided in the package examples. Kolmogorov flow is known to be unstable and chaotic, so small numerical errors can accumulate over time and cause trajectories to diverge.

	\paragraph{Continuum dislocation dynamics (CDD) model}
	Continuum dislocation dynamics (\gls{CDD}) \cite{hochrainer2014continuum} 
	is a set of \glspl{PDE} that 
	describe the transport and evolution of curved lines under a given velocity field. Lines can be represented as quasi-discrete, single lines  or even as ``smeared-out'' continuum-like bundles of many lines (because the \glspl{PDE} are the result of a statistical field theory).
	
	In the 2D setting considered here, we assume a velocity $\boldsymbol{v}$ with constant magnitude $|\boldsymbol{v}| = \text{const}$ that is by construction always perpendicular to the local line tangent. As a consequence, a circular dislocation loop will expand, whereas a straight line segment will experience only pure translation. The key difference between these two extreme geometrical objects is that expanding loops increase the total line length in the system, while straight segments merely move without changing the total length. 
	
	This purely geometric behavior is---in the field of materials science---central to dislocation plasticity, where the motion and evolution of line-like defects inside metals and semiconductors govern plastic deformation, as demonstrated for example in Sandfeld and Po~\cite{sandfeld2015microstructural}. There, the purely geometrical aspects described above are central aspects.
	
	In this work we consider a \gls{CDD} model following the formulation in~\cite{hochrainer2014continuum}. The state is represented by three fields: the total density $\rho_\mathrm{t}$ (whose volume integral gives the total line length inside the volume), a vector of geometrically necessary dislocation density $\boldsymbol{\varrho} = [\varrho_{\mathrm{s}}, \varrho_{\mathrm{e}}]$ (corresponding to screw and edge dislocations), and a curvature density $q_\mathrm{t}$ (whose volume integral gives the number of closed  loops as a multiple of $2\pi$). The temporal evolution is governed by transport-type equations
	\begin{align}
		\label{eq:cdd}
		\partial_t \rho_\mathrm{t} &= -\nabla \cdot (v \boldsymbol{\varrho}^{\perp}) + v q_\mathrm{t}, \\ \nonumber
		\partial_t \boldsymbol{\varrho} &= -\nabla (v \rho_\mathrm{t}), \\ \nonumber
		\partial_t q_\mathrm{t} &= -\nabla \cdot \bigl(-v \boldsymbol{Q}^{(1)} + \boldsymbol{A}^{(2)} \cdot \nabla v\bigr),
	\end{align}
	where a dot denotes the scalar product and $\boldsymbol{\varrho}^{\perp} = [\varrho_{\mathrm{e}}, -\varrho_{\mathrm{s}}]$ is the $90^\circ$-rotated GND density vector. Following~\cite{monavari2014comparison}, we assume
	\begin{align}
		\boldsymbol{Q}^{(1)} &=
		-\boldsymbol{\varrho}^{\perp} \,\frac{q_\mathrm{t}}{\rho_\mathrm{t}}, \\ \nonumber
		\boldsymbol{A}^{(2)} &=
		\frac{\rho_\mathrm{t}}{2}
		\left[ (1+\Psi)\,\boldsymbol{l}_\varrho \otimes \boldsymbol{l}_\varrho
		+ (1-\Psi)\,\boldsymbol{l}_{\varrho^{\perp}} \otimes \boldsymbol{l}_{\varrho^{\perp}} \right],
	\end{align}
	where $\boldsymbol{l}_\varrho = \boldsymbol{\varrho} / |\boldsymbol{\varrho}|$ is the average line direction and $\boldsymbol{l}_{\varrho^{\perp}}$ is the unit vector perpendicular to $\boldsymbol{l}_\varrho$. The anisotropy parameter $\Psi$ is approximated as
	\begin{align}
		\Psi \approx
		\frac{\left( |\boldsymbol{\varrho}| / \rho_\mathrm{t} \right)^2
			\left(1 + \left(|\boldsymbol{\varrho}| / \rho_\mathrm{t}\right)^4\right)}{2}.
	\end{align}
	%
	
	We implement the \gls{CDD} model in FEniCS. For simplicity, we assign a constant velocity $v$, corresponding to dislocation glide on a fixed slip system, and neglect interactions between dislocation loops (e.g., annihilation, cross slip). The number of circular dislocation loops is chosen randomly in the range $[50, 500]$ with radius $r_0 = 400b$. The spatial mesh is $100 \times 100$, and each simulation runs for $1000$ time steps with fields saved every $10$ time steps (resulting in 100 stored snapshots). We consider an ensemble of dislocation loops with the same Burgers vector $\boldsymbol{b}$, and a domain of size $l_x = l_y = 2000b$ with $b = \SI{0.256}{\nm}$. Periodic boundary conditions are used in both directions so that loops leaving the domain on one side re-enter on the opposite side. Further details and parameter choices can be found in \cite{sandfeld2015microstructural}.
	
	
	\paragraph{Reduced \gls{CDD} model}
	To produce the datasets DS-3a and DS-3b, containing advection-like or diffusion-like motion of loop distributions, we retain only the first two equations in \cref{eq:cdd}, drop the curvature-density term and add the diffusion terms. The equation becomes,

	\begin{align}
		\label{eq:reduced_cdd}
		\partial_t \rho_\mathrm{t} &= -\nabla \cdot (v \boldsymbol{\varrho}^{\perp}) + D \nabla^2 \rho_\mathrm{t}, \\ \nonumber
		\partial_t \boldsymbol{\varrho} &= -\nabla (v \rho_\mathrm{t}) + D \nabla^2 \boldsymbol{\varrho}, 
	\end{align}

	All parameters for solving this equation are similar to that of the \gls{CDD} model. We assume a velocity $\boldsymbol{v}$ with constant magnitude $|\boldsymbol{v}| = \text{const}$ and diffusion coefficient $D=0$ for dataset DS-3a; $|\boldsymbol{v}| = 0$ and $D= \text{const}$ for DS-3b.

	\section{Training behaviour}
	
\begin{figure*}[h!tb]
		\centering
		\begin{subfigure}{0.9\textwidth}
			\centering
			\includegraphics[width=1\textwidth]{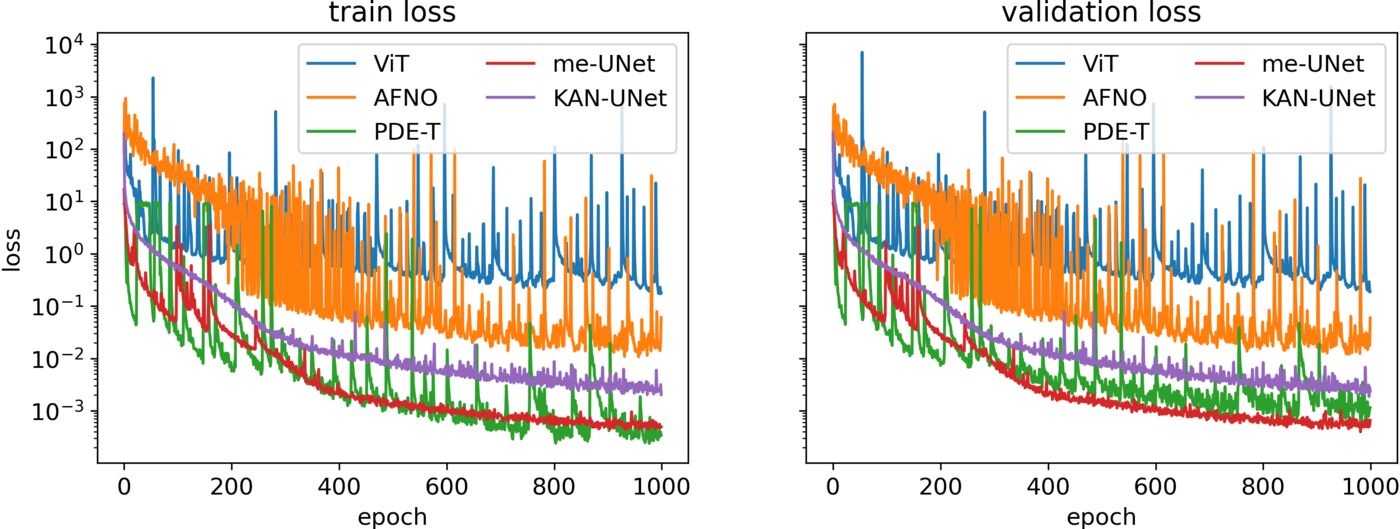}\\
			\caption{}
		\end{subfigure}
		\begin{subfigure}{0.9\textwidth}
			\centering
			\includegraphics[width=1.\textwidth]{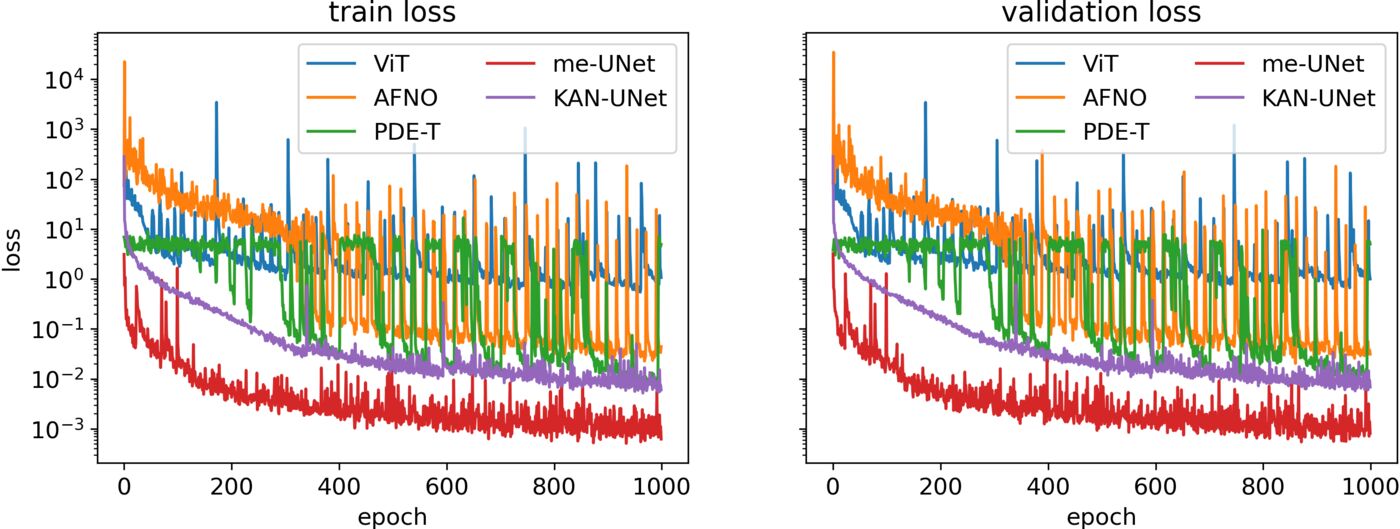}\\
			\caption{}
		\end{subfigure}
		\caption{Training and validation loss curves (MSE + perceptual loss) for the different architectures on (a) the \gls{CDD} dataset (DS-4) and (b) the Gray--Scott dataset DS-6a. me-UNet converges reliably and shows no signs of overfitting under the common training setup, whereas ViT and AFNO struggle in the small-data regime.}
		\label{fig:train_val_loss}
	\end{figure*}

	Figure~\ref{fig:train_val_loss} shows the convergence behavior of training and 
	validation loss for the \gls{CDD} and Gray--Scott DS-6a datasets. me-UNet 
	converges faster and more smoothly than the other architectures, without 
	evidence of overfitting, while ViT and AFNO have difficulty training with the 
	limited data. PDE-Transformer converges on DS-4 but struggles on DS-6a, 
	consistent with its weaker predictive performance on the latter.

	\section{Visualization of physics-based metrics across architectures}
	\label{app:physics_diagnostic_results}
	
\begin{figure*}[htb]
		\centering
		\begin{subfigure}{0.45\textwidth}
			\centering
			\includegraphics[width=0.9\textwidth]{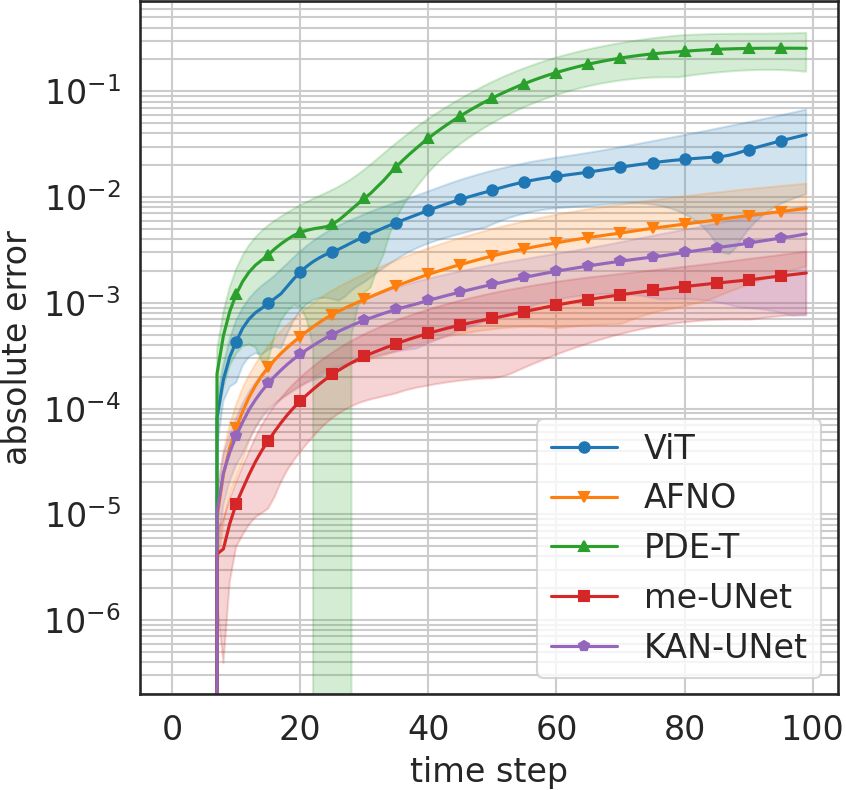}\\
			\bigskip
			\caption{}
		\end{subfigure}
		\centering
		\begin{subfigure}{0.45\textwidth}
			\centering
			\includegraphics[width=0.9\textwidth]{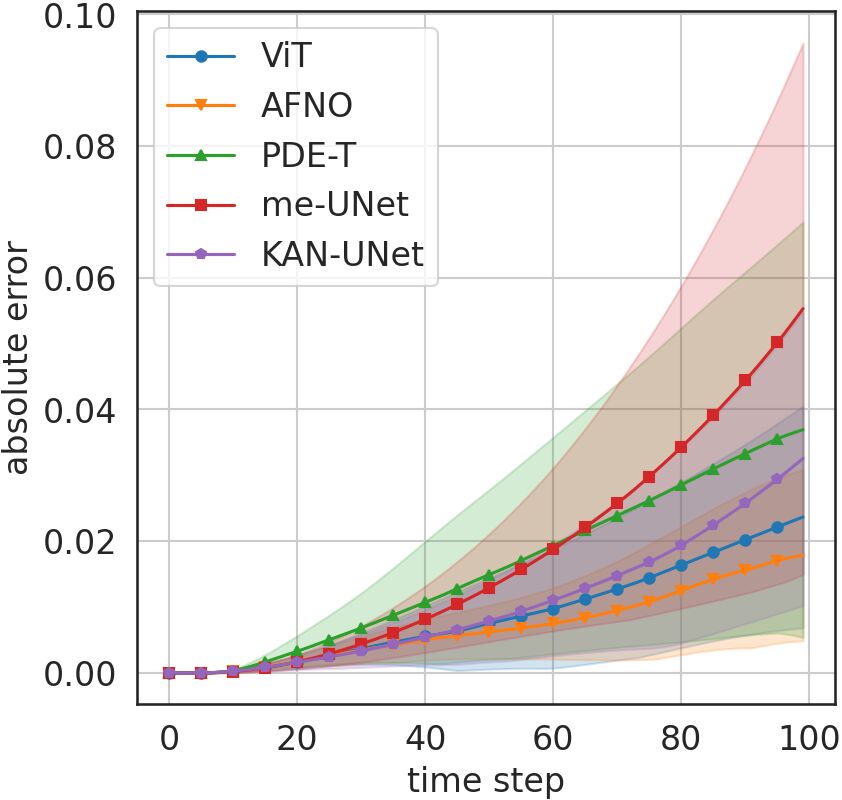}\\
			\bigskip
			\caption{}
		\end{subfigure}
		\centering
		\begin{subfigure}{0.45\textwidth}
			\centering
			\includegraphics[width=0.9\textwidth]{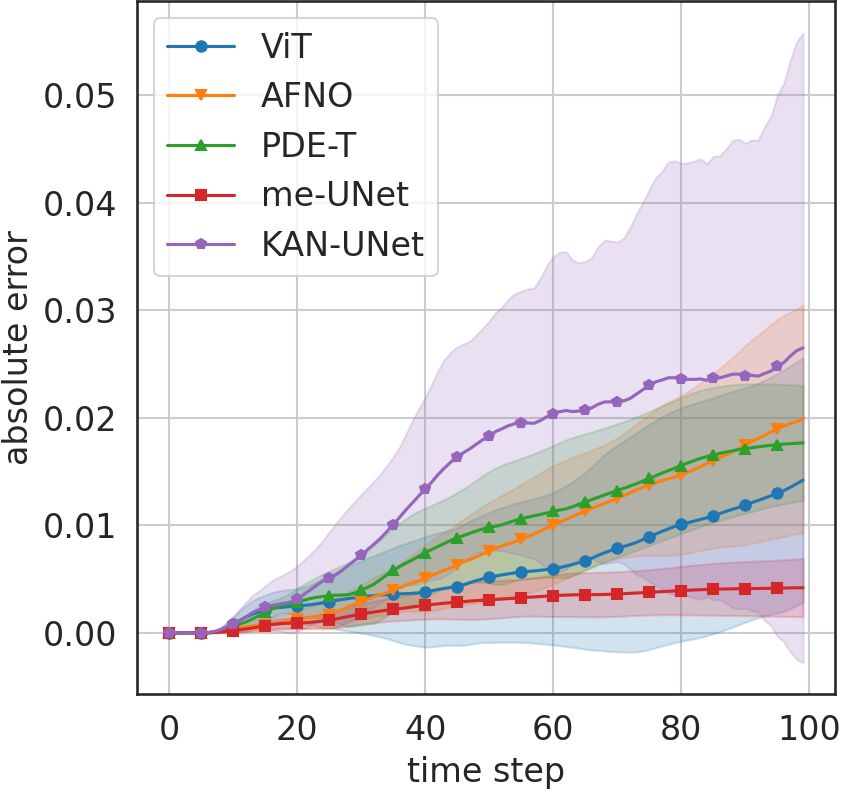}\\
			\bigskip
			\caption{}
		\end{subfigure}
		\centering
		\begin{subfigure}{0.45\textwidth}
			\centering
			\includegraphics[width=0.9\textwidth]{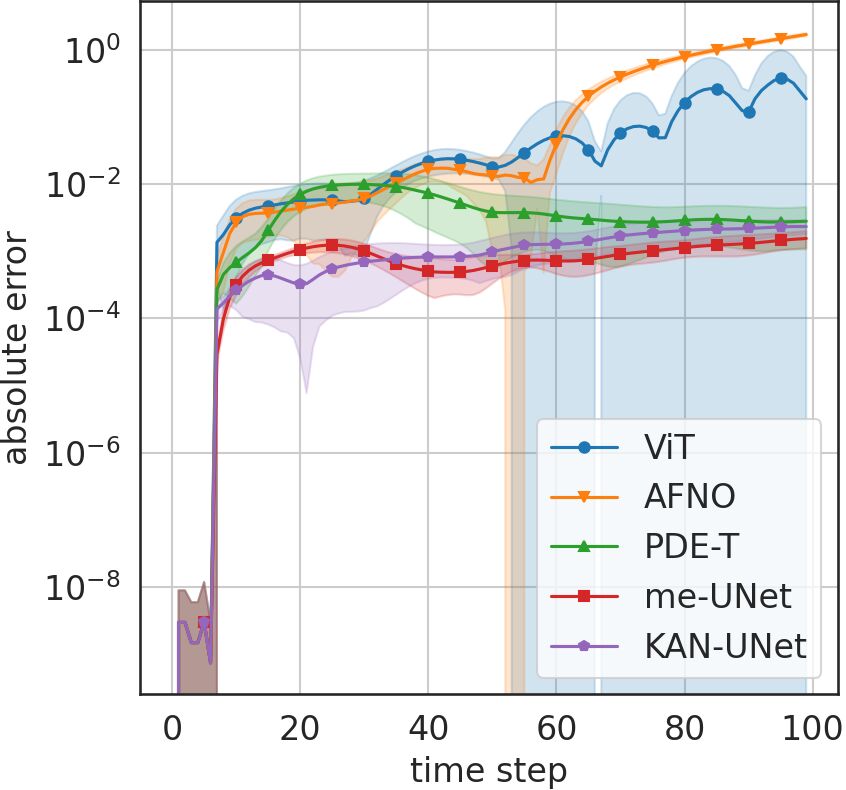}\\
			\bigskip
			\caption{}
		\end{subfigure}
		\caption{Normalized absolute error of physics-based metrics for (a) DS-3a, (b) DS-4, (c) DS-5, and (d) DS-6a.}
		\label{fig:physics_diagnostic}
	\end{figure*}

	In this section we provide additional plots of the physics-based metrics introduced in \cref{sec:metrics} for in-distribution rollouts. 
	Each panel in \cref{fig:physics_diagnostic} shows the normalized absolute error of a conserved or approximately conserved quantity 
	(e.g., mass, energy, total dislocation density) as a function of time for all architectures. 
	Across the four datasets, me-UNet generally yields the smallest deviation from the reference curves over 100 time steps, 
	with KAN-UNet often second-best, and the transformer- and operator-based models showing larger drift.
	
	The accumulated error for DS-4 (\cref{fig:physics_diagnostic}b, \cref{fig:physics_diagnostic_ood}a and \cref{fig:physics_diagnostic_ood}b) 
	may arise from the large range of values between the minimum and maximum of the original dataset (before scaling to $[-1, 1]$). 
	Since the number of loops used as initial values spans a wide range (50--500 loops), the number of simulations may be relatively small 
	compared to the variability in total line length. This can result in a systematic shift of the spatial integral of the state variable 
	(see \cref{fig:physics_diagnostic_ood}a,b). However, the main physical behavior of localization (i.e., the overall patterns) is still captured 
	by the trained models (see \cref{fig:ood_cdd}).
	
	For the other datasets, such as DS-6a (\cref{fig:physics_diagnostic}d, 
	\cref{fig:physics_diagnostic_ood}c and \cref{fig:physics_diagnostic_ood}d), the minimum and maximum values of the original data 
	for all simulations, both in-distribution and \gls{ood}, lie in a much narrower range. In these cases, the physical phenomena are 
	well captured, both in terms of localization and generalization behavior (see \cref{fig:ood_gs_maze_d2}, 
	\cref{fig:physics_diagnostic_ood}c,d).

	\clearpage
	\section{Visualization of physics-based metrics for \gls{ood} initial conditions}
	\label{app:physics_diagnostic_results_ood}
\begin{figure*}[h!tb]
		\centering
		\begin{subfigure}{0.45\textwidth}
			\centering
			\includegraphics[width=0.9\textwidth]{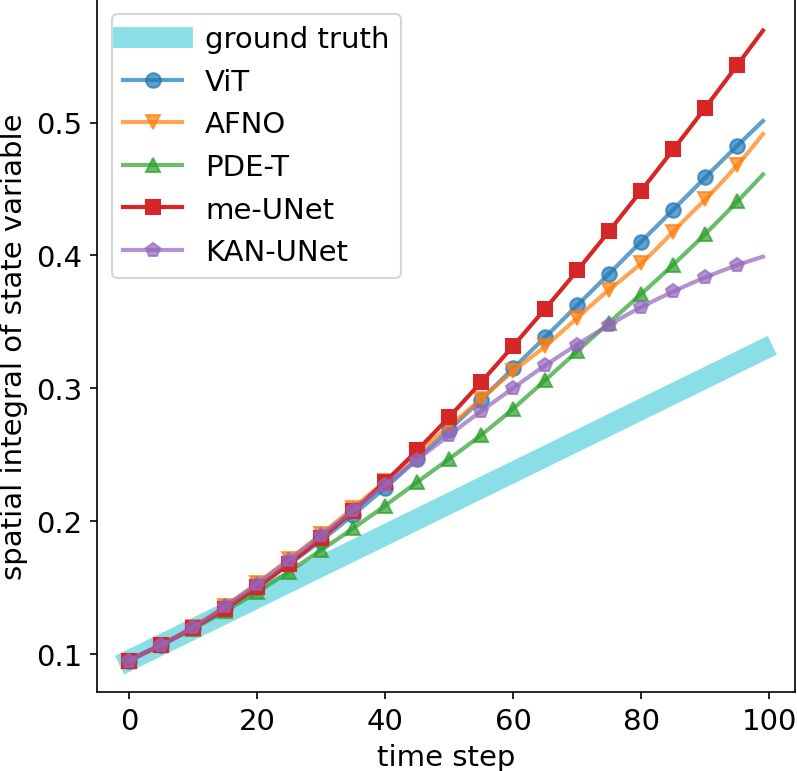}\\
			\bigskip
			\caption{}
		\end{subfigure}
		\centering
		\begin{subfigure}{0.45\textwidth}
			\centering
			\includegraphics[width=0.9\textwidth]{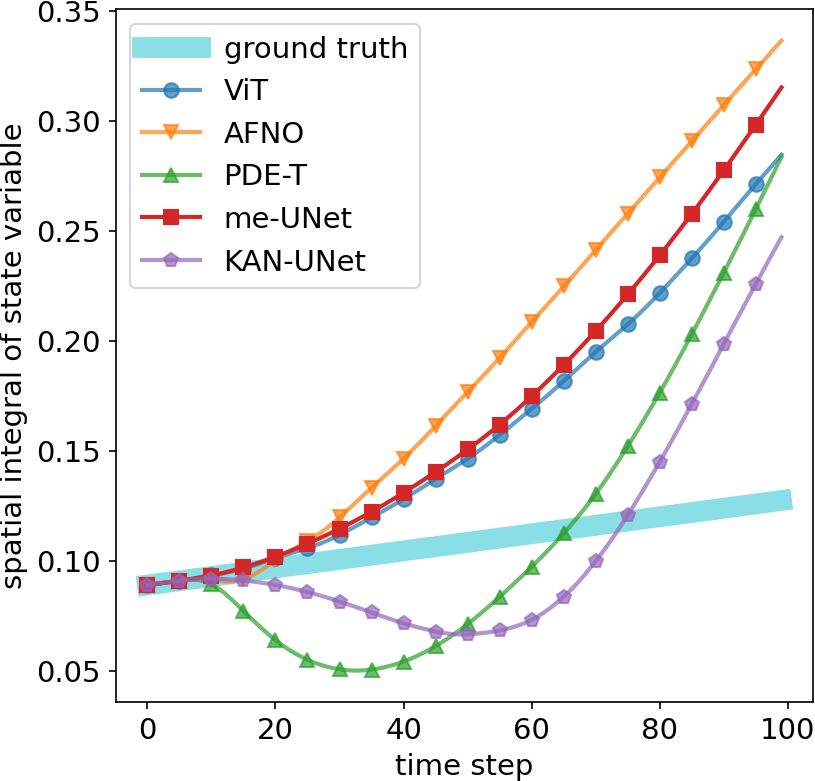}\\
			\bigskip
			\caption{}
		\end{subfigure}
		\centering
		\begin{subfigure}{0.45\textwidth}
			\centering
			\includegraphics[width=0.9\textwidth]{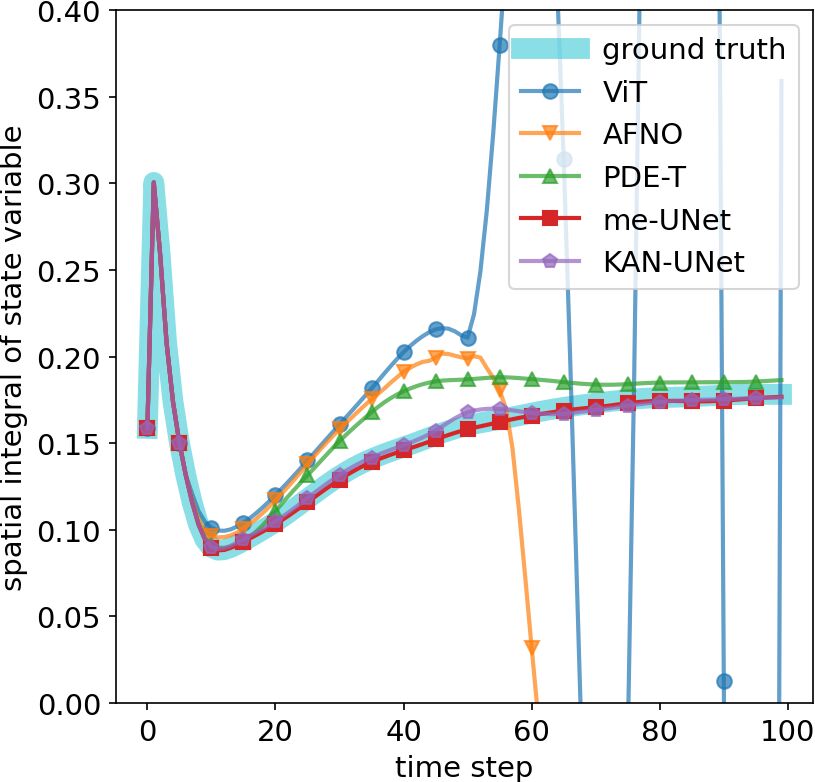}\\
			\bigskip
			\caption{}
		\end{subfigure}
		\centering
		\begin{subfigure}{0.45\textwidth}
			\centering
			\includegraphics[width=0.9\textwidth]{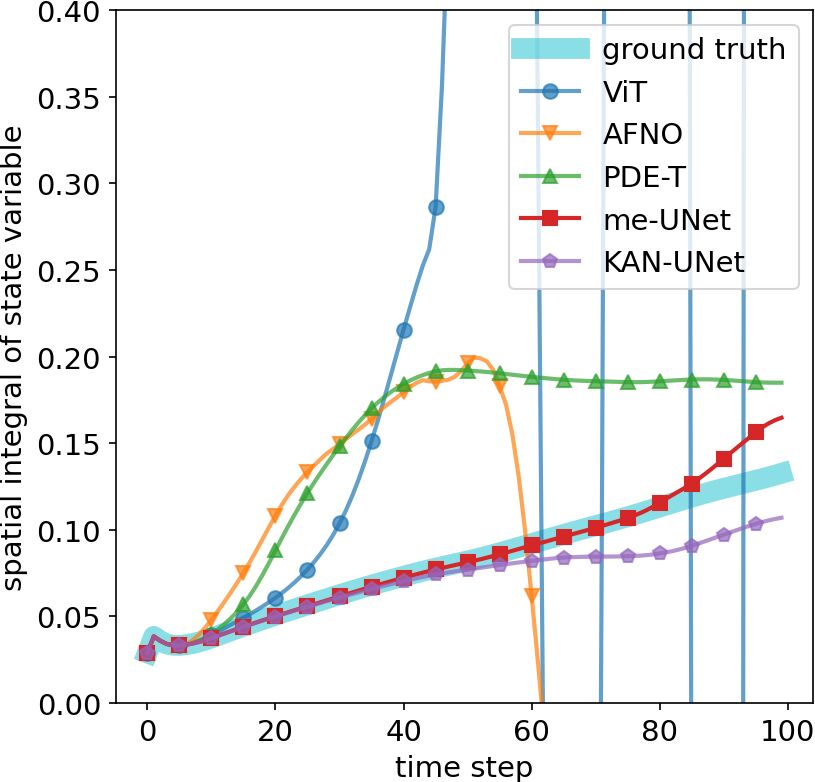}\\
			\bigskip
			\caption{}
		\end{subfigure}
		\caption{Spatial integrals of the monitored quantities for \gls{ood} rollouts on (a) DS-4 with initial few loops, (b) DS-4 with mixed lines and loops, (c) DS-6a with line-like initial conditions, and (d) DS-6a with blob-like initial conditions.}
		\label{fig:physics_diagnostic_ood}
	\end{figure*}

	In this section we show the corresponding physics-based metrics for the \gls{ood} initial-condition experiments described in \cref{sec:results}. 
	Figure~\ref{fig:physics_diagnostic_ood} displays spatial integrals of the monitored quantities for \gls{CDD} and Gray--Scott rollouts 
	under different \gls{ood} initial states. As in the in-distribution case, me-UNet tracks the reference integrals most closely over time, 
	while several baselines exhibit noticeable drift, particularly for the more challenging line- and blob-type initial conditions.

	\clearpage
	\section{Visualization of prediction performance across architectures}
	\label{app:prediction_results}
	This section contains additional qualitative rollouts and metric curves for all datasets, complementing the examples in \cref{sec:results}. 
	Each figure shows, for a fixed dataset, autoregressive predictions of all architectures over 100 time steps together with per-time-step \gls{RMSE} and 
	PSD cosine similarity. These plots illustrate in more detail the trends discussed in the main text: me-UNet and, to a lesser extent, KAN-UNet produce 
	stable pattern evolution with low error and high spectral similarity, whereas several of the heavier architectures either smooth out fine-scale 
	structure or develop high-frequency artifacts over long rollouts.

	\begin{figure*}[htb!]
		\centering
		\tabskip=0pt
		\valign{#\cr
			\hbox{%
				\begin{subfigure}[b]{.69\textwidth}
					\centering
					\includegraphics[width=1\textwidth]{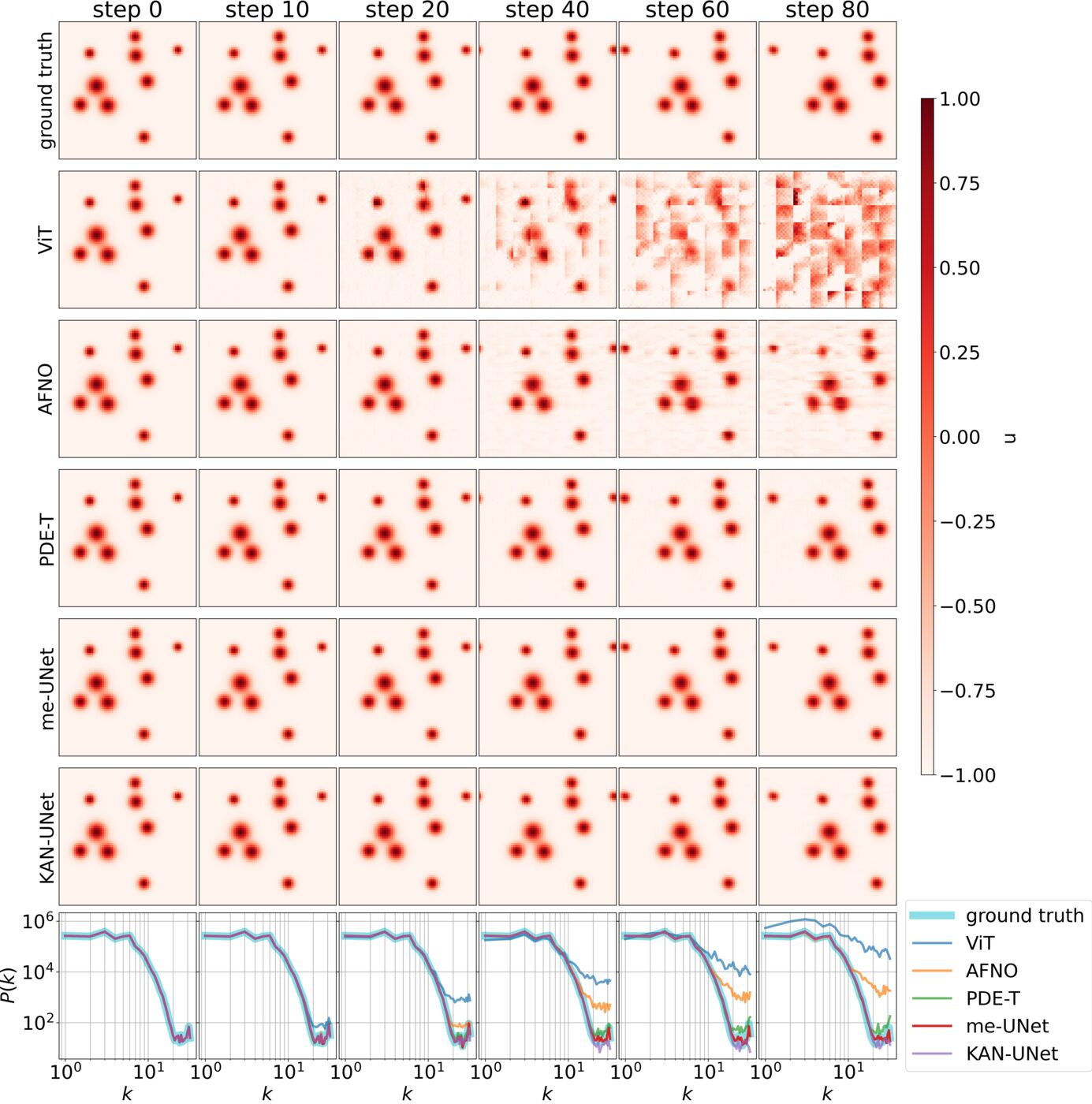}
					\caption{}
				\end{subfigure}%
			}\cr
			\hbox{%
				\begin{subfigure}{.3\textwidth}
					\centering
					\includegraphics[width=0.9\textwidth]{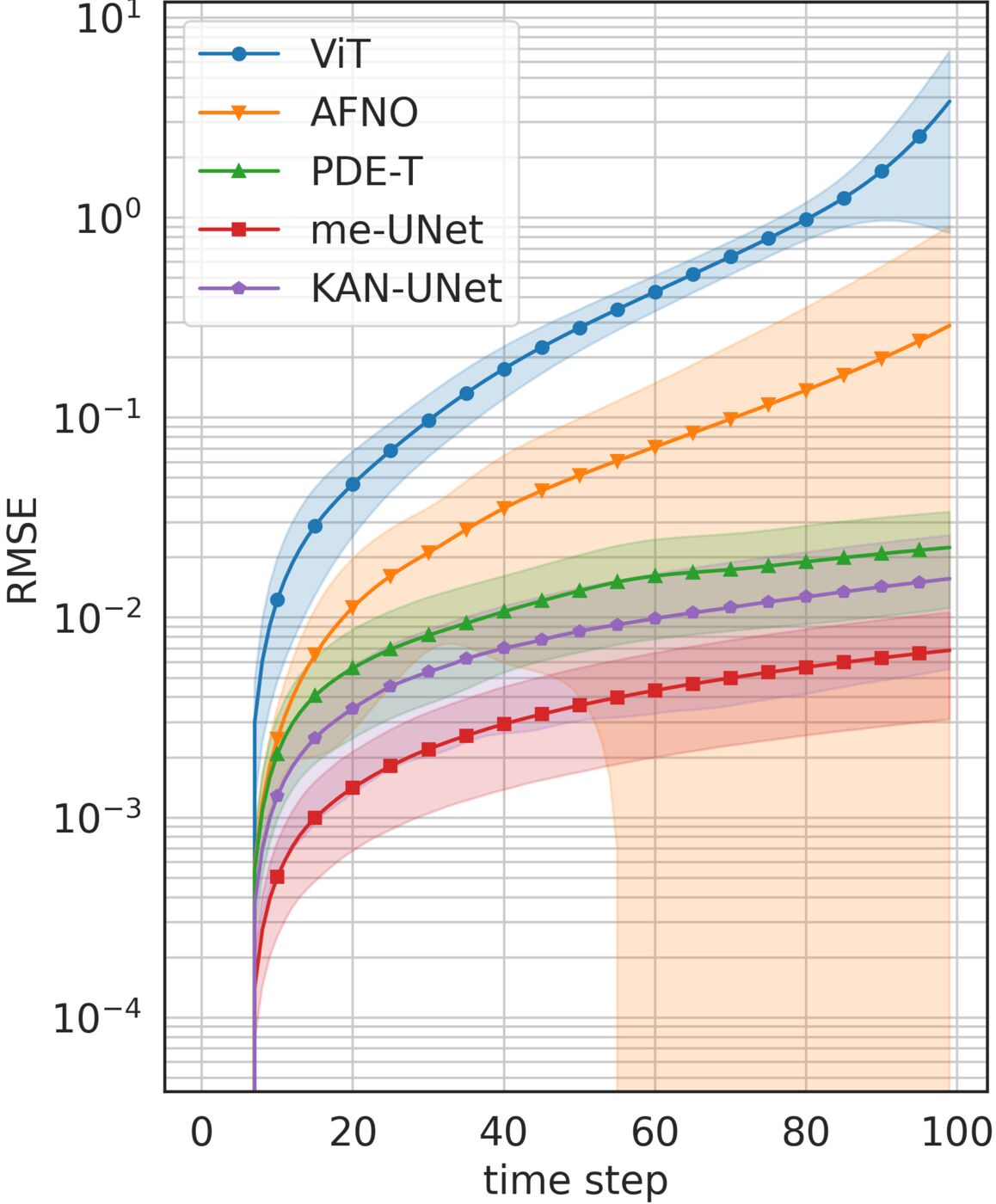}
					\caption{}
				\end{subfigure}%
			}\vfill
			\hbox{%
				\begin{subfigure}{.3\textwidth}
					\centering
					\includegraphics[width=0.9\textwidth]{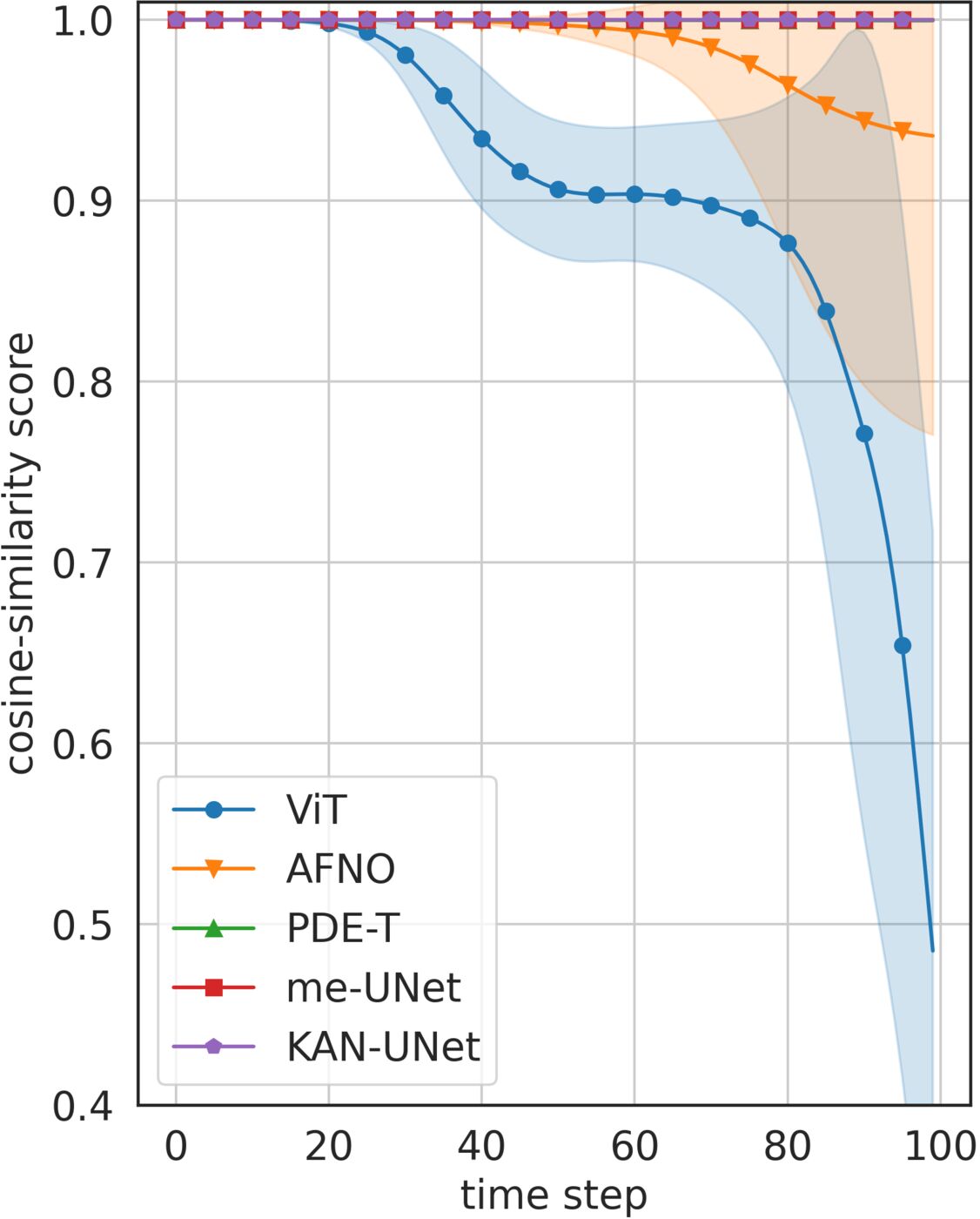}
					\caption{}
				\end{subfigure}%
			}\cr
		}
		\caption{Visualization of the performance of the different neural networks on the DS-1 dataset: (a) autoregressive prediction results; (b) \gls{RMSE}; and (c) cosine similarity of the PSD curves.}
		\label{fig:id_adv}
	\end{figure*}
	
	\begin{figure*}[htb!]
		\centering
		\tabskip=0pt
		\valign{#\cr
			\hbox{%
				\begin{subfigure}[b]{.69\textwidth}
					\centering
					\includegraphics[width=1\textwidth]{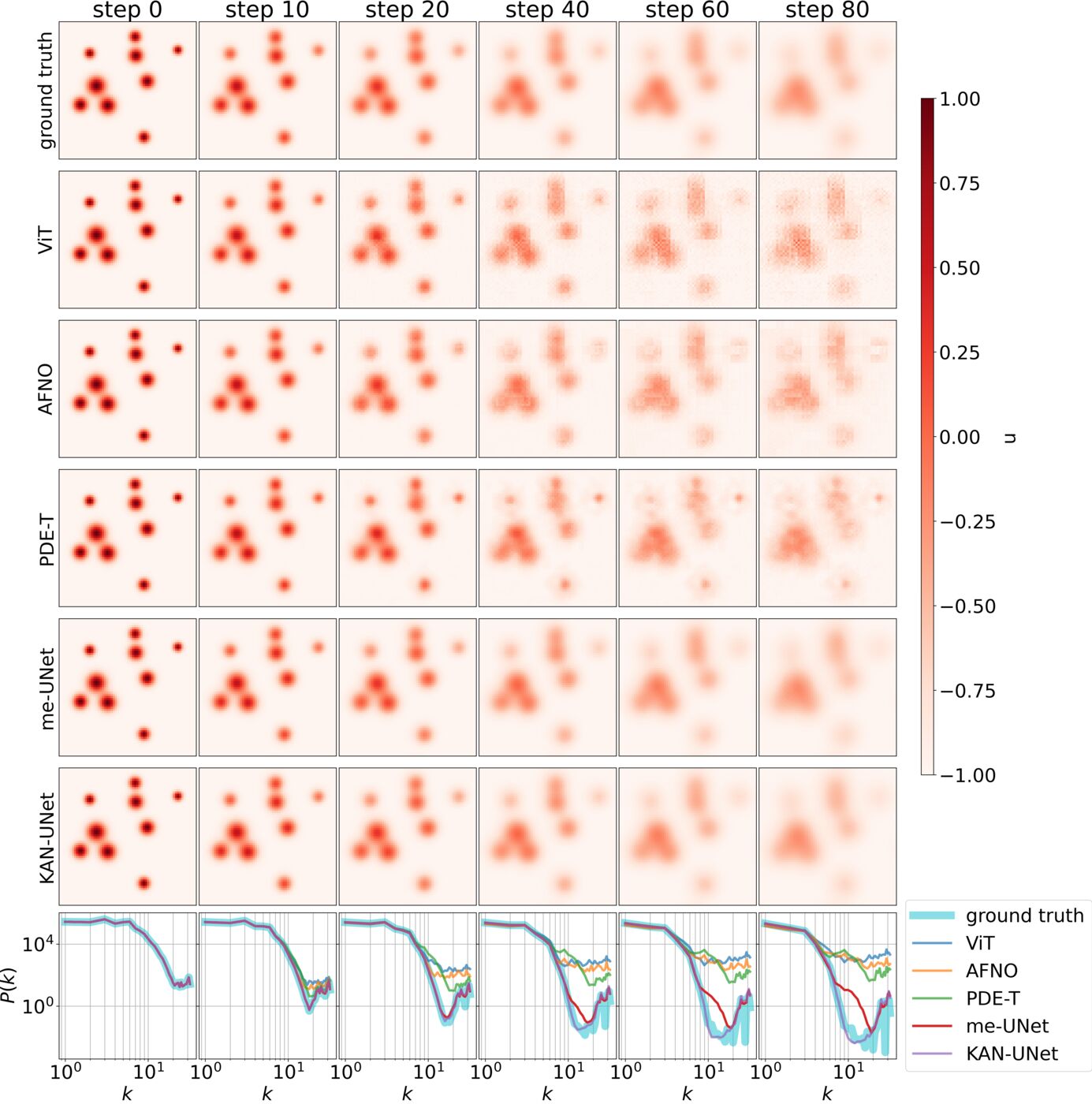}
					\caption{}
				\end{subfigure}%
			}\cr
			\hbox{%
				\begin{subfigure}{.3\textwidth}
					\centering
					\includegraphics[width=0.9\textwidth]{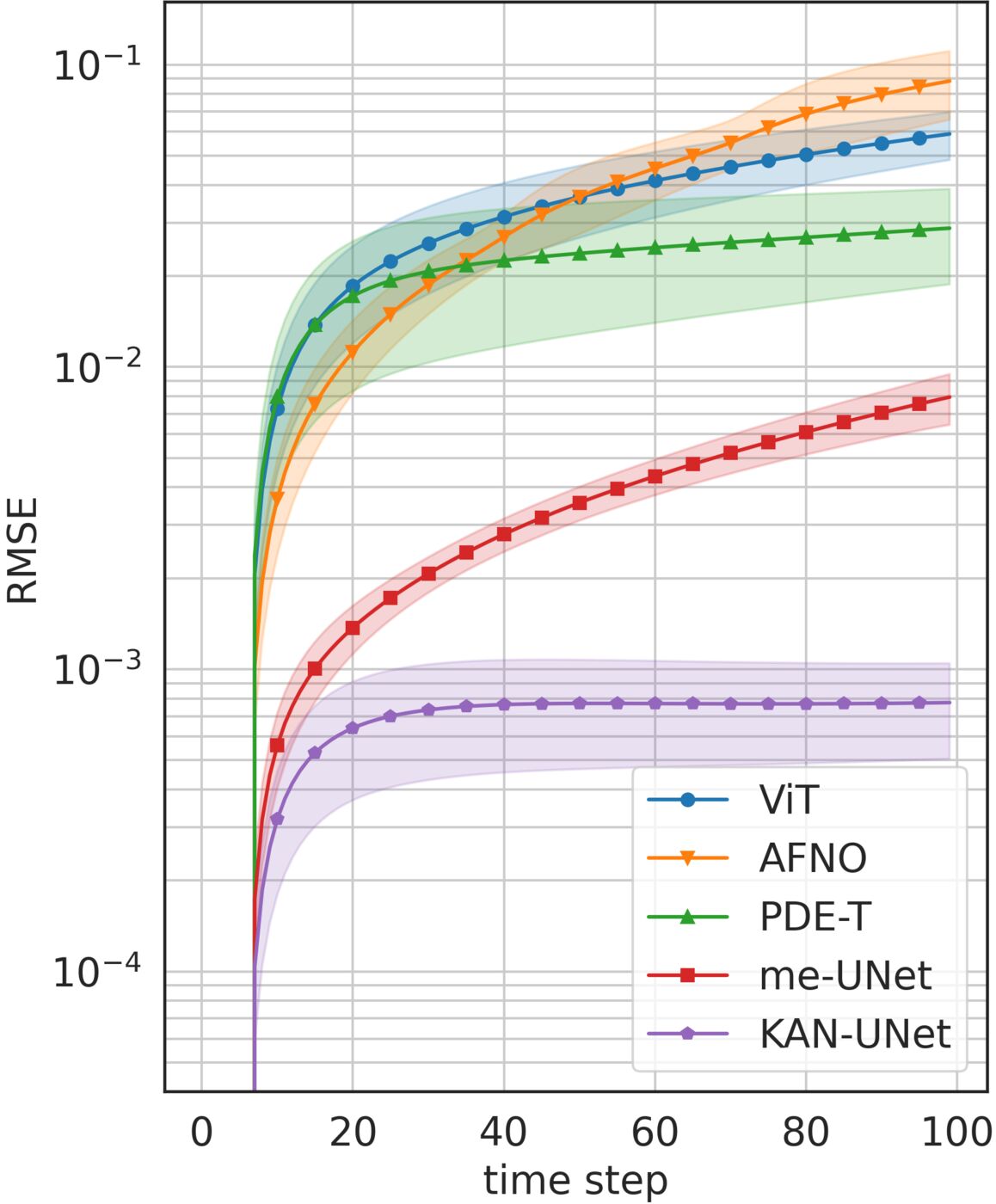}
					\caption{}
				\end{subfigure}%
			}\vfill
			\hbox{%
				\begin{subfigure}{.3\textwidth}
					\centering
					\includegraphics[width=0.9\textwidth]{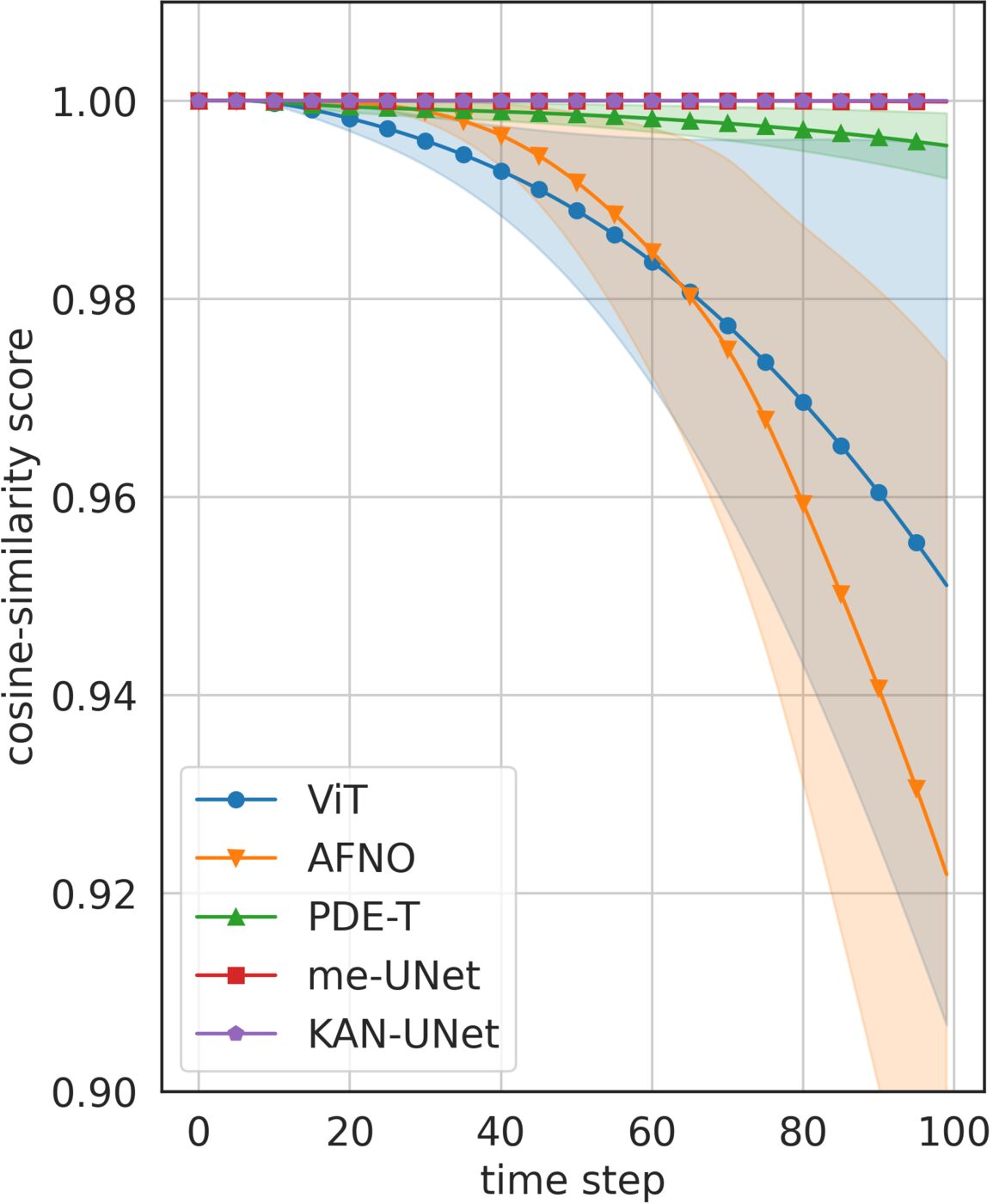}
					\caption{}
				\end{subfigure}%
			}\cr
		}
		\caption{Visualization of the performance of the different neural networks on the DS-2 dataset: (a) autoregressive prediction results; (b) \gls{RMSE}; and (c) cosine similarity of the PSD curves.}
		\label{fig:id_diff}
	\end{figure*}
	
	\begin{figure*}[htb!]
		\centering
		\tabskip=0pt
		\valign{#\cr
			\hbox{%
				\begin{subfigure}[b]{.69\textwidth}
					\centering
					\includegraphics[width=1\textwidth]{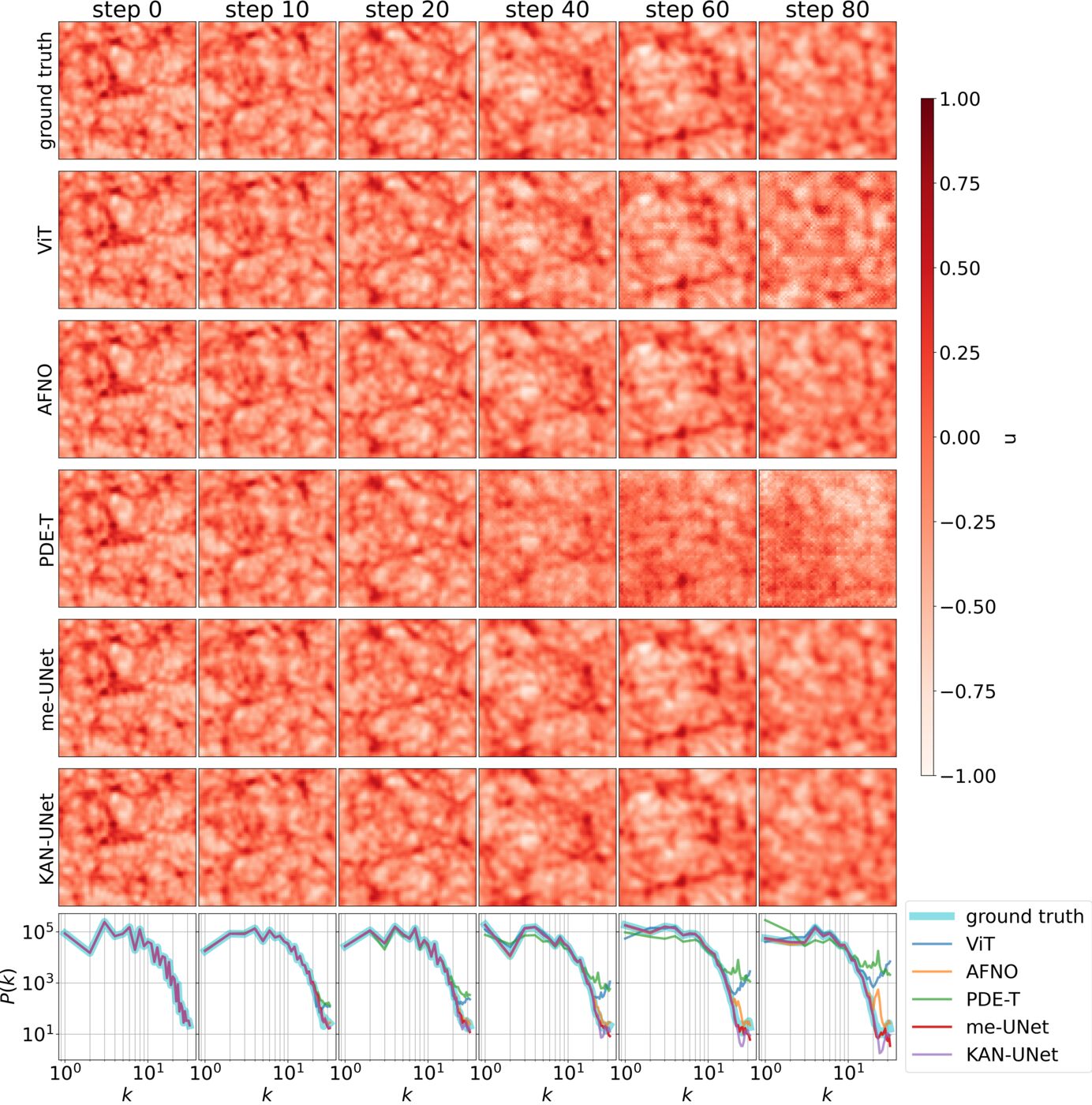}
					\caption{}
				\end{subfigure}%
			}\cr
			\hbox{%
				\begin{subfigure}{.3\textwidth}
					\centering
					\includegraphics[width=0.9\textwidth]{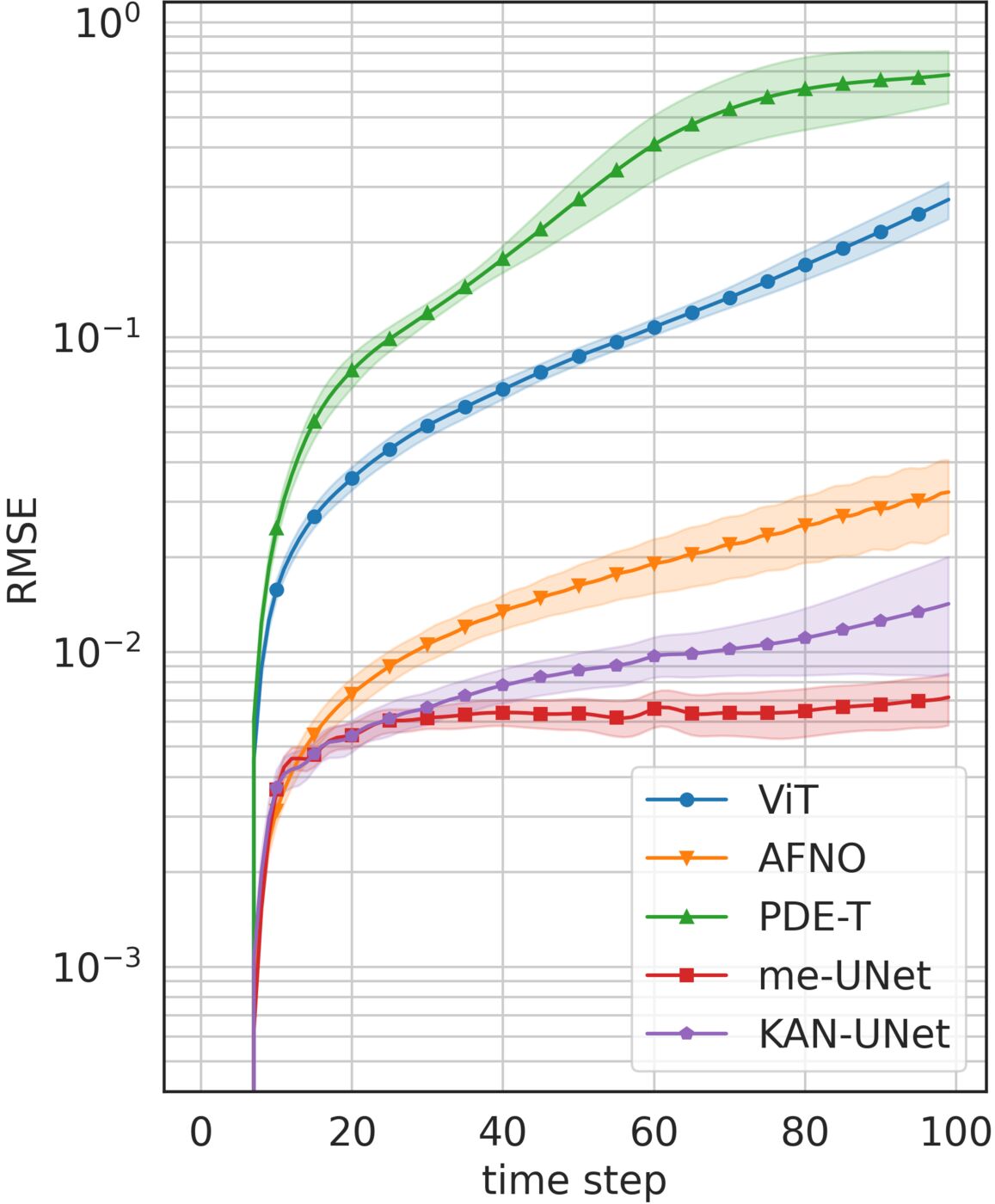}
					\caption{}
				\end{subfigure}%
			}\vfill
			\hbox{%
				\begin{subfigure}{.3\textwidth}
					\centering
					\includegraphics[width=0.9\textwidth]{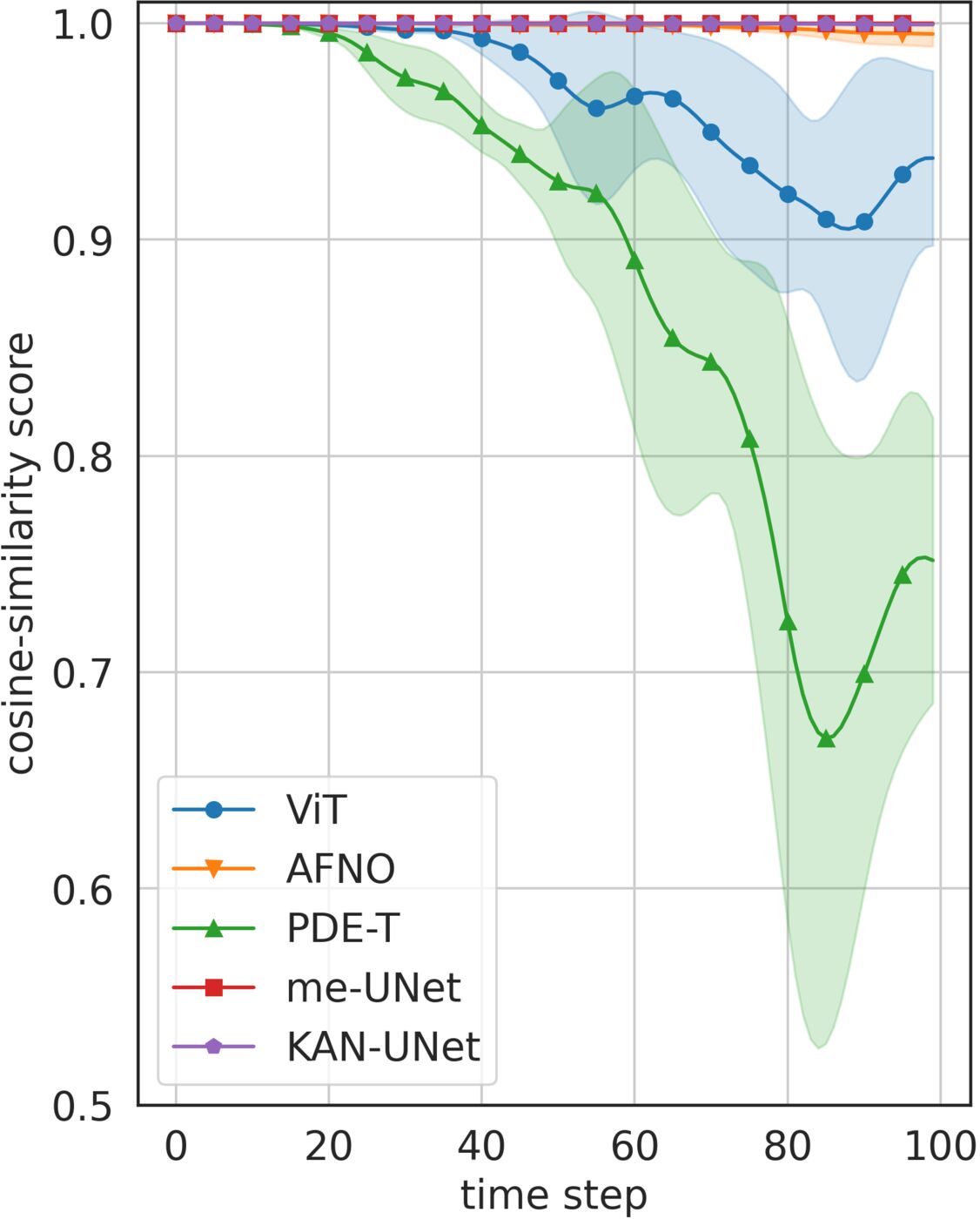}
					\caption{}
				\end{subfigure}%
			}\cr
		}
		\caption{Visualization of the performance of the different neural networks on the DS-3a dataset: (a) autoregressive prediction results; (b) \gls{RMSE}; and (c) cosine similarity of the PSD curves.}
		\label{fig:id_cdd_adv}
	\end{figure*}
	
	\begin{figure*}[htb!]
		\centering
		\tabskip=0pt
		\valign{#\cr
			\hbox{%
				\begin{subfigure}[b]{.69\textwidth}
					\centering
					\includegraphics[width=1\textwidth]{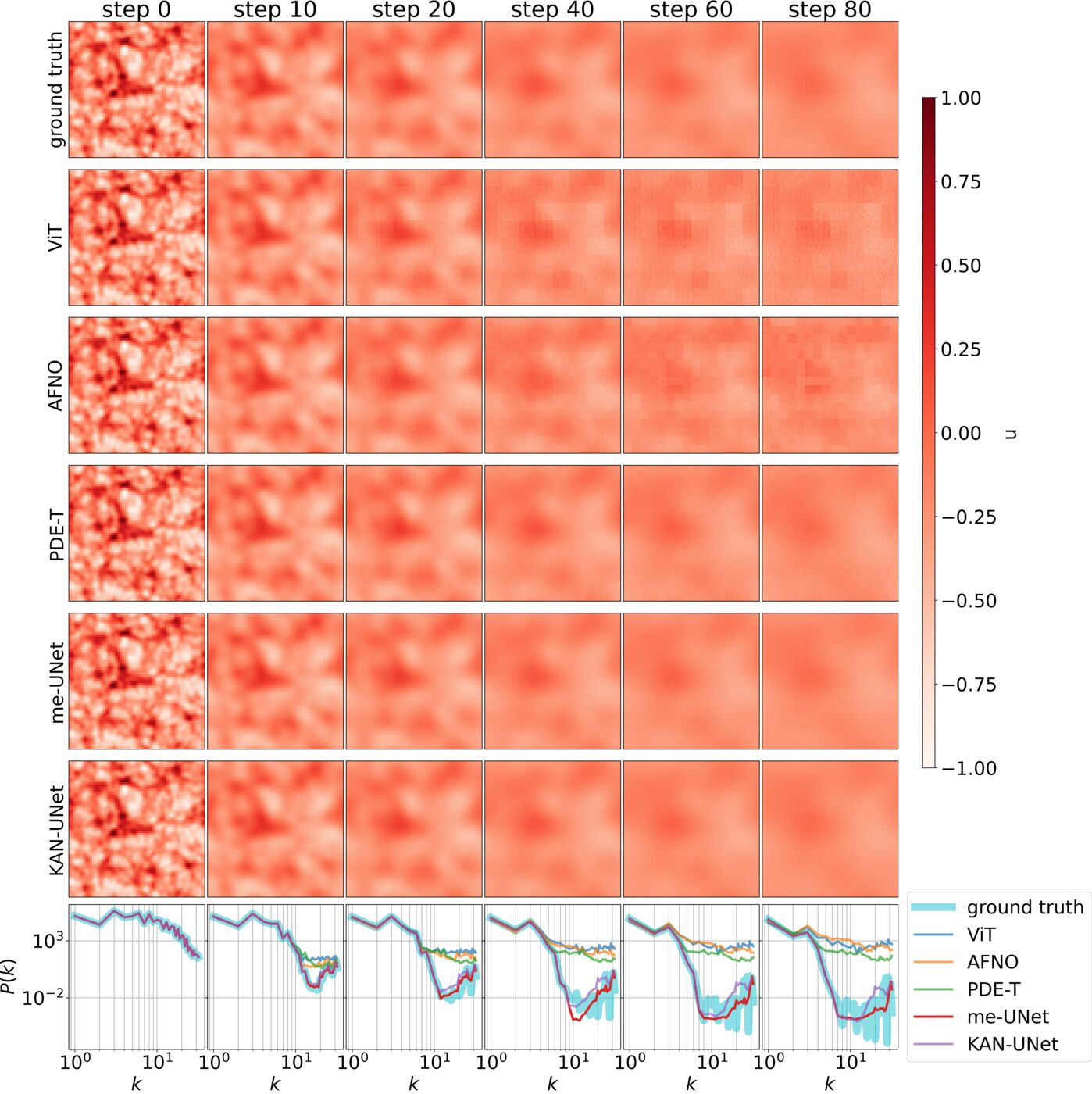}
					\caption{}
				\end{subfigure}%
			}\cr
			\hbox{%
				\begin{subfigure}{.3\textwidth}
					\centering
					\includegraphics[width=0.9\textwidth]{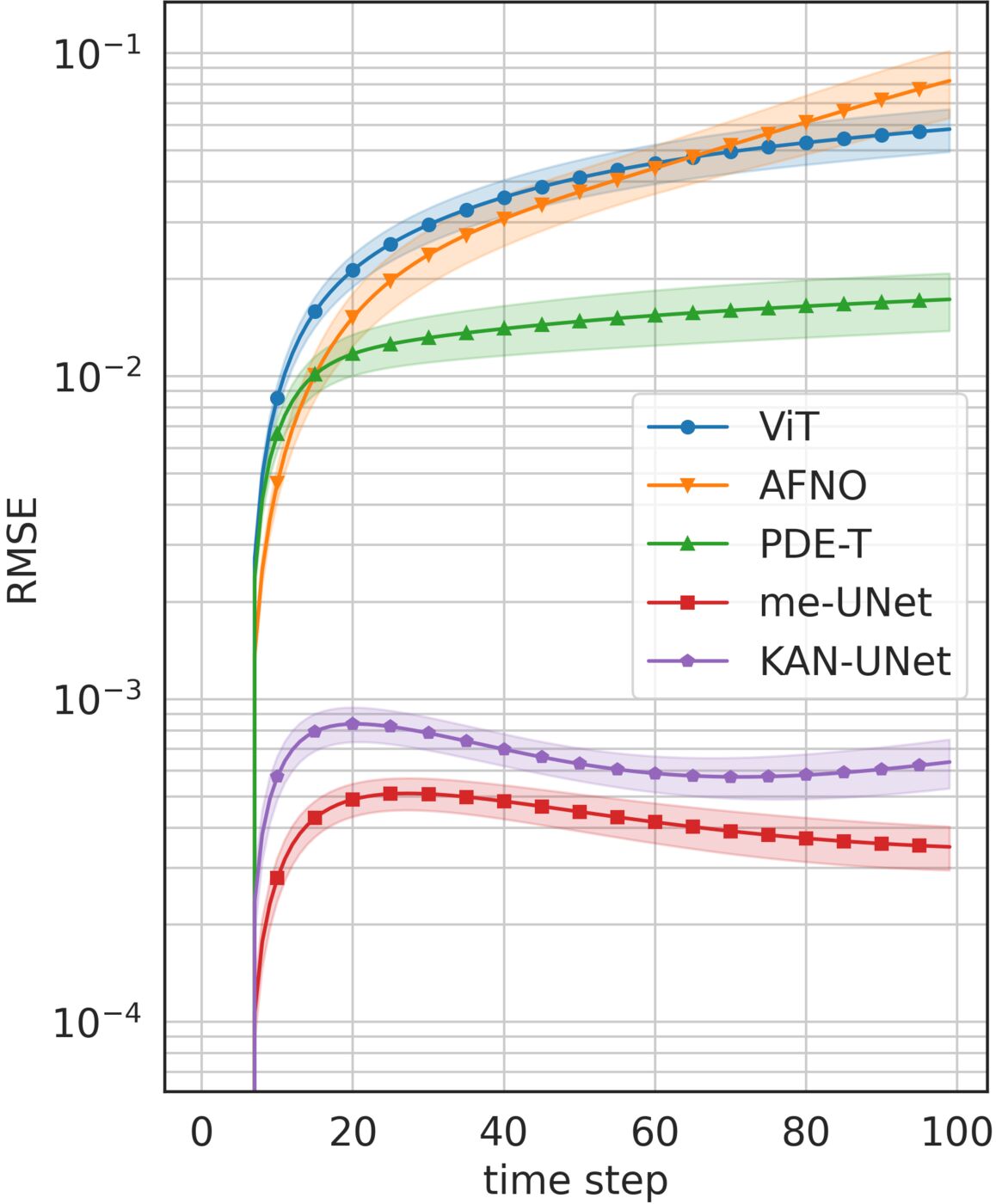}
					\caption{}
				\end{subfigure}%
			}\vfill
			\hbox{%
				\begin{subfigure}{.3\textwidth}
					\centering
					\includegraphics[width=0.9\textwidth]{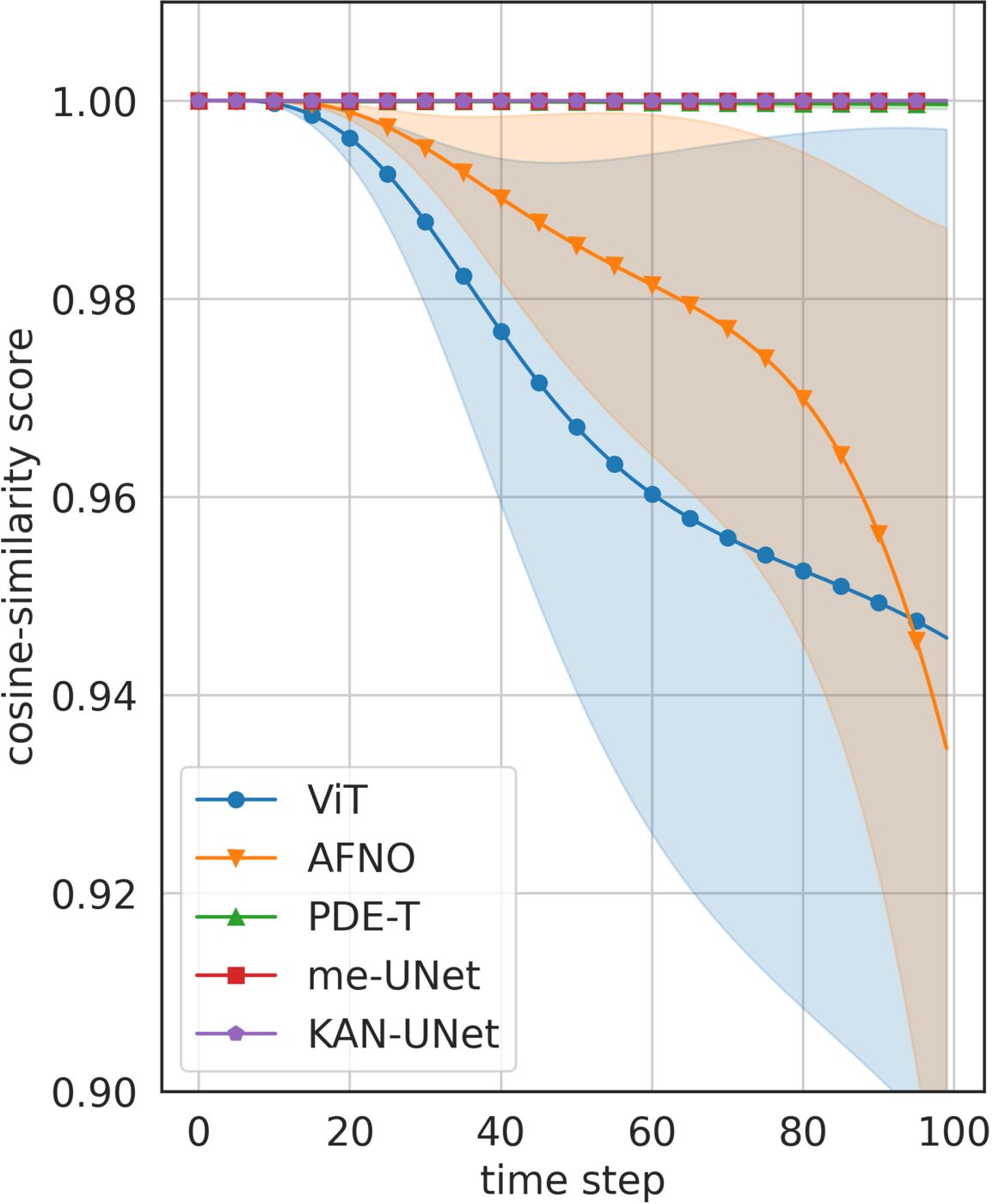}
					\caption{}
				\end{subfigure}%
			}\cr
		}
		\caption{Visualization of the performance of the different neural networks on the DS-3b dataset: (a) autoregressive prediction results; (b) \gls{RMSE}; and (c) cosine similarity of the PSD curves.}
		\label{fig:id_cdd_diff}
	\end{figure*}
	
	\begin{figure*}[htb!]
		\centering
		\tabskip=0pt
		\valign{#\cr
			\hbox{%
				\begin{subfigure}[b]{.69\textwidth}
					\centering
					\includegraphics[width=1\textwidth]{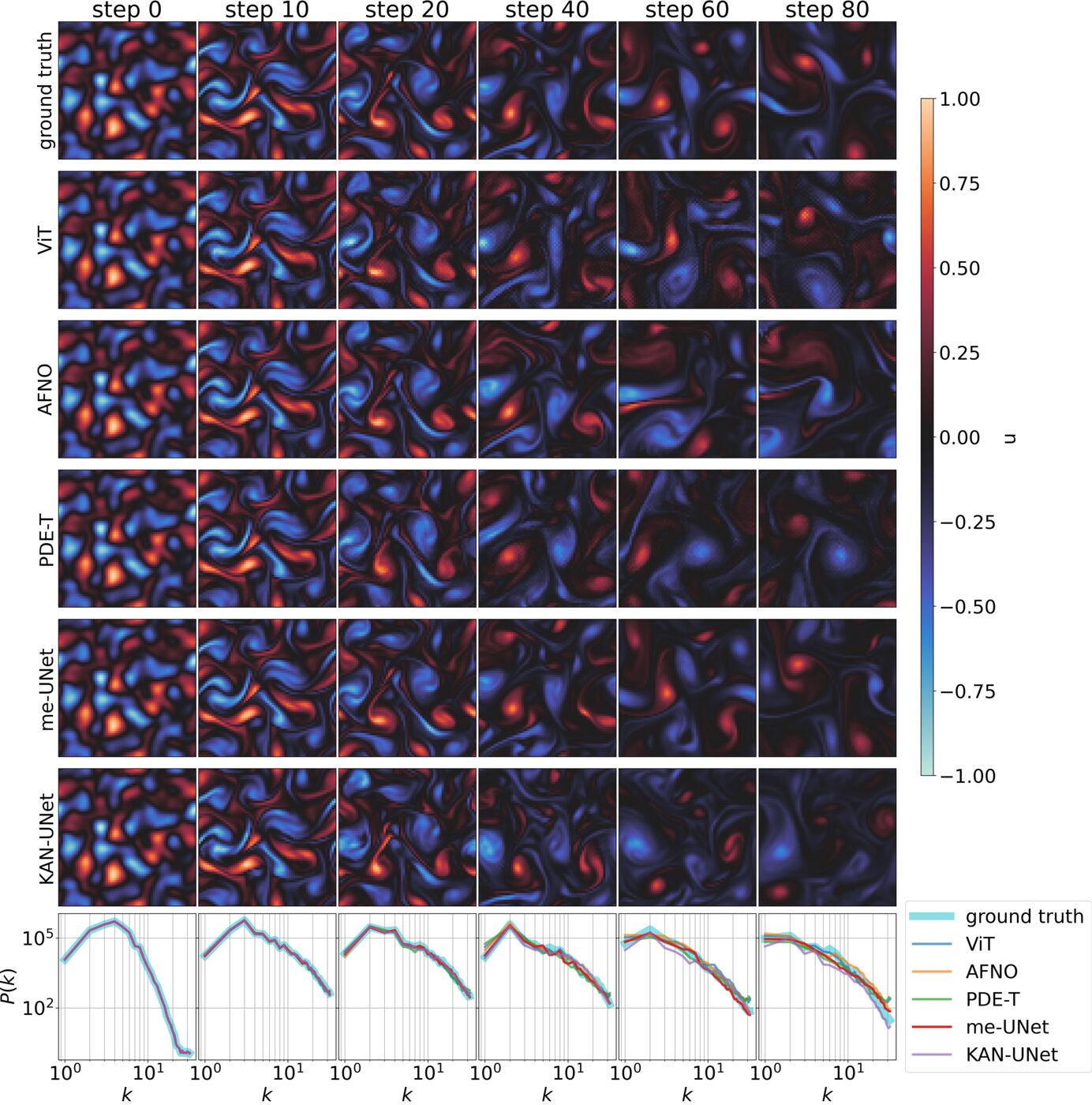}
					\caption{}
				\end{subfigure}%
			}\cr
			\hbox{%
				\begin{subfigure}{.3\textwidth}
					\centering
					\includegraphics[width=0.9\textwidth]{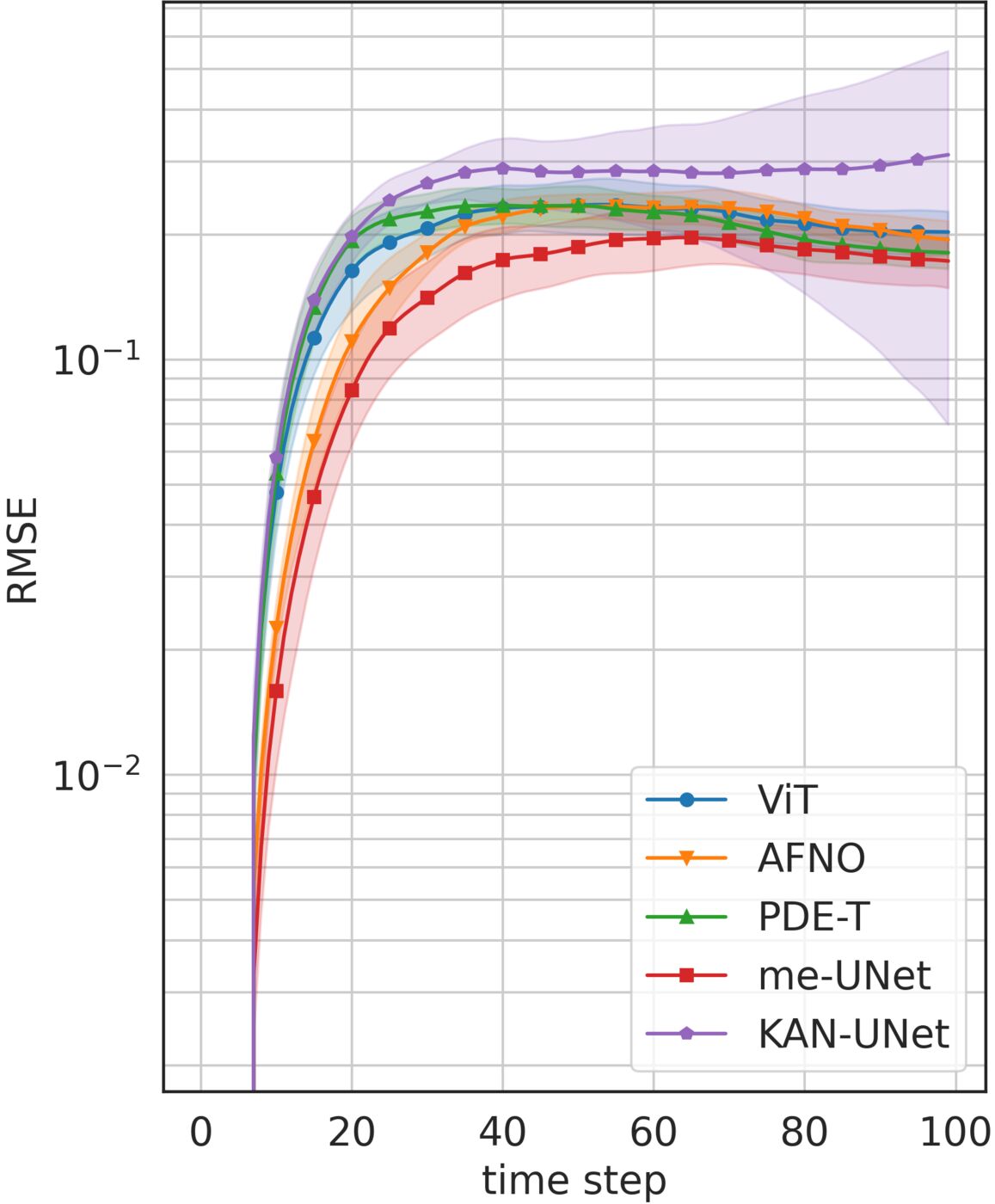}
					\caption{}
				\end{subfigure}%
			}\vfill
			\hbox{%
				\begin{subfigure}{.3\textwidth}
					\centering
					\includegraphics[width=0.9\textwidth]{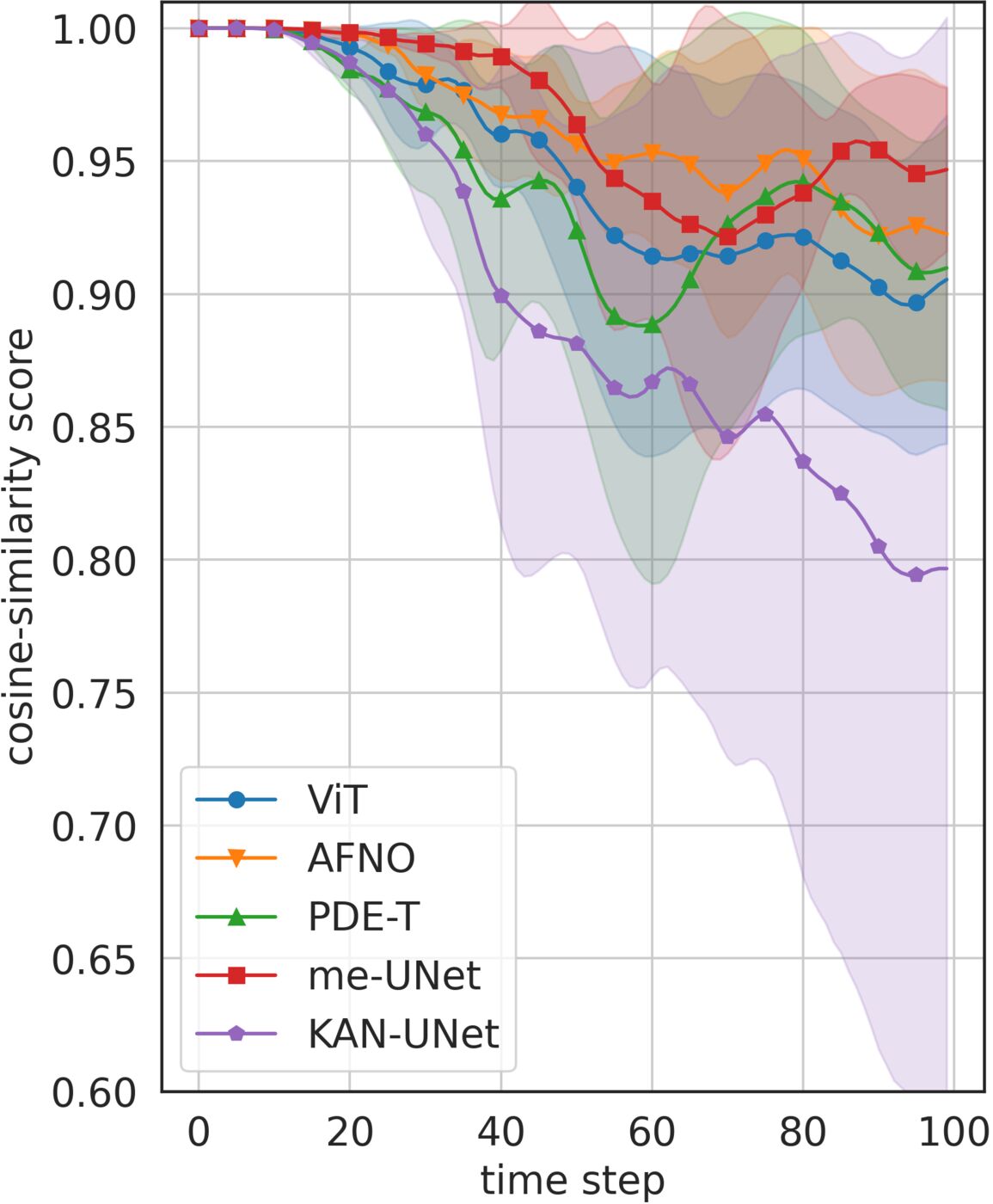}
					\caption{}
				\end{subfigure}%
			}\cr
		}
		\caption{Visualization of the performance of the different neural networks on the DS-5 dataset: (a) autoregressive prediction results; (b) \gls{RMSE}; and (c) cosine similarity of the PSD curves.}
		\label{fig:id_cfd}
	\end{figure*}
	
	\begin{figure*}[htb!]
		\centering
		\tabskip=0pt
		\valign{#\cr
			\hbox{%
				\begin{subfigure}[b]{.69\textwidth}
					\centering
					\includegraphics[width=1\textwidth]{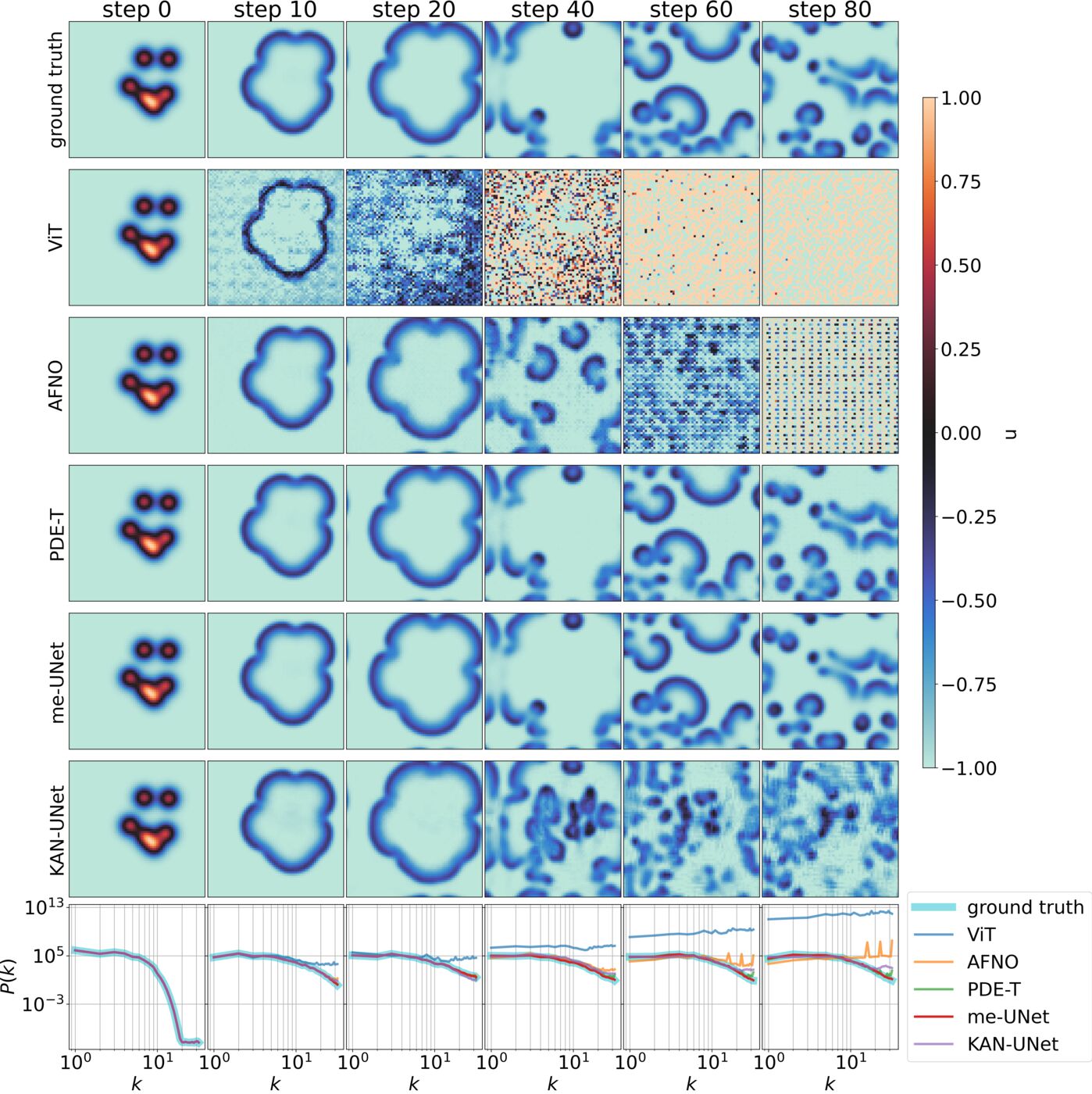}
					\caption{}
				\end{subfigure}%
			}\cr
			\hbox{%
				\begin{subfigure}{.3\textwidth}
					\centering
					\includegraphics[width=0.9\textwidth]{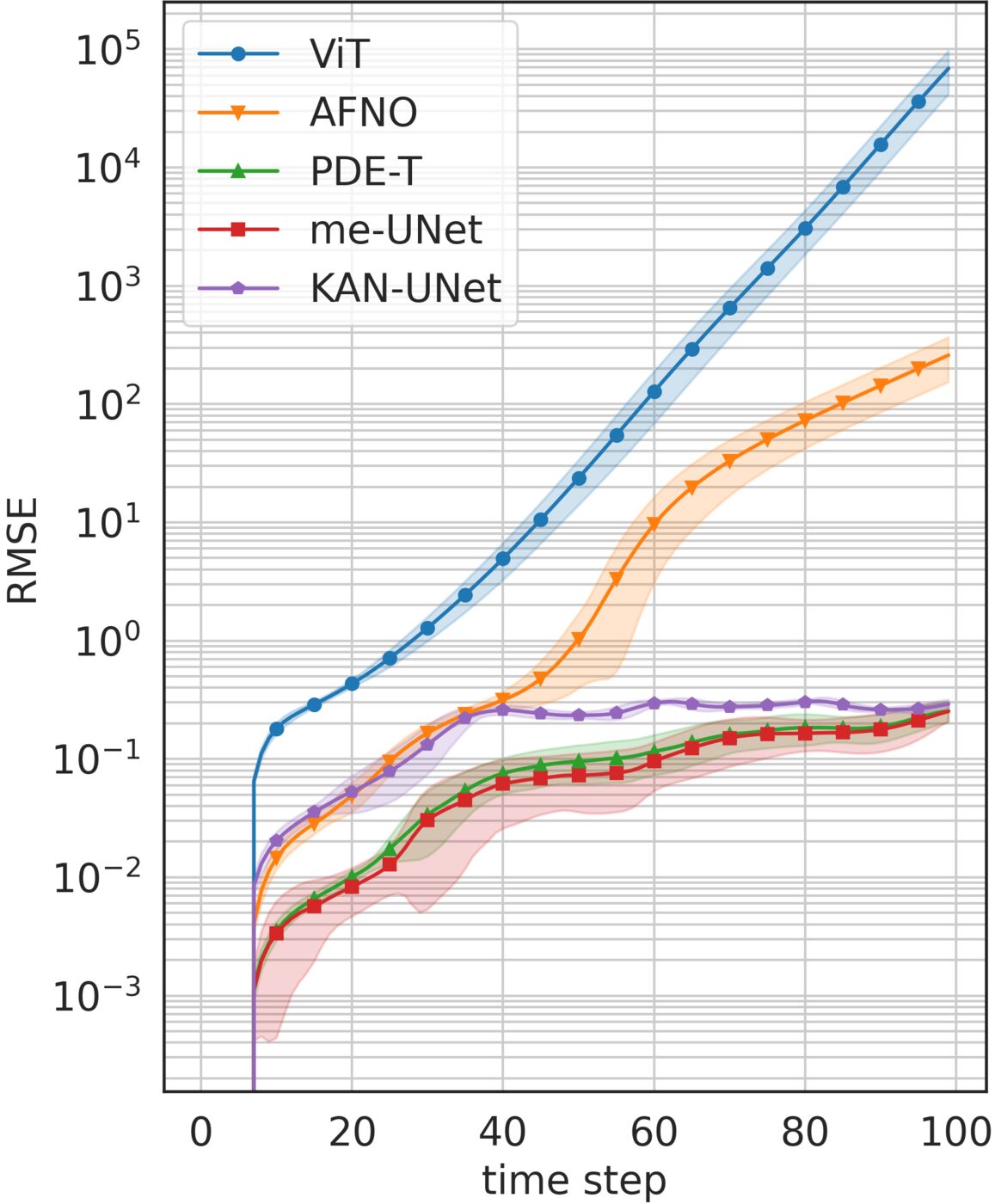}
					\caption{}
				\end{subfigure}%
			}\vfill
			\hbox{%
				\begin{subfigure}{.3\textwidth}
					\centering
					\includegraphics[width=0.9\textwidth]{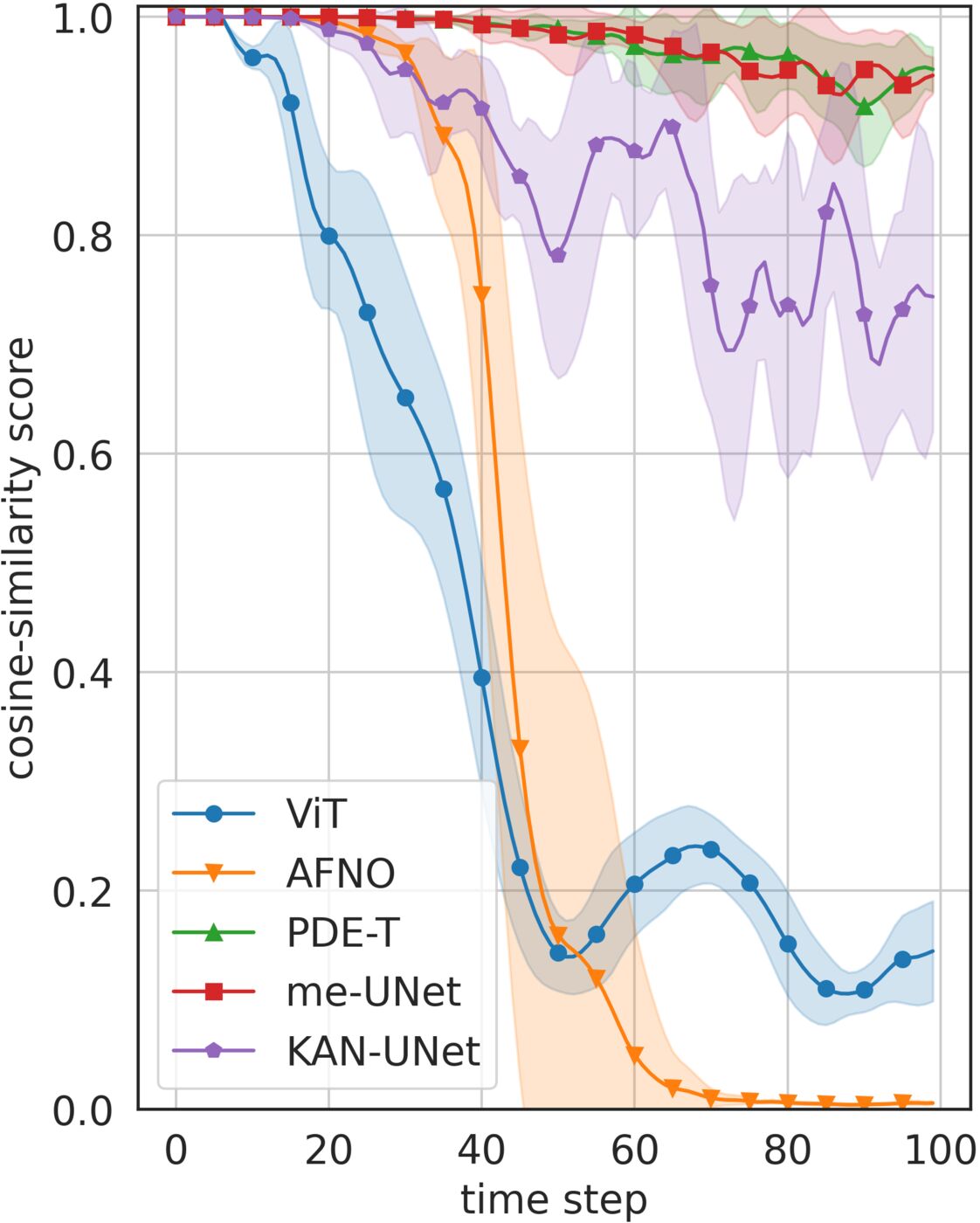}
					\caption{}
				\end{subfigure}%
			}\cr
		}
		\caption{Visualization of the performance of the different neural networks on the DS-6b dataset: (a) autoregressive prediction results; (b) \gls{RMSE}; and (c) cosine similarity of the PSD curves.}
		\label{fig:id_gs_alpha_v2}
	\end{figure*}
	
	\begin{figure*}[htb!]
		\centering
		\tabskip=0pt
		\valign{#\cr
			\hbox{%
				\begin{subfigure}[b]{.69\textwidth}
					\centering
					\includegraphics[width=1\textwidth]{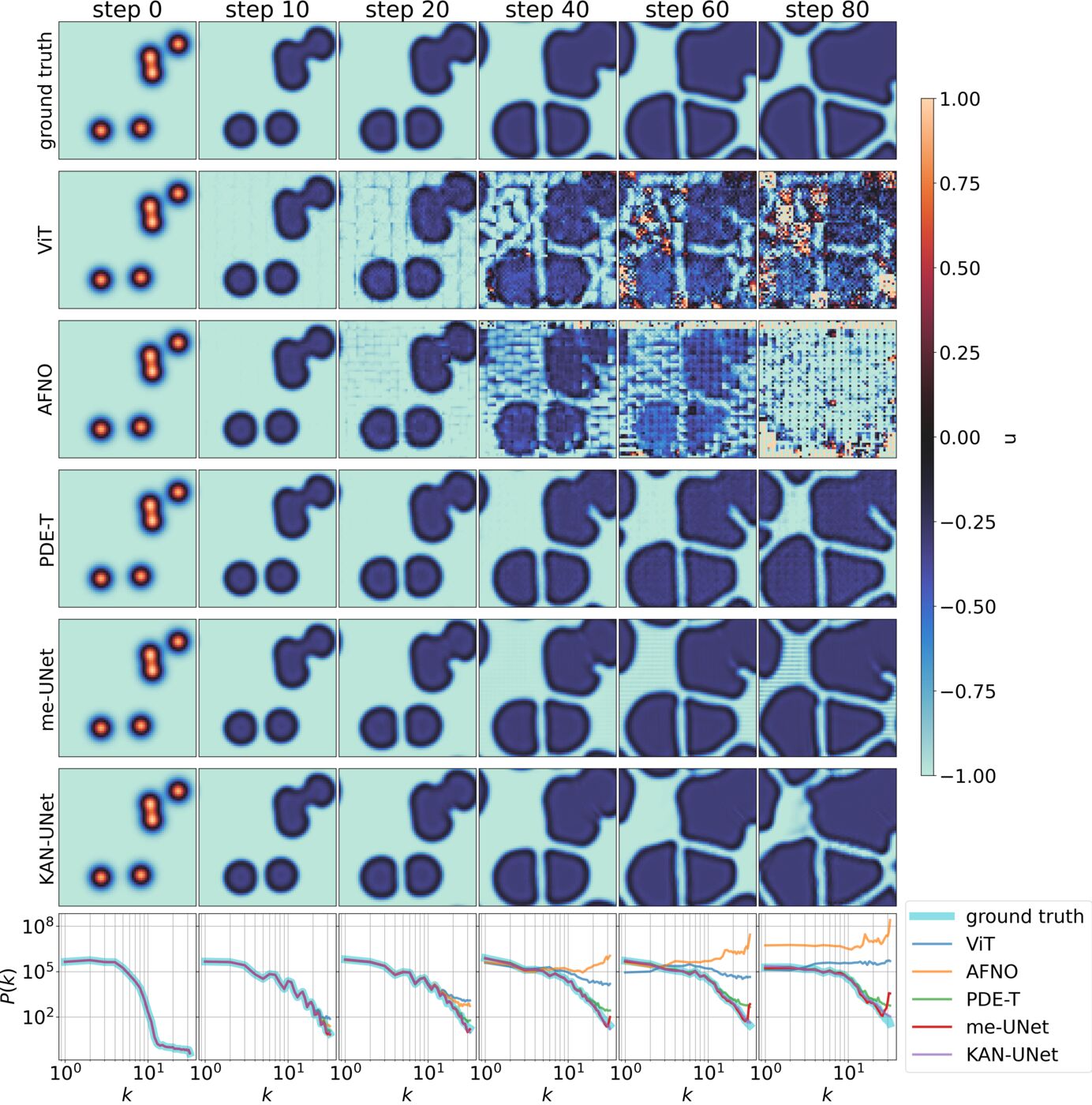}
					\caption{}
				\end{subfigure}%
			}\cr
			\hbox{%
				\begin{subfigure}{.3\textwidth}
					\centering
					\includegraphics[width=0.9\textwidth]{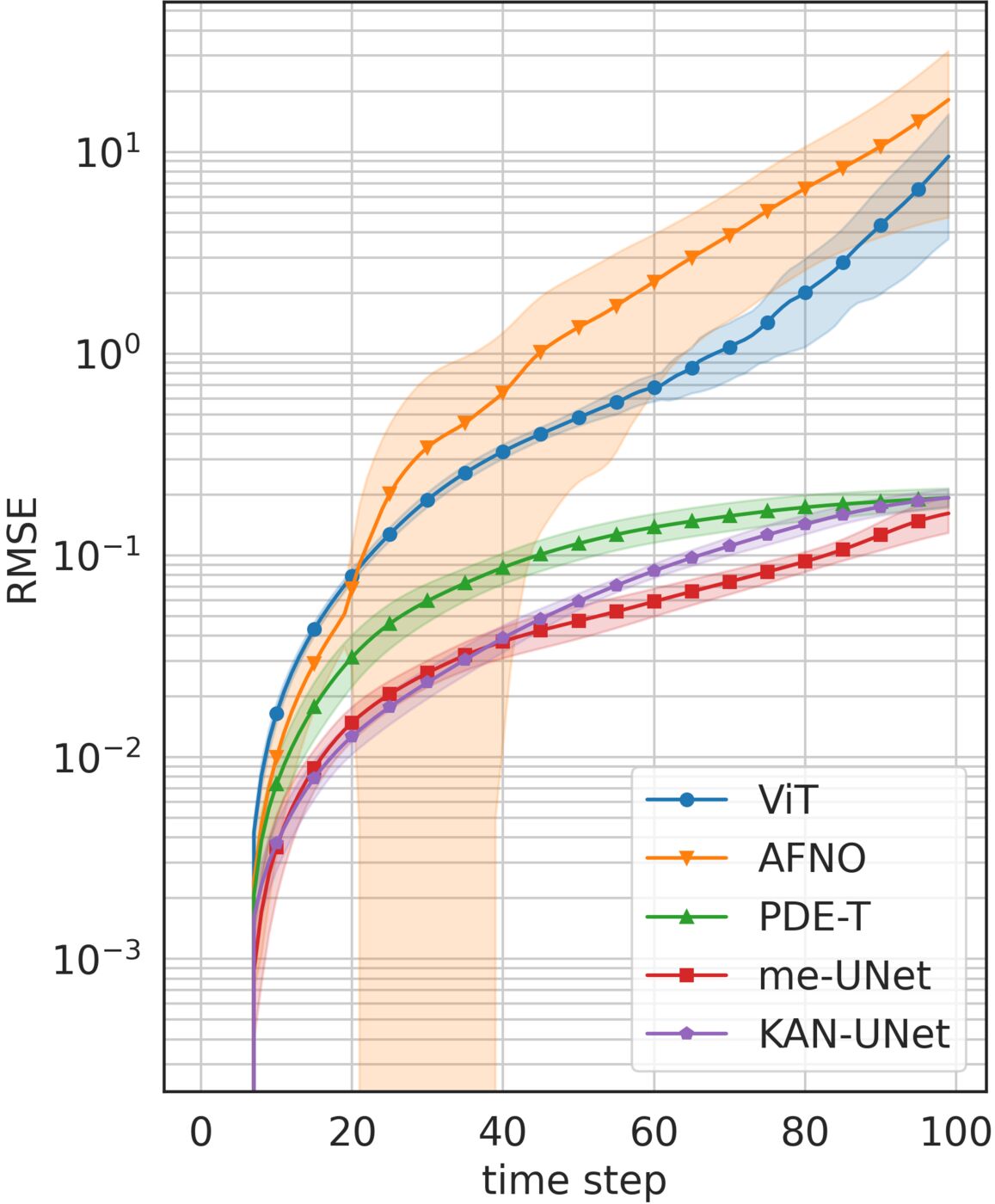}
					\caption{}
				\end{subfigure}%
			}\vfill
			\hbox{%
				\begin{subfigure}{.3\textwidth}
					\centering
					\includegraphics[width=0.9\textwidth]{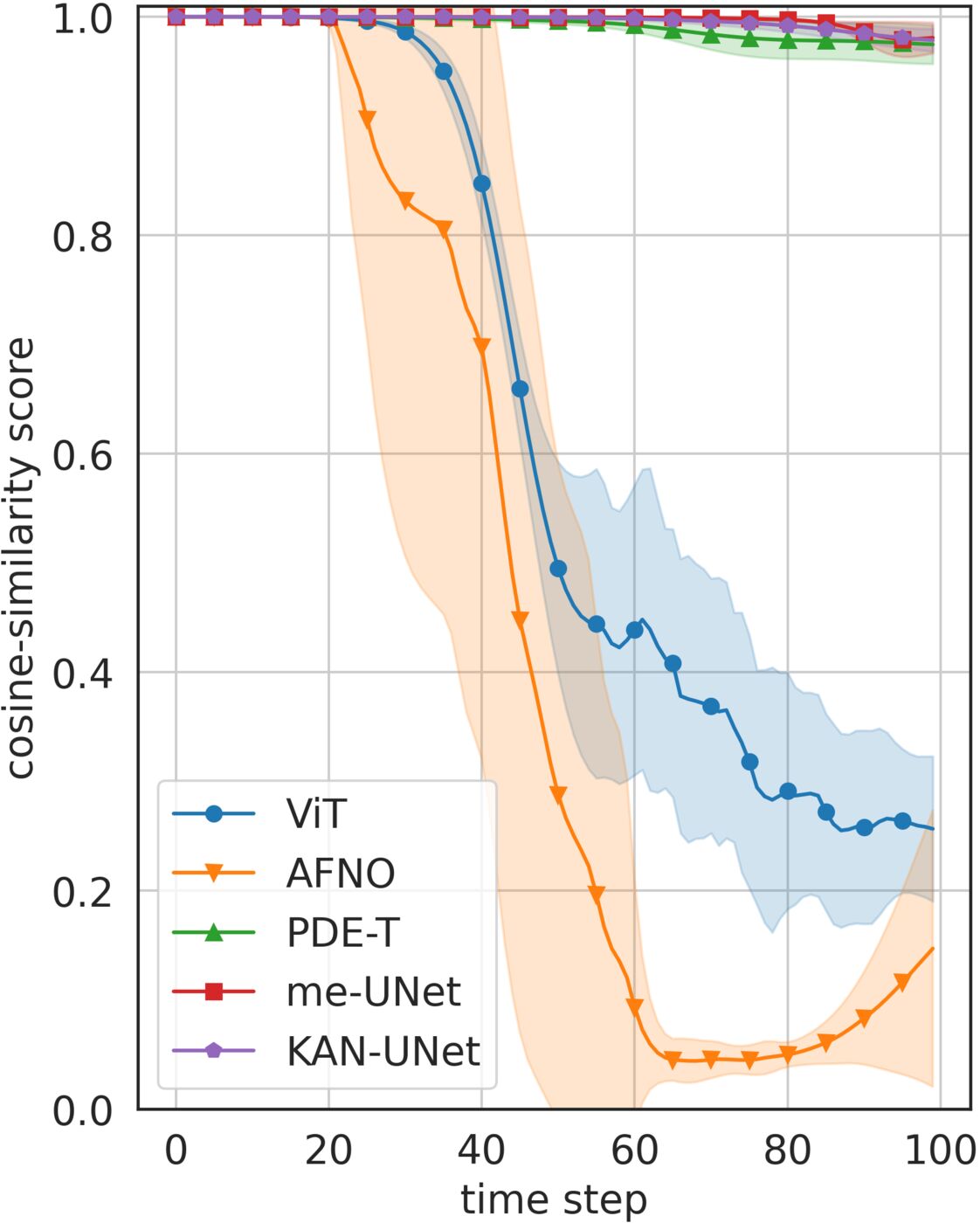}
					\caption{}
				\end{subfigure}%
			}\cr
		}
		\caption{Visualization of the performance of the different neural networks on the DS-6c dataset: (a) autoregressive prediction results; (b) \gls{RMSE}; and (c) cosine similarity of the PSD curves.}
		\label{fig:id_gs_bubbles}
	\end{figure*}
	
	\begin{figure*}[htb!]
		\centering
		\tabskip=0pt
		\valign{#\cr
			\hbox{%
				\begin{subfigure}[b]{.69\textwidth}
					\centering
					\includegraphics[width=1\textwidth]{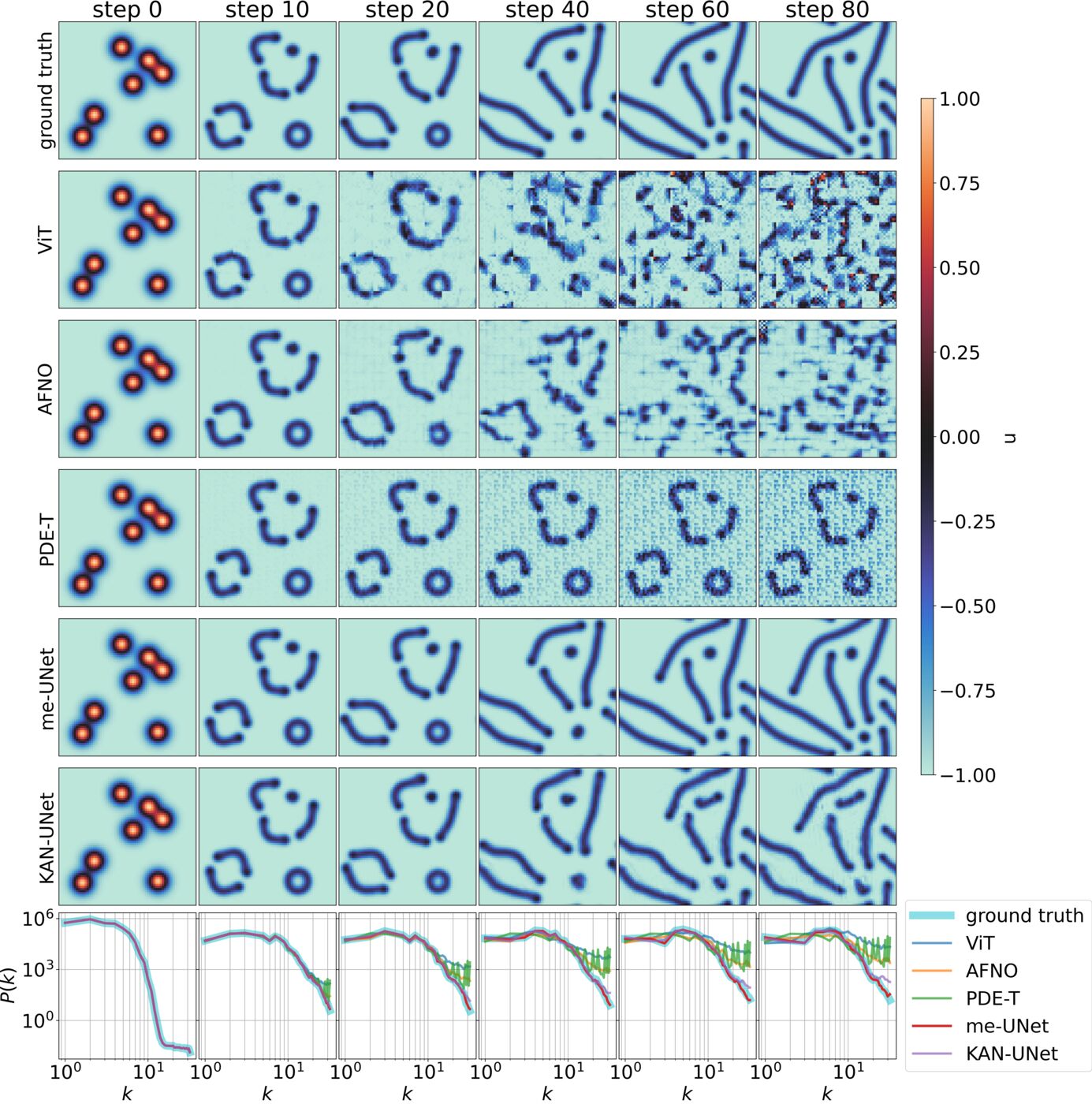}
					\caption{}
				\end{subfigure}%
			}\cr
			\hbox{%
				\begin{subfigure}{.3\textwidth}
					\centering
					\includegraphics[width=0.9\textwidth]{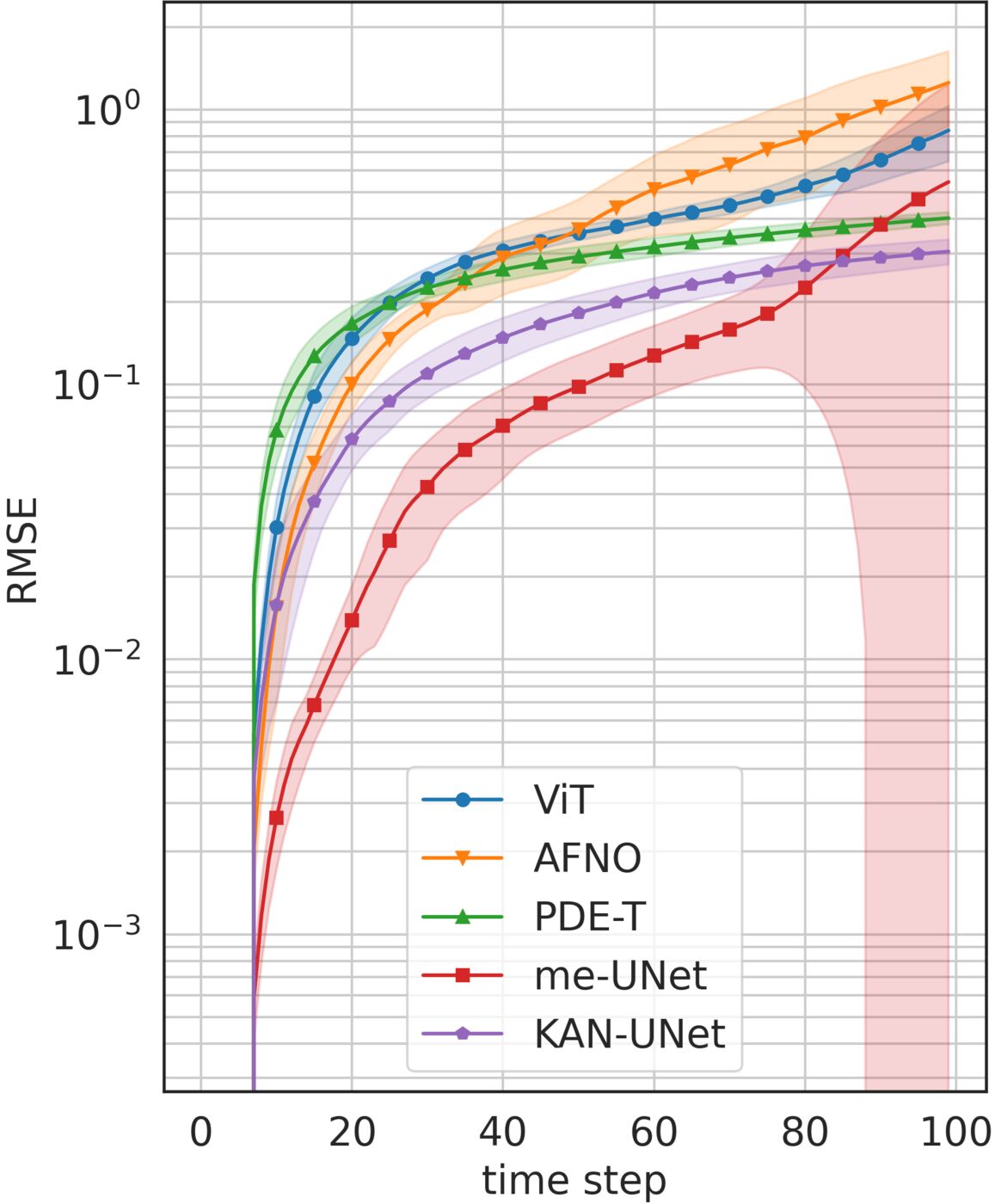}
					\caption{}
				\end{subfigure}%
			}\vfill
			\hbox{%
				\begin{subfigure}{.3\textwidth}
					\centering
					\includegraphics[width=0.9\textwidth]{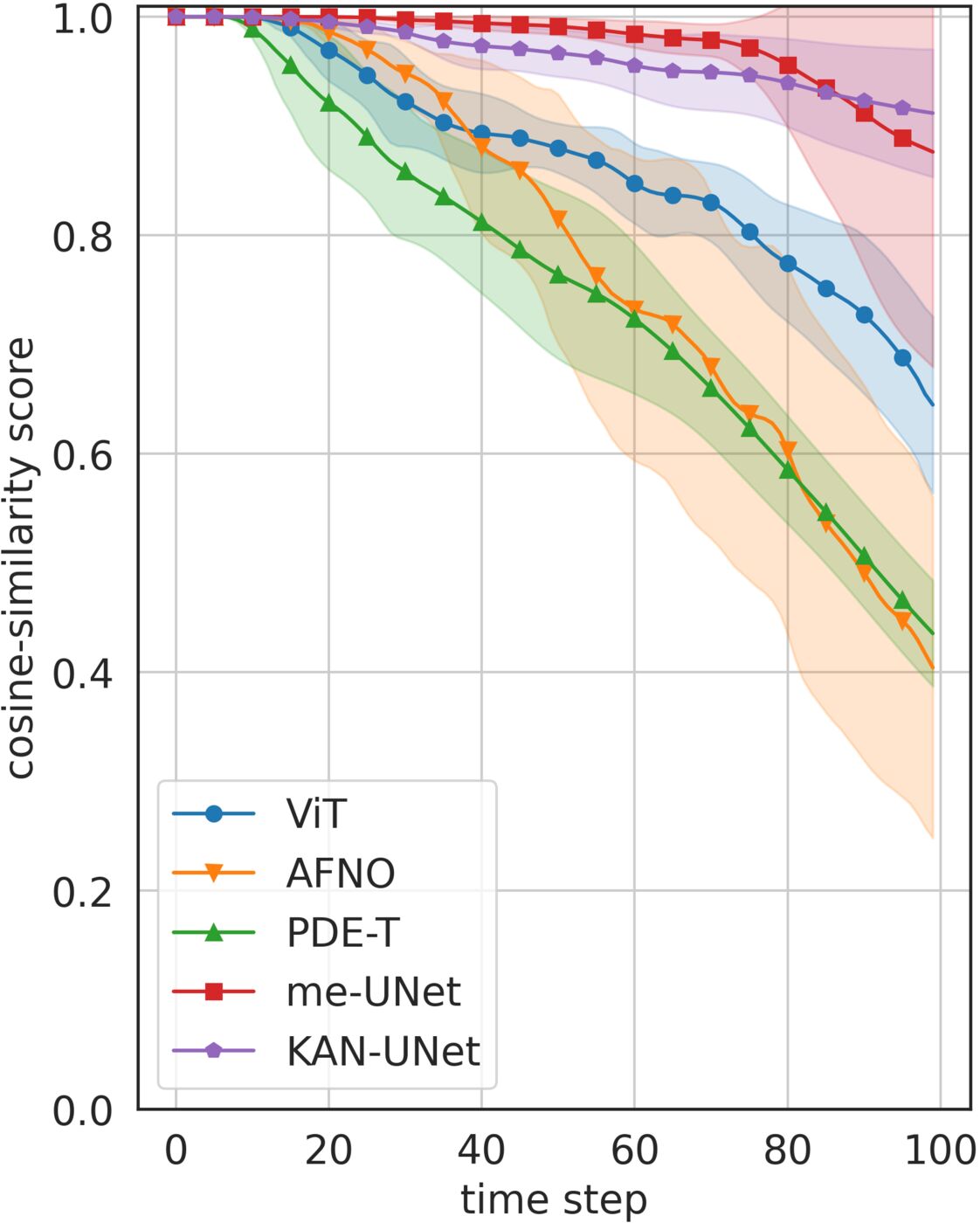}
					\caption{}
				\end{subfigure}%
			}\cr
		}
		\caption{Visualization of the performance of the different neural networks on the DS-6d dataset: (a) autoregressive prediction results; (b) \gls{RMSE}; and (c) cosine similarity of the PSD curves.}
		\label{fig:id_gs_worms_mu}
	\end{figure*}
	
	\clearpage
	\section{Visualization of \gls{RMSE} and cosine similarity for \gls{ood} predictions}
	\label{app:ood_rmse_cos_sim}
	
\begin{figure*}[h!tb]
		\centering
		\begin{subfigure}{0.9\textwidth}
			\centering
			\includegraphics[width=0.7\textwidth]{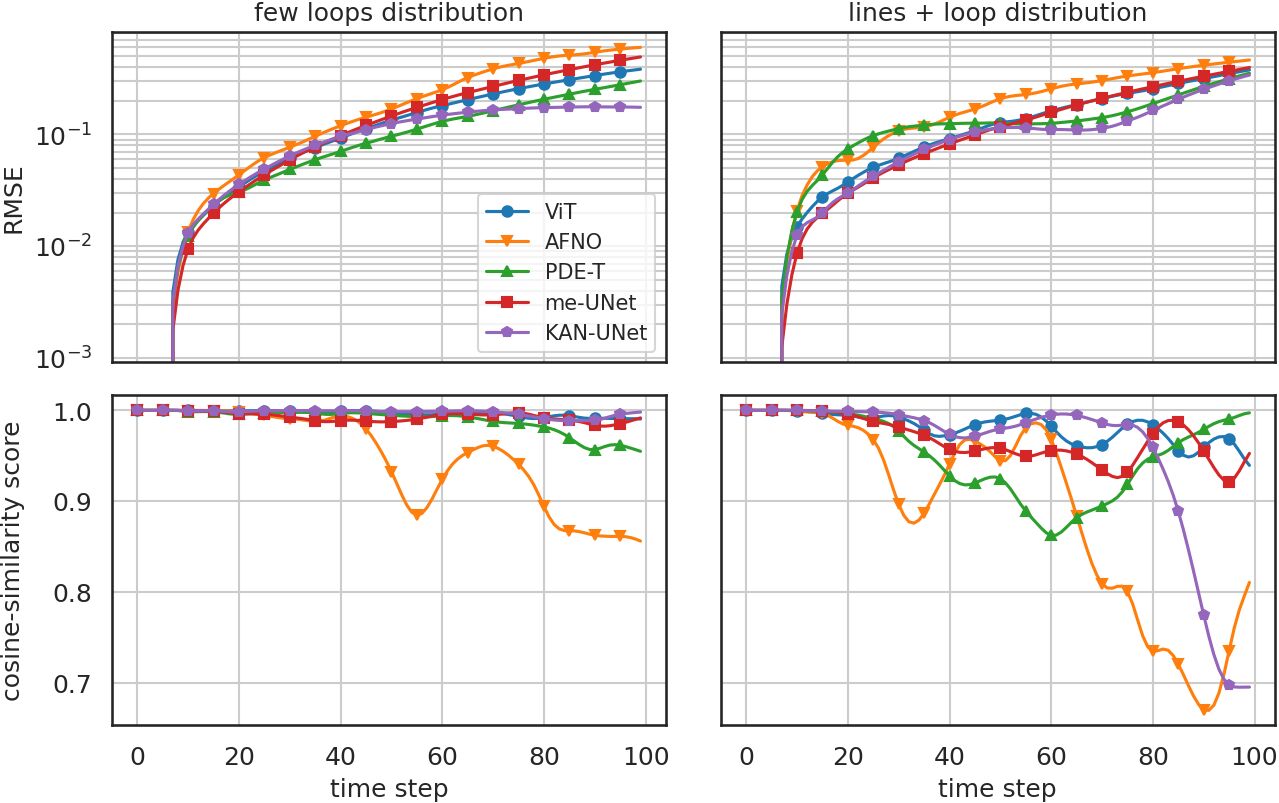}\\
			\caption{}
		\end{subfigure}
		\begin{subfigure}{0.9\textwidth}
			\centering
			\includegraphics[width=0.7\textwidth]{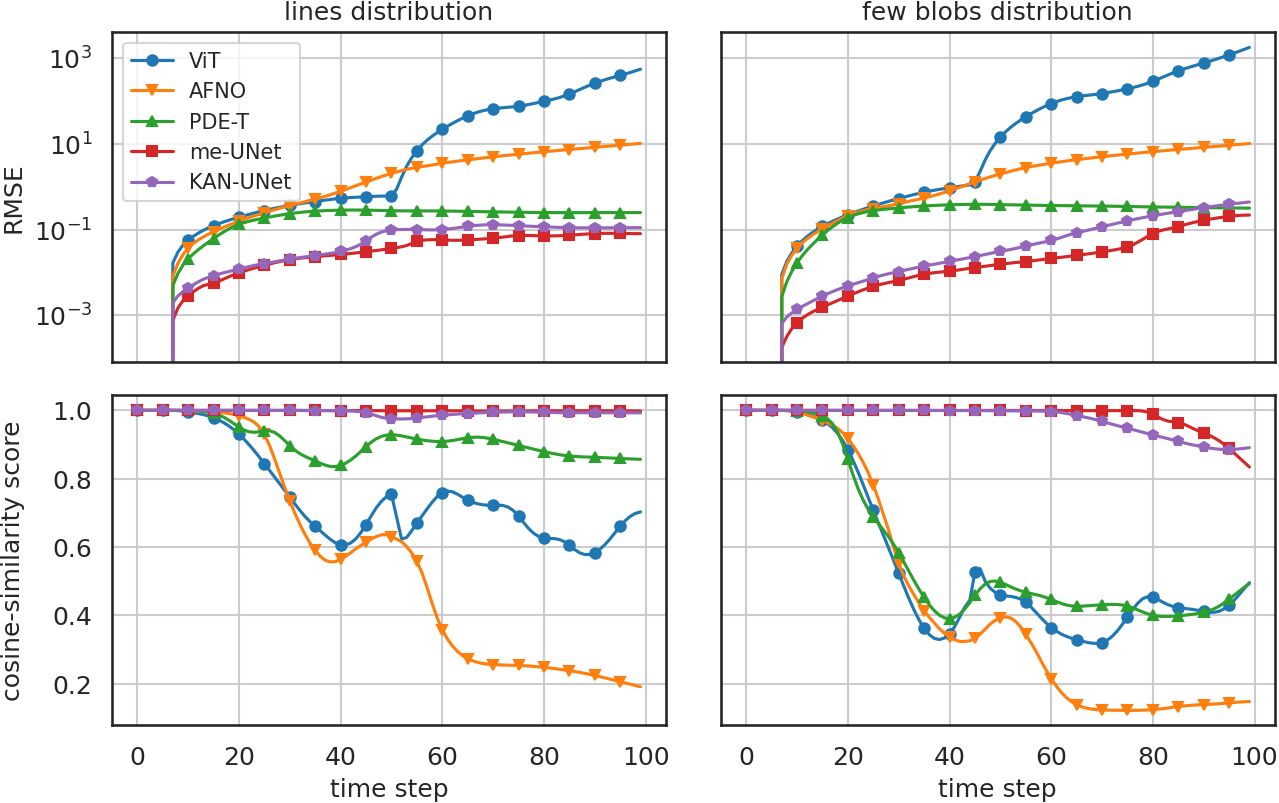}\\
			\caption{}
		\end{subfigure}
		\caption{\gls{ood} performance of the different neural network architectures on (a) the \gls{CDD} model and (b) the Gray--Scott model (DS-6a), in terms of \gls{RMSE} and PSD cosine similarity.}
		\label{fig:ood_rmse_cos_sim}
	\end{figure*}

	As expected from the main results in \cref{sec:results}, me-UNet lies near the bottom of the error curves and near the top of the cosine-similarity curves in both panels, 
	while ViT and AFNO show much larger degradation under \gls{ood} initial conditions. Fig.~\ref{fig:ood_rmse_cos_sim} shows these measures for all models and all simulation time steps.

	\clearpage
	\section{Visualization of me-UNet behavior during training (Grad-CAM}
	\label{app:grad_cam_results}
	In what follows we provide full Grad-CAM visualizations for all datasets, complementing the qualitative discussion in \cref{sec:results}. 
	For each dataset, the top panel shows Grad-CAM maps for all encoder/decoder blocks over training epochs, 
	and the bottom panel shows the average Grad-CAM value per block. These plots illustrate how attention concentrates on physically relevant regions 
	over the course of training and how different datasets activate different parts of the U-Net. Note that during each training epoch \emph{all} simulation steps are simultaneously considered by the model; therefore, the early epochs \emph{cannot} be correlated with the early simulation time steps.
	
\begin{figure*}[htb!]
		\centering
		\tabskip=0pt
		\halign{#\cr
			\hbox{%
				\begin{subfigure}[b]{.9\textwidth}
					\centering
					\includegraphics[width=0.9\textwidth]{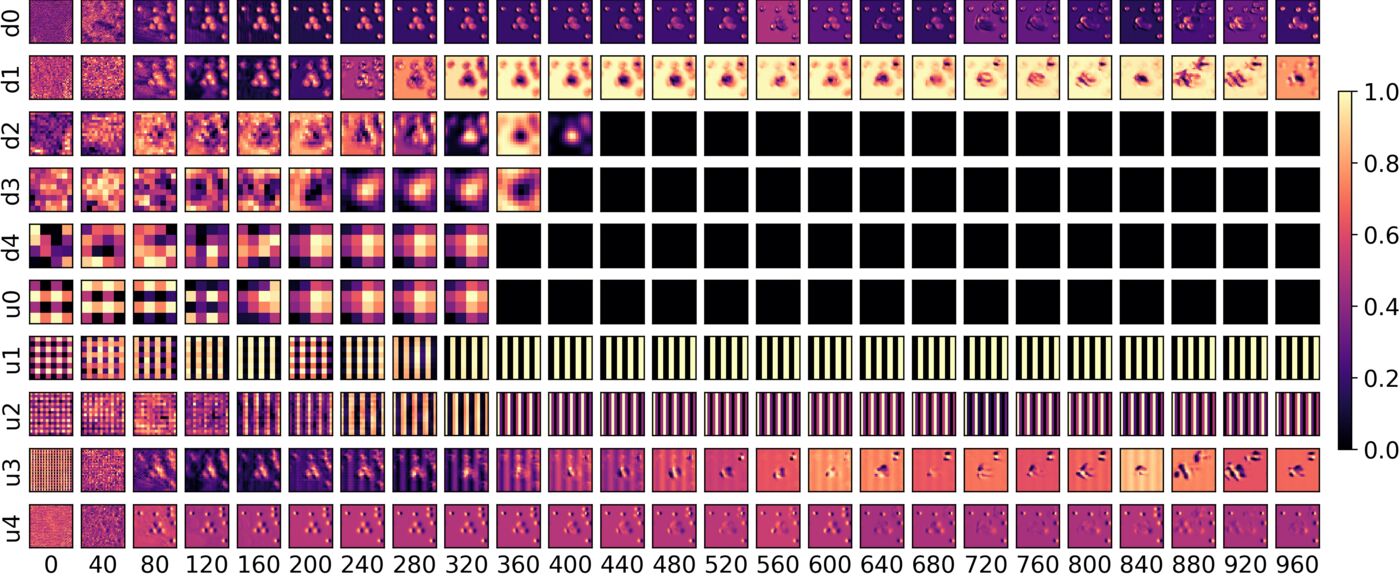}
					\caption{}
				\end{subfigure}%
			}\cr
			\hbox{%
				\begin{subfigure}{.9\textwidth}
					\centering
					\includegraphics[width=0.9\textwidth]{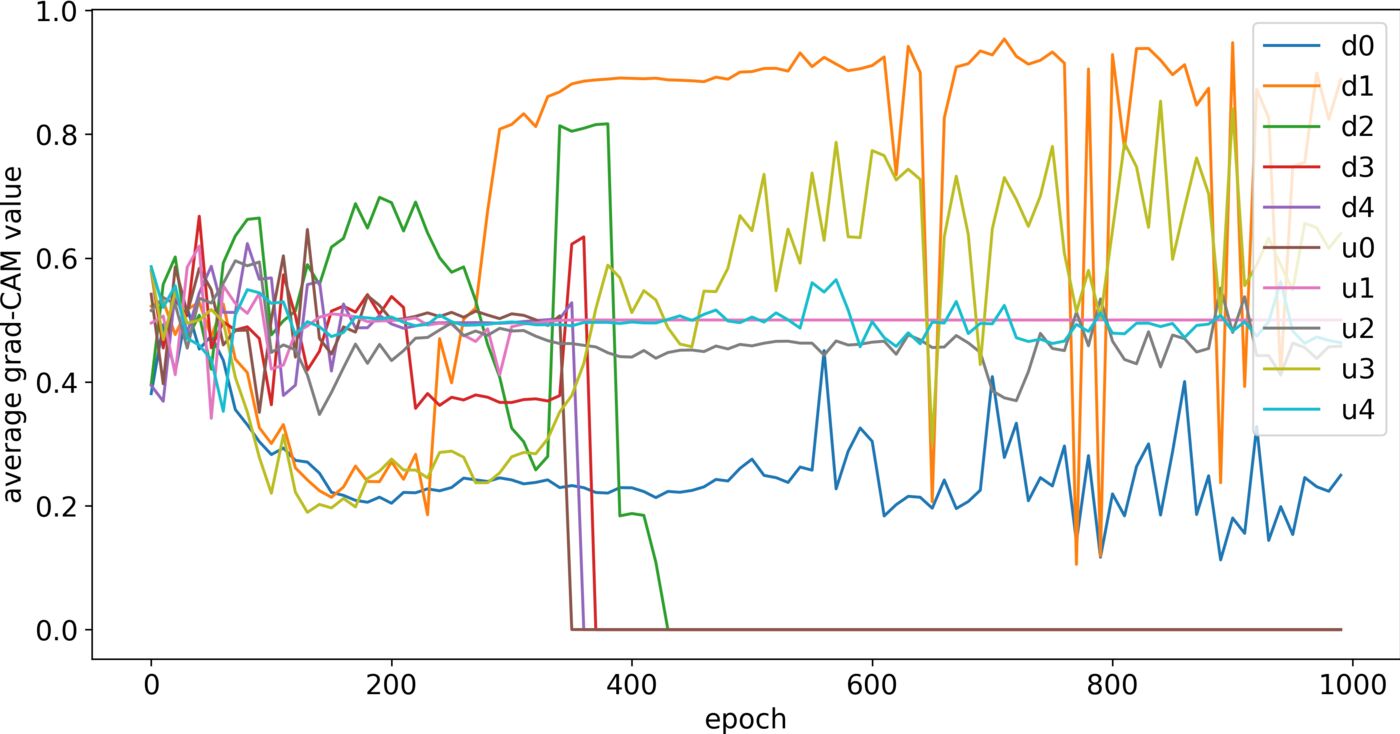}
					\caption{}
				\end{subfigure}%
			}\cr
		}
		\caption{Behavior of me-UNet during training on the DS-1 dataset: (a) Grad-CAM maps over training epochs; (b) average Grad-CAM value per block as a function of epoch.}
		\label{fig:grad_cam_adv}
	\end{figure*}

	\paragraph{DS-1 (\cref{fig:grad_cam_adv})}
	Across the $1000$ training epochs, block layers d0, d1, u1, u2, u3, and u4 remain active, whereas layers d2, d3, d4 and u0 become inactive at epochs $440$, $400$, $360$ and $360$, respectively.
	This dataset exhibits purely advective behavior in the x-direction. The Grad-CAM visualizations reveal a pronounced and consistent vertical streak structure, most clearly visible in u1 and u2, corresponding to the transport of objects along the x-axis. If, instead, we would have had transport along the y-axis, horizontal streaks would occur (not shown here).
	
\begin{figure*}[htb!]
		\centering
		\tabskip=0pt
		\halign{#\cr
			\hbox{%
				\begin{subfigure}[b]{.9\textwidth}
					\centering
					\includegraphics[width=0.9\textwidth]{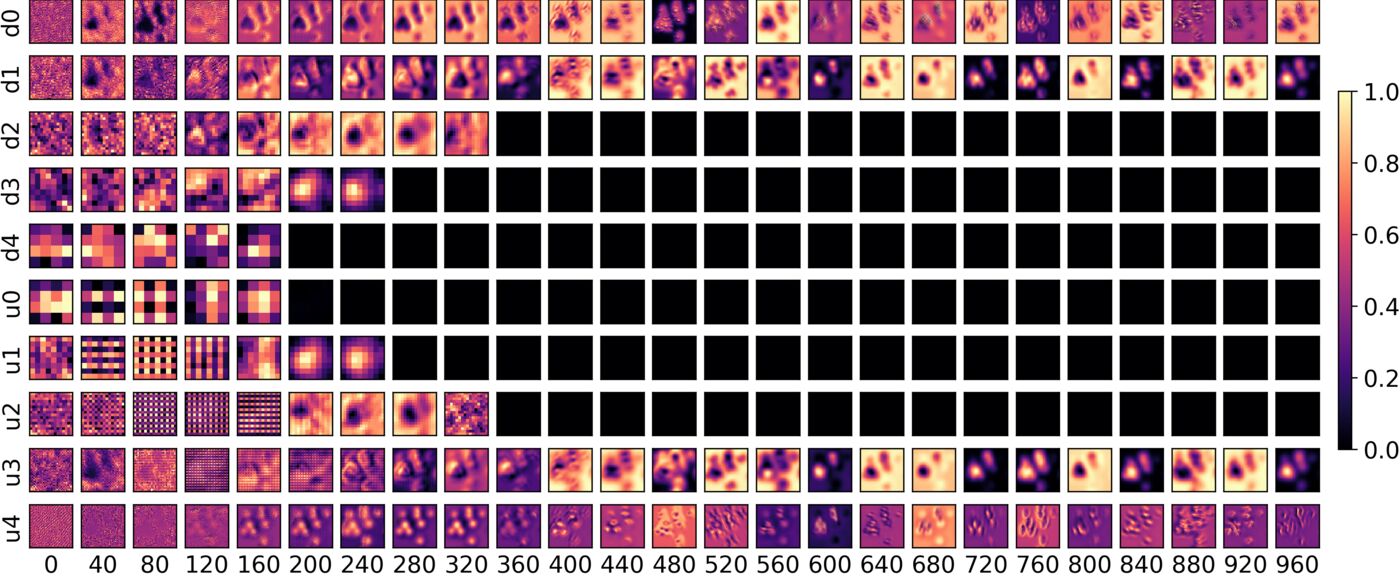}
					\caption{}
				\end{subfigure}%
			}\cr
			\hbox{%
				\begin{subfigure}{.9\textwidth}
					\centering
					\includegraphics[width=0.9\textwidth]{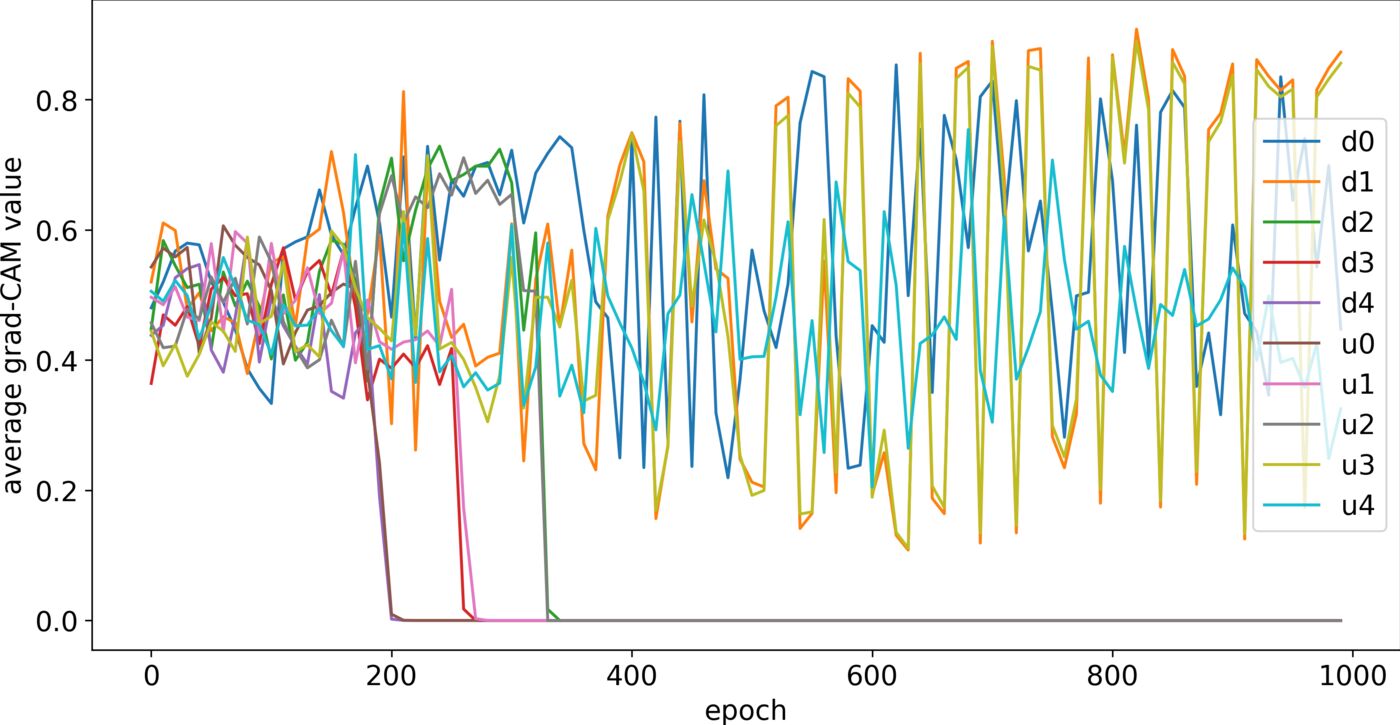}
					\caption{}
				\end{subfigure}%
			}\cr
		}
		\caption{Behavior of me-UNet during training on the DS-2 dataset: (a) Grad-CAM maps; (b) average Grad-CAM value per block over epochs.}
		\label{fig:grad_cam_diff}
	\end{figure*}

	\paragraph{DS-2 (\cref{fig:grad_cam_diff})}
	Throughout 1000 training epochs, layers d0, d1, u3, and u4 remain active, while layers d2, d3, d4, u0, u1, and u2 deactivate at epochs $360$, $280$, $200$, $200$, $280$ and $360$, respectively.
	This dataset exhibits purely diffusive behavior. However, diffusion is a type of  transport process where, e.g., a concentration is moved along both directions. Initially, this transport phenomenon is particularly pronounced, producing characteristic cross-shaped patterns in u1 and u2. As training progresses, the model transitions to alternative feature representations, reflecting that diffusive transport is slowing down such that only minimal state changes occur.
	
\begin{figure*}[htb!]
		\centering
		\tabskip=0pt
		\halign{#\cr
			\hbox{%
				\begin{subfigure}[b]{.9\textwidth}
					\centering
					\includegraphics[width=0.9\textwidth]{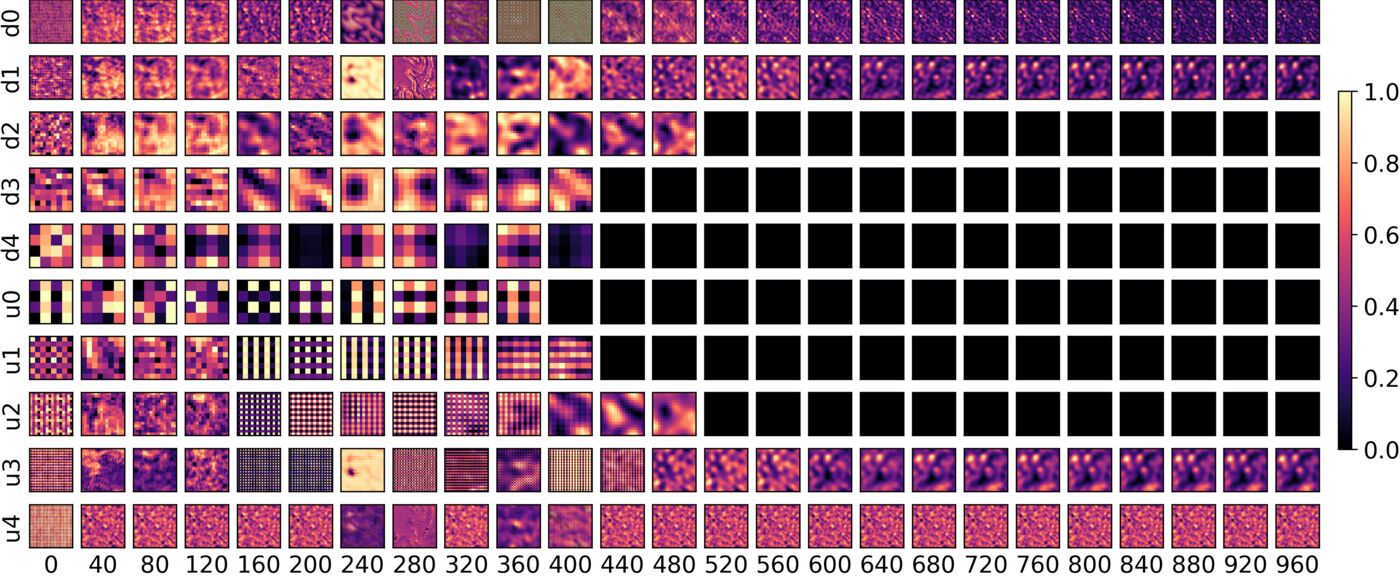}
					\caption{}
				\end{subfigure}%
			}\cr
			\hbox{%
				\begin{subfigure}{.9\textwidth}
					\centering
					\includegraphics[width=0.9\textwidth]{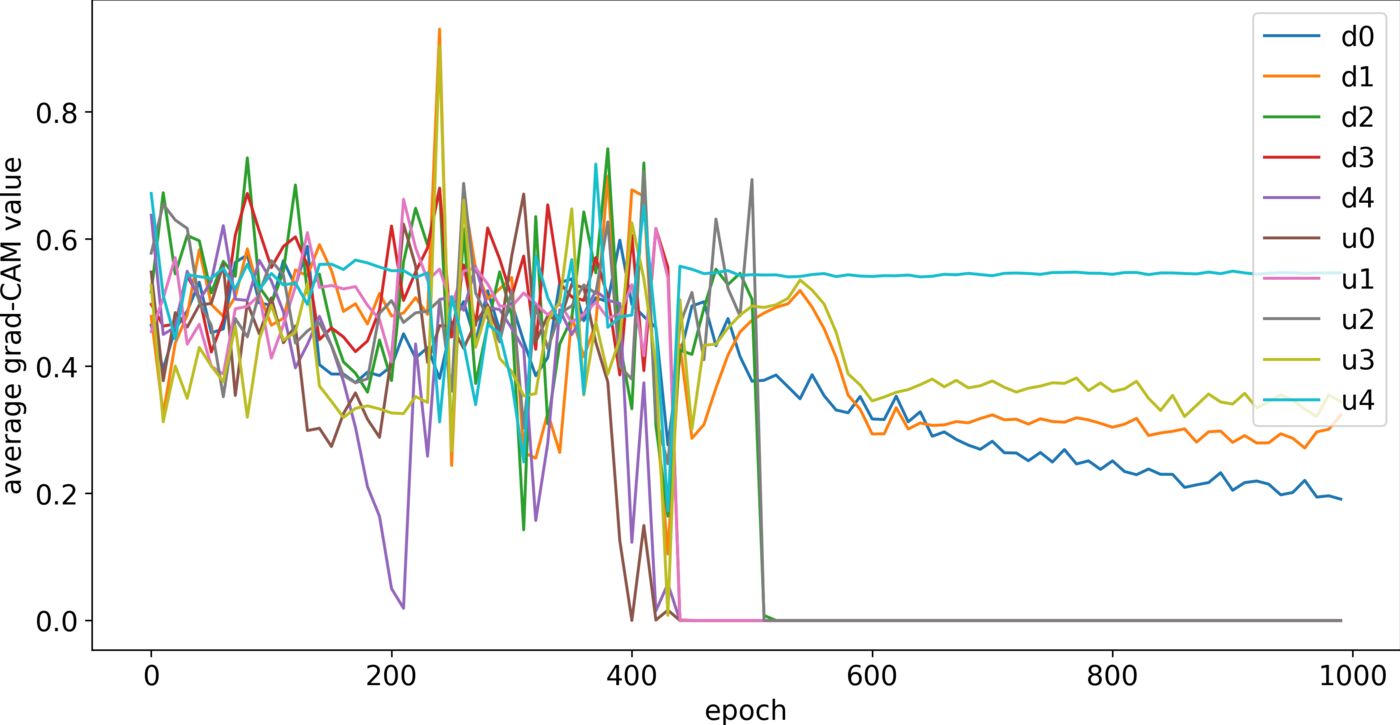}
					\caption{}
				\end{subfigure}%
			}\cr
		}
		\caption{Behavior of me-UNet during training on the DS-3a dataset: (a) Grad-CAM maps; (b) average Grad-CAM value per block over epochs.}
		\label{fig:grad_cam_cdd_adv}
	\end{figure*}

	\paragraph{DS-3a (\cref{fig:grad_cam_cdd_adv})}
	During the 1000-epoch training, layers d0, d1, u3, and u4 remain active, while layers d2, d3, d4, u0, u1, and u2 deactivate at epochs $520$, $440$, $400$, $400$, $440$ and $520$, respectively.
	This dataset contains bidirectional transport. Consistent with this, the Grad-CAM maps display cross-pattern structures across multiple scales of the network.
	
\begin{figure*}[htb!]
		\centering
		\tabskip=0pt
		\halign{#\cr
			\hbox{%
				\begin{subfigure}[b]{.9\textwidth}
					\centering
					\includegraphics[width=0.9\textwidth]{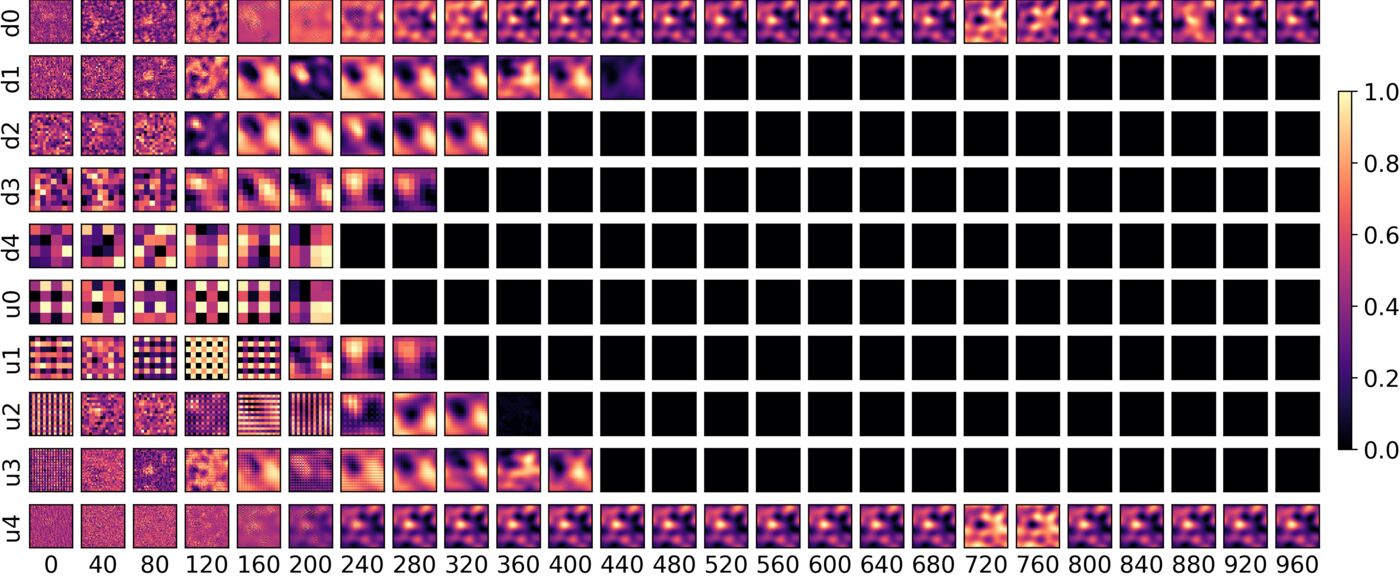}
					\caption{}
				\end{subfigure}%
			}\cr
			\hbox{%
				\begin{subfigure}{.9\textwidth}
					\centering
					\includegraphics[width=0.9\textwidth]{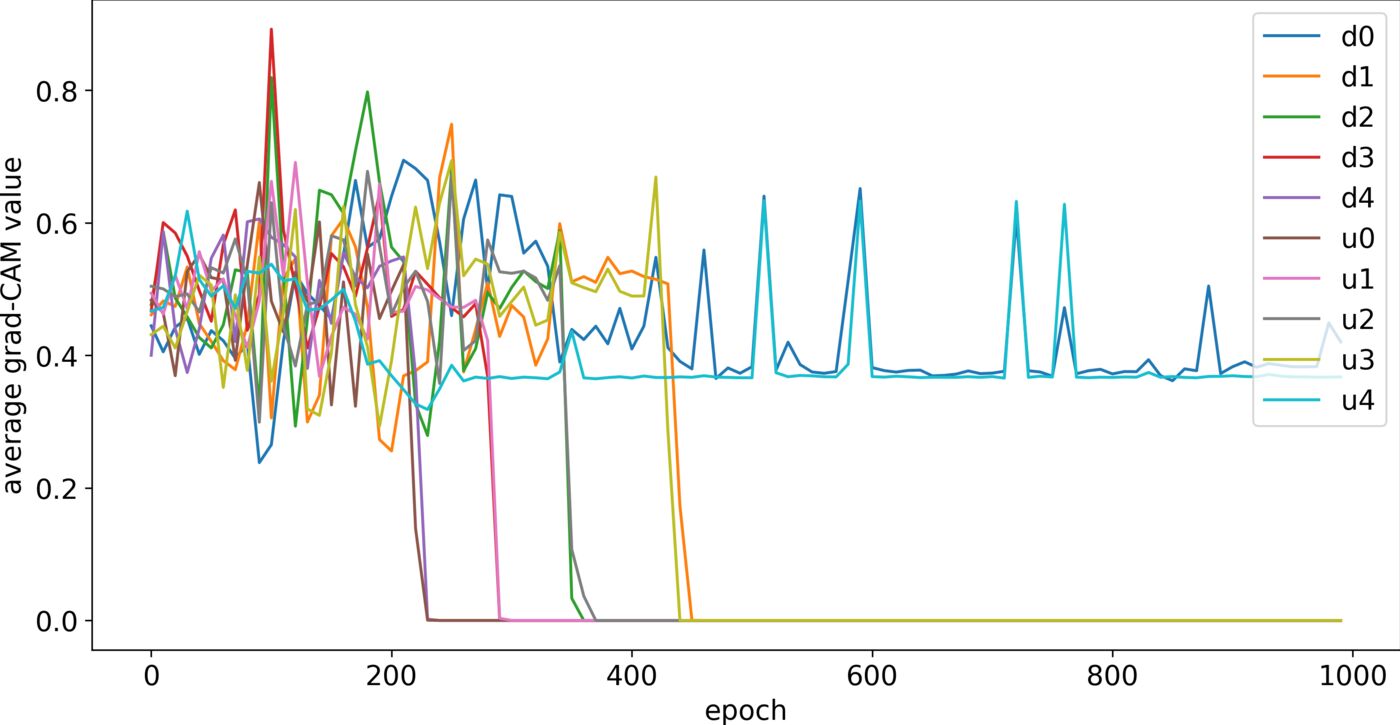}
					\caption{}
				\end{subfigure}%
			}\cr
		}
		\caption{Behavior of me-UNet during training on the DS-3b dataset: (a) Grad-CAM maps; (b) average Grad-CAM value per block over epochs.}
		\label{fig:grad_cam_cdd_diff}
	\end{figure*}

	\paragraph{DS-3b (\cref{fig:grad_cam_cdd_diff})}
	Across the 1000 epochs, only layers d0 and u4 remain consistently active. Layers d1–d4 and u0–u3 deactivate at epochs $440$, $360$, $320$, $240$, $240$, $320$, $360$ and $440$, respectively.
	As in DS-2, the presence of transport in both directions leads to cross-pattern activations in u1 and u2.
	
\begin{figure*}[htb!]
		\centering
		\tabskip=0pt
		\halign{#\cr
			\hbox{%
				\begin{subfigure}[b]{.9\textwidth}
					\centering
					\includegraphics[width=0.9\textwidth]{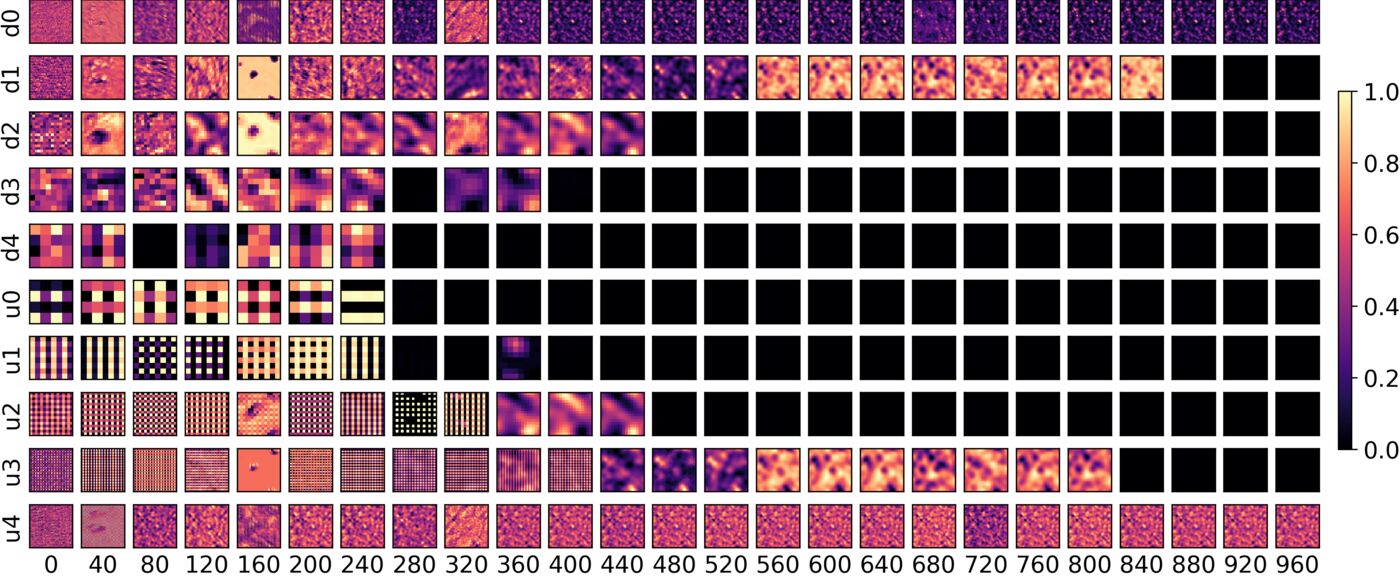}
					\caption{}
				\end{subfigure}%
			}\cr
			\hbox{%
				\begin{subfigure}{.9\textwidth}
					\centering
					\includegraphics[width=0.9\textwidth]{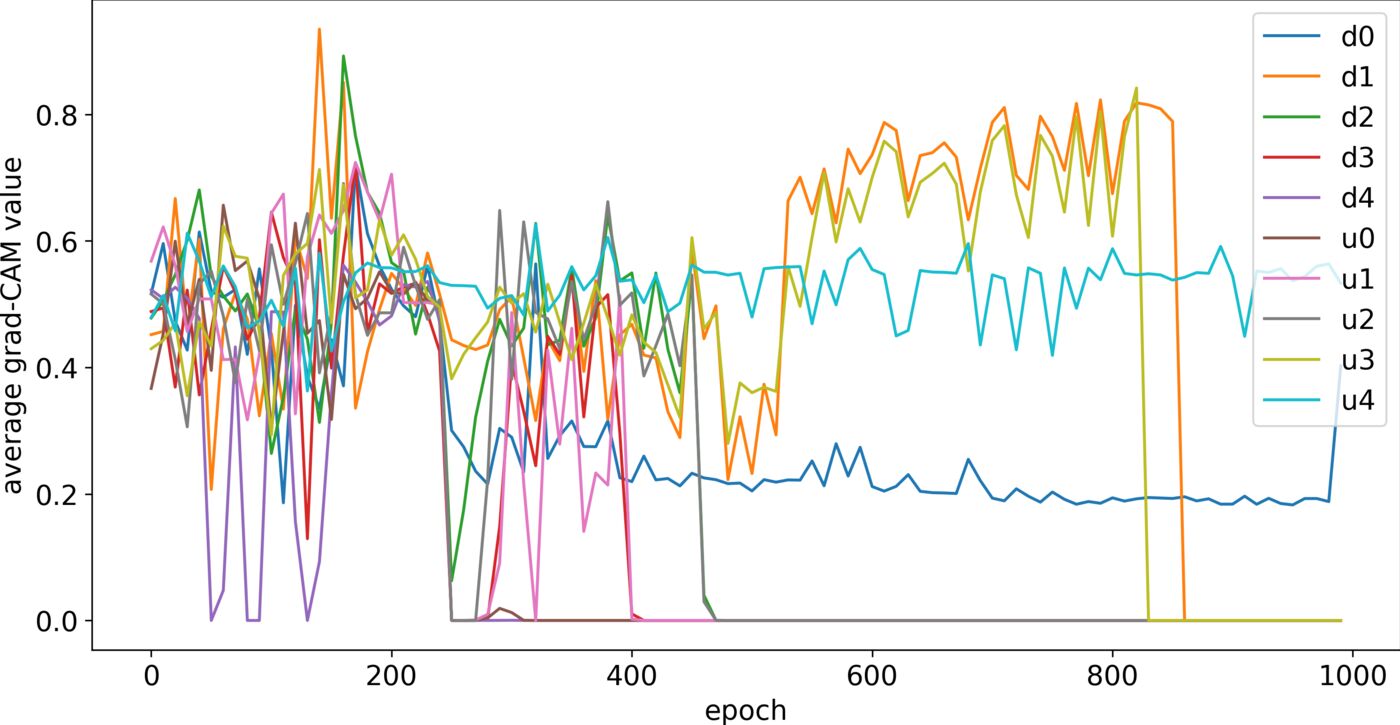}
					\caption{}
				\end{subfigure}%
			}\cr
		}
		\caption{Behavior of me-UNet during training on the DS-4 dataset: (a) Grad-CAM maps; (b) average Grad-CAM value per block over epochs.}
		\label{fig:grad_cam_cdd}
	\end{figure*}

	\paragraph{DS-4 (\cref{fig:grad_cam_cdd})}
	Throughout training, layers d0 and u4 stay active, while layers d1–d4 and u0–u3 deactivate at epochs $880$, $480$, $400$, $280$, $280$, $400$, $480$ and $840$, respectively.
	Here, bidirectional transport is intrinsic to the governing equations, yielding pronounced and persistent cross-patterns in layers u1, u2, and u3.
	
\begin{figure*}[htb!]
		\centering
		\tabskip=0pt
		\halign{#\cr
			\hbox{%
				\begin{subfigure}[b]{.9\textwidth}
					\centering
					\includegraphics[width=0.9\textwidth]{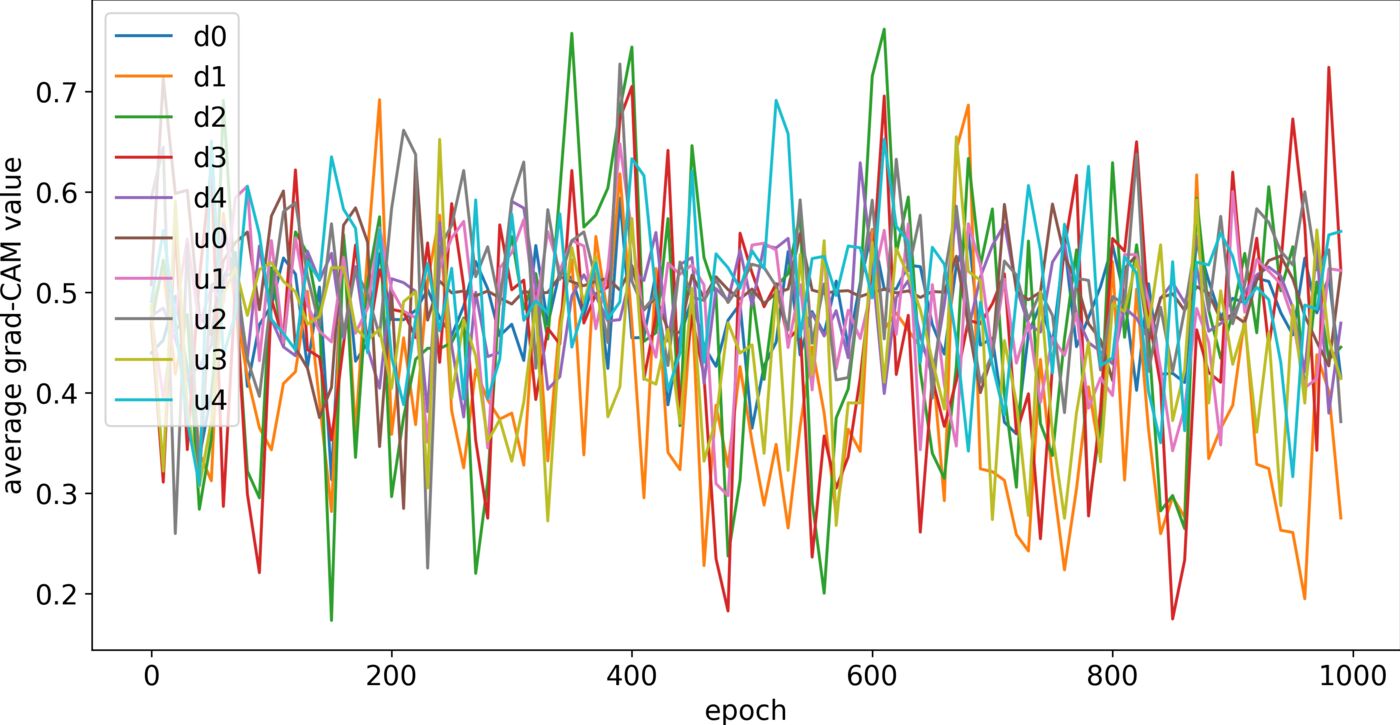}
					\caption{}
				\end{subfigure}%
			}\cr
			\hbox{%
				\begin{subfigure}{.9\textwidth}
					\centering
					\includegraphics[width=0.9\textwidth]{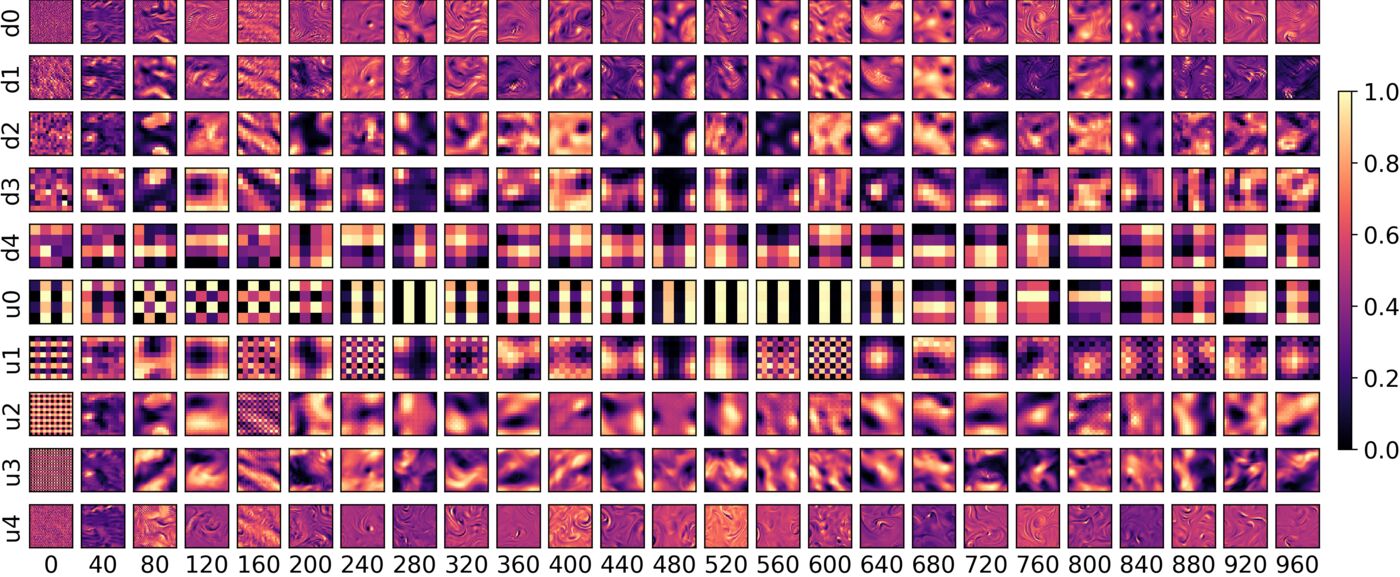}
					\caption{}
				\end{subfigure}%
			}\cr
		}
		\caption{Behavior of me-UNet during training on the DS-5 dataset: (a) Grad-CAM maps; (b) average Grad-CAM value per block over epochs.}
		\label{fig:grad_cam_cfd}
	\end{figure*}

	\paragraph{DS-5 (\cref{fig:grad_cam_cfd})}
	All block layers remain active over the 1000 training epochs.
	Although the dataset contains complex transport phenomena, the Grad-CAM maps still exhibit cross-pattern structures, particularly in u1 and u2, albeit embedded within more complex feature combinations. From the gradCAM visualization we can clearly see that a number of different physical phenomena are active throughtout the whole time range of the simulation.
	
\begin{figure*}[htb!]
		\centering
		\tabskip=0pt
		\halign{#\cr
			\hbox{%
				\begin{subfigure}[b]{.9\textwidth}
					\centering
					\includegraphics[width=0.9\textwidth]{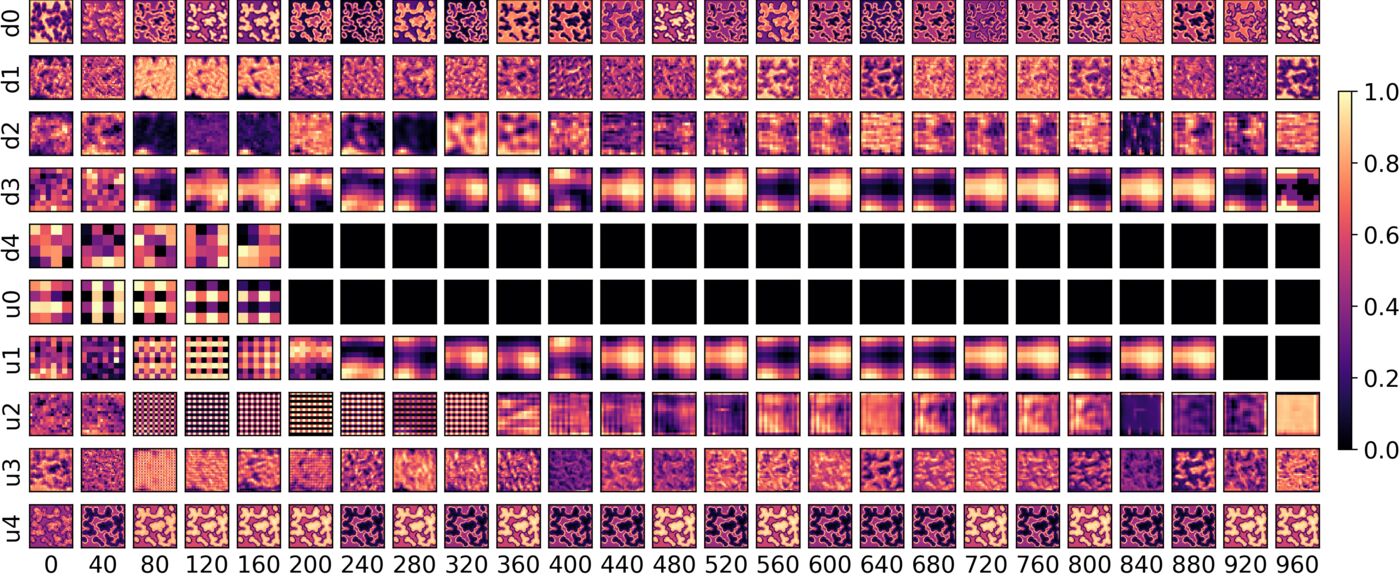}
					\caption{}
				\end{subfigure}%
			}\cr
			\hbox{%
				\begin{subfigure}{.9\textwidth}
					\centering
					\includegraphics[width=0.9\textwidth]{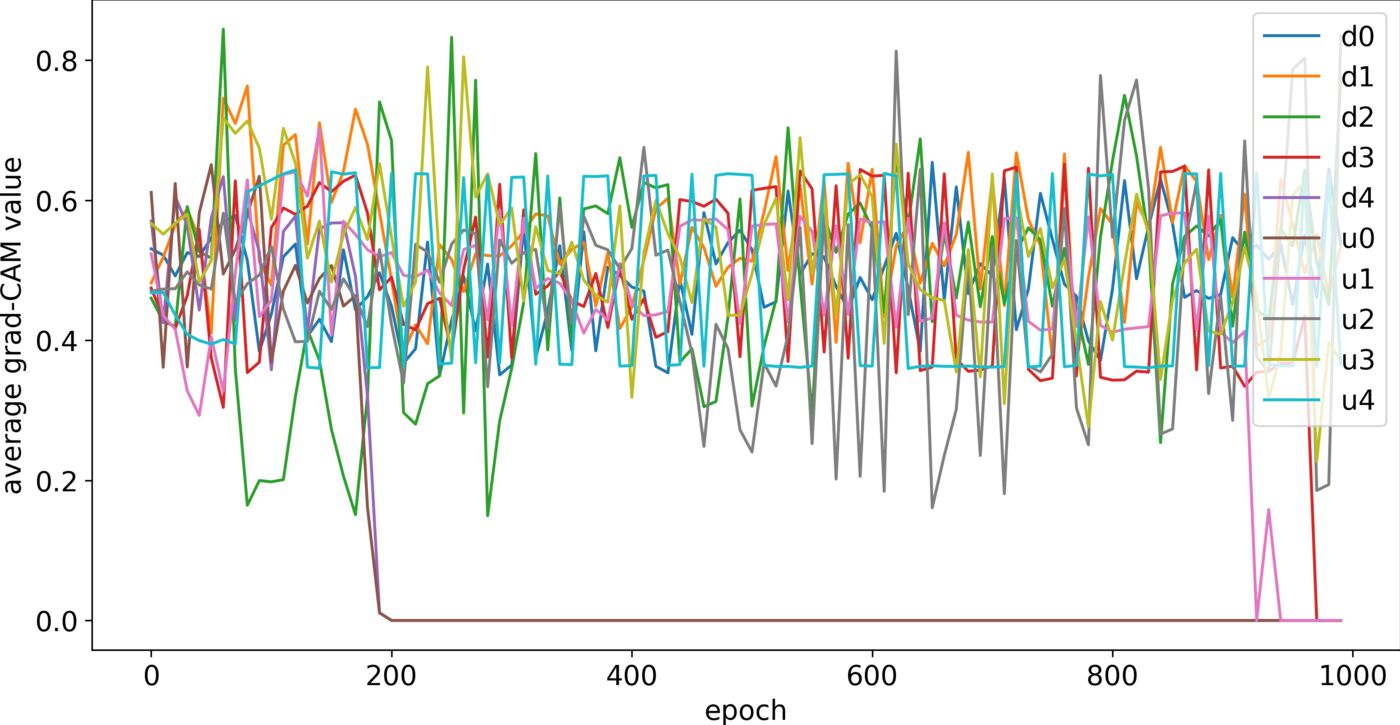}
					\caption{}
				\end{subfigure}%
			}\cr
		}
		\caption{Behavior of me-UNet during training on the DS-6a dataset: (a) Grad-CAM maps; (b) average Grad-CAM value per block over epochs.}
		\label{fig:grad_cam_gs_maze_d2}
	\end{figure*}

	\paragraph{DS-6a (\cref{fig:grad_cam_gs_maze_d2})}
	Layers d0–d3 and u2–u4 stay active during training, while d4, u0, and u1 deactivate at epochs $200$, $200$ and $920$, respectively.
	The dataset contains bidirectional transport, reflected in the maze-like growth dynamics of the simulation. Correspondingly, cross-pattern features appear in u1 and u2.
	
\begin{figure*}[htb!]
		\centering
		\tabskip=0pt
		\halign{#\cr
			\hbox{%
				\begin{subfigure}[b]{.9\textwidth}
					\centering
					\includegraphics[width=0.9\textwidth]{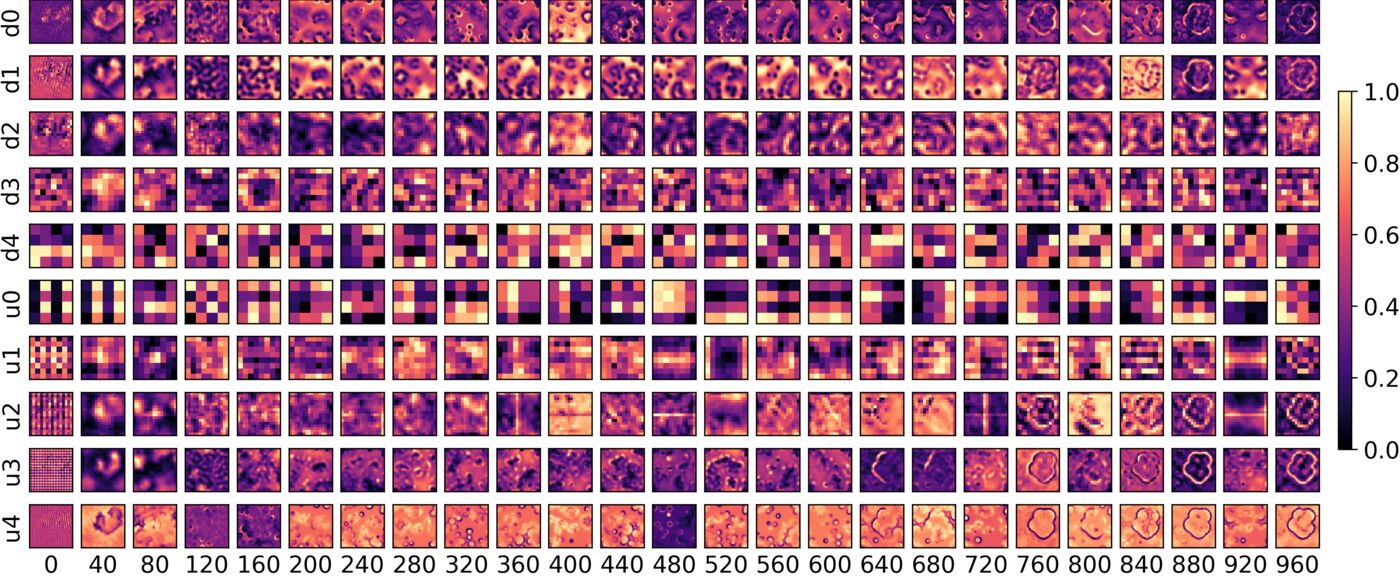}
					\caption{}
				\end{subfigure}%
			}\cr
			\hbox{%
				\begin{subfigure}{.9\textwidth}
					\centering
					\includegraphics[width=0.9\textwidth]{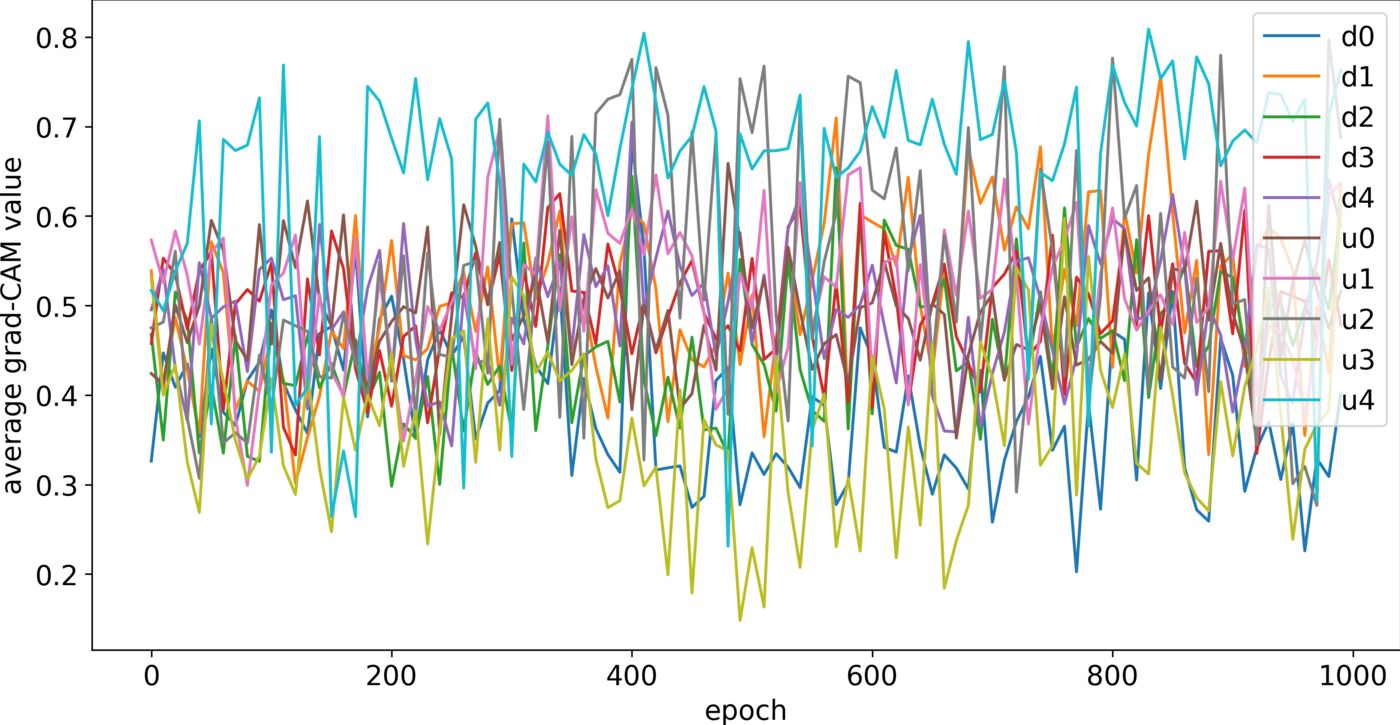}
					\caption{}
				\end{subfigure}%
			}\cr
		}
		\caption{Behavior of me-UNet during training on the DS-6b dataset: (a) Grad-CAM maps; (b) average Grad-CAM value per block over epochs.}
		\label{fig:grad_cam_gs_alpha_v2}
	\end{figure*}

	\paragraph{DS-6b (\cref{fig:grad_cam_gs_alpha_v2})}
	All block layers remain active throughout the 1000 epochs.
	Early in the simulation, expanding blob structures induce transport in both spatial directions, resulting in early-epoch cross-pattern signatures in the Grad-CAM maps before other features become dominant at later stages. As opposed to the previous dataset, the model is still undergoing changes also for later epochs. 
	
\begin{figure*}[htb!]
		\centering
		\tabskip=0pt
		\halign{#\cr
			\hbox{%
				\begin{subfigure}[b]{.9\textwidth}
					\centering
					\includegraphics[width=0.9\textwidth]{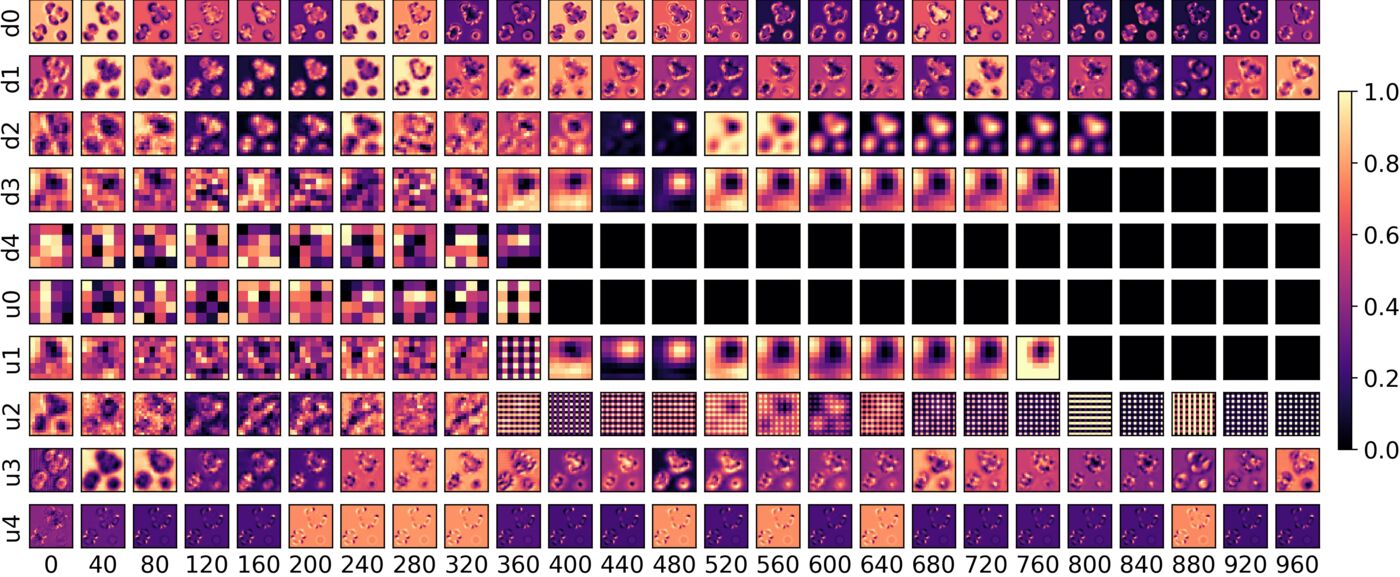}
					\caption{}
				\end{subfigure}%
			}\cr
			\hbox{%
				\begin{subfigure}{.9\textwidth}
					\centering
					\includegraphics[width=0.9\textwidth]{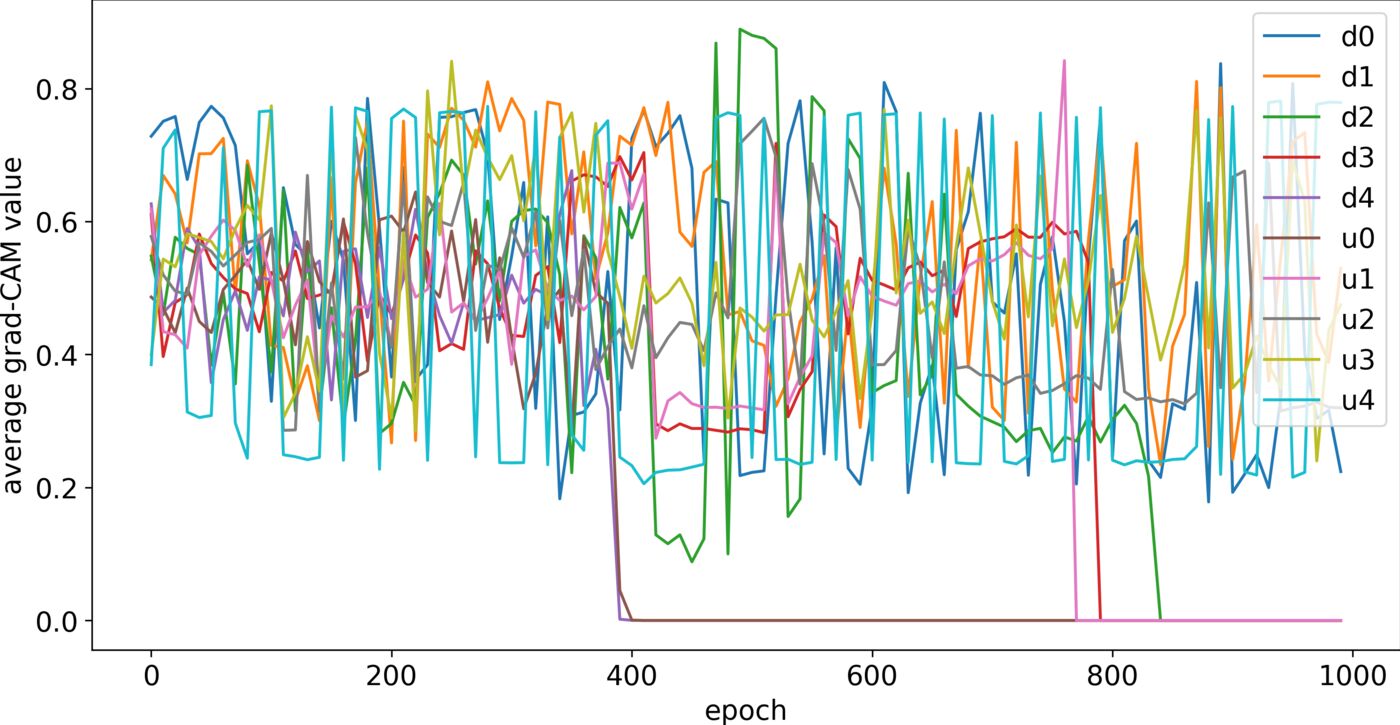}
					\caption{}
				\end{subfigure}%
			}\cr
		}
		\caption{Behavior of me-UNet during training on the DS-6d dataset: (a) Grad-CAM maps; (b) average Grad-CAM value per block over epochs.}
		\label{fig:grad_cam_gs_worms_mu}
	\end{figure*}

	\paragraph{DS-6c (\cref{fig:grad_cam_gs_worms_mu})}
	Layers d0, d1, u3, and u4 stay active, while d2, d3, d4, u0, u1, and u2 deactivate at epochs $640$, $440$, $280$, $440$ and $640$, respectively.
	Bidirectional transport is again present in this dataset. Cross-patterns appear only locally within the Grad-CAM visualizations, indicating a more spatially heterogeneous transport behavior.
	
\begin{figure*}[htb!]
		\centering
		\tabskip=0pt
		\halign{#\cr
			\hbox{%
				\begin{subfigure}[b]{.9\textwidth}
					\centering
					\includegraphics[width=0.9\textwidth]{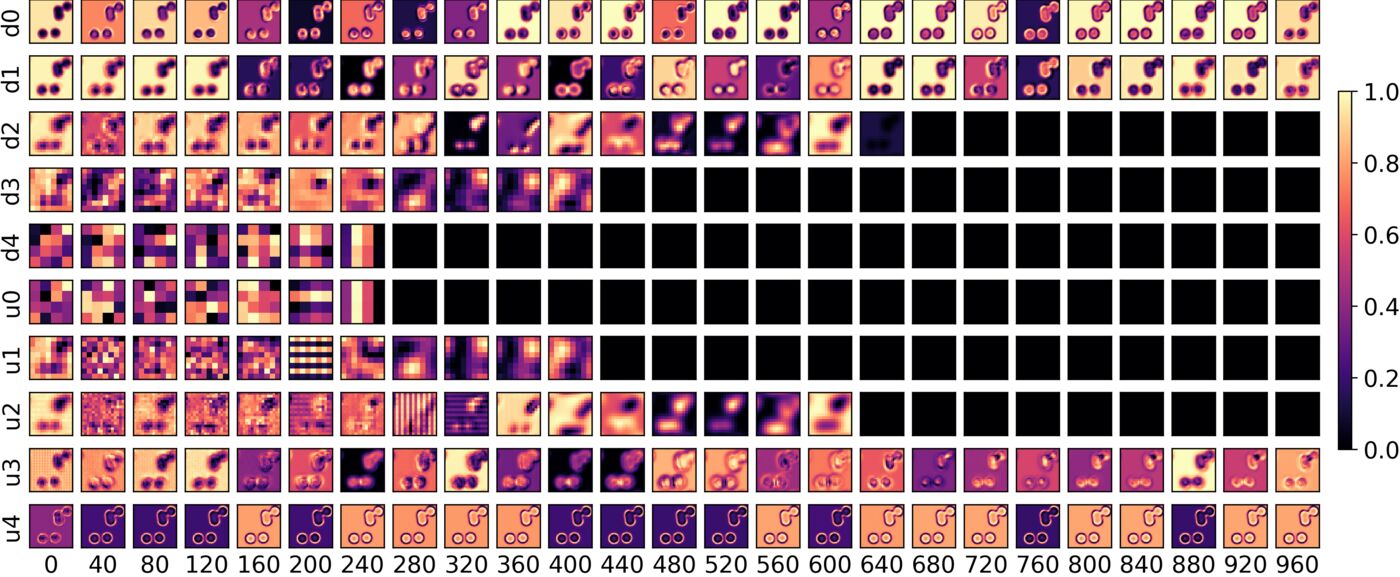}
					\caption{}
				\end{subfigure}%
			}\cr
			\hbox{%
				\begin{subfigure}{.9\textwidth}
					\centering
					\includegraphics[width=0.9\textwidth]{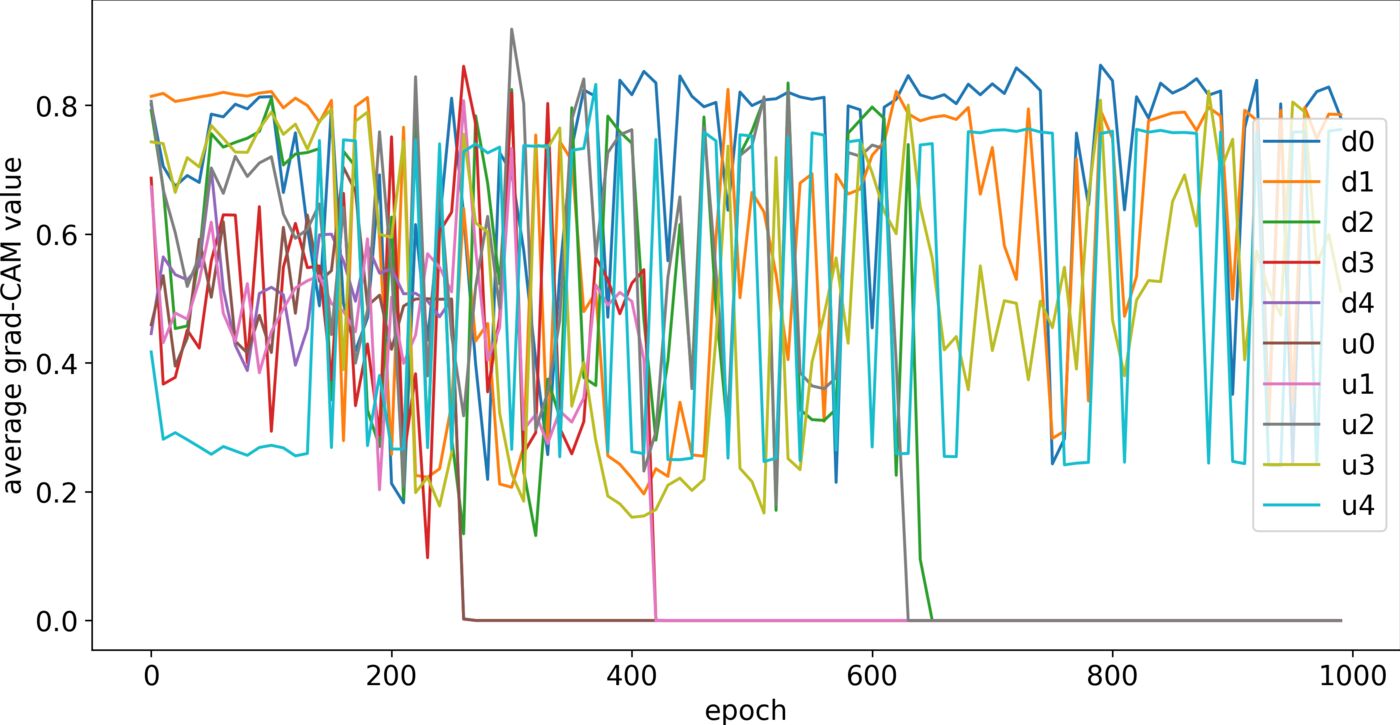}
					\caption{}
				\end{subfigure}%
			}\cr
		}
		\caption{Behavior of me-UNet during training on the DS-6c dataset: (a) Grad-CAM maps; (b) average Grad-CAM value per block over epochs.}
		\label{fig:grad_cam_gs_bubbles}
	\end{figure*}

	\paragraph{DS-6d (\cref{fig:grad_cam_gs_bubbles})}
	Layers d0, d1, u2, u3, and u4 remain active, while layers d2–d4 and u0–u1 deactivate at epochs $840$, $800$, $400$, $400$, and $800$, respectively.
	After an ``incubation period'' the dataset contains strong bidirectional transport dynamics, yielding clearly identifiable cross-patterns in the Grad-CAM maps, with layer u2 even converging to a stable cross-pattern representation. The absense of these patterns during the early phase of training suggests that initially, it was beneficial to focus the training on non-transport aspects.
	
	Overall, there are common transport activities in almost all presented datasets, which result in the appearance of streak- or crossed-pattern, depending on which direction this movement is dominated. Their appearance depends on the amount of training data which contains these behaviors. If a large part of the simulation clearly shows transport activities (e.g., DS-1, DS-3a, or DS-4), then in the grad-CAM visualization these patterns appear often and may even converge (\cref{fig:grad_cam_adv}). If only a small part of the simulation that the transport activities are clearly observed (e.g., DS-2, DS-3b, DS-6b, or DS-6c), the crossed patterns appear at the beginning or only in some places during the training process, then other patterns dominate. 
	
\end{document}